\pdfoutput=1
\documentclass{article}
\usepackage{iclr2027_conference,times}

\usepackage{amsmath,amssymb}
\usepackage{booktabs}
\usepackage{multirow}
\usepackage{array}
\usepackage{tabularx}
\usepackage{longtable}
\usepackage{calc}
\usepackage{adjustbox}
\usepackage{needspace}
\usepackage{ragged2e}
\usepackage{enumitem}
\usepackage{xcolor}
\usepackage{colortbl}
\usepackage{tcolorbox}
\tcbuselibrary{breakable,skins}
\usepackage{threeparttable}
\usepackage{float}
\usepackage{placeins}
\usepackage{chngcntr}
\usepackage[figuresleft]{rotating}
\usepackage{arydshln}
\usepackage{graphicx}
\usepackage{tikz}
\usetikzlibrary{positioning,calc,arrows.meta}
\usepackage{xcolor}
\usepackage{microtype}
\usepackage{hyperref}
\usepackage{url}

\graphicspath{{figures/}}

\author{\parbox[t]{\dimexpr\textwidth-2\tabcolsep\relax}{\raggedright
Jin~Huang$^{1*}$\quad
Diego~Ferreras~Garrucho$^{2*}$\quad
Yutong~Xie$^{1}$\quad
Walter~M.~Yuan$^{3}$\quad
Qiaozhu~Mei$^{1**}$\quad
Chen~Lian$^{4**}$\quad
Jonathon~Hazell$^{2**}$\\[1ex]
{\normalfont\small
$^{1}$University of Michigan, Ann Arbor\quad
$^{2}$London School of Economics\\
$^{3}$MobLab Inc\quad
$^{4}$University of California, Berkeley\\
$^{*}$Equal Contribution\quad
$^{**}$Equal Senior Supervision}}}

\iclrfinalcopy

\usepackage{xspace}
\newcommand{\bench}{HouseholdBench\xspace}

\title{\bench: Evaluating Large Language Models as Predictors of Household Economic Behavior}

\begin{document}

\maketitle
\lhead{Preprint}

\begin{abstract}
    Large language models (LLMs) have the potential to meet a key goal in economics: a quantitative model of household decision  making, across a variety of settings. Yet existing evaluations cover few surveys and outcomes, and do not study how households adjust to changing economic conditions. We introduce a new evaluation, \bench, which unites 6 U.S. household surveys and 32 prediction tasks spanning numeric, categorical and probabilistic outcomes, related to consumption, income, labor, expectations, and housing. Using past behavior, demographics and macroeconomic conditions, the tasks test whether LLMs predict behavior, including how households adjust to changes in various policies.
    We evaluate 13 proprietary and open-weight LLMs against a no-change baseline and a gradient-boosted tree model. Most LLMs outperform the no-change baseline, including for policy response tasks---with the best model lowering error for numeric outcomes by 12.2\%. Across most tasks,  gradient-boosted trees rank first; leading proprietary LLMs approach their performance, but open-weight models lag.
    LLMs exhibit systematic over- and underprediction across different tasks. We identify methods that enable a 4 billion parameter open-weight model to match proprietary models' performance: fine-tuning and aggregating 16 predictions per observation. Improvements generalize to policy-response tasks, which are excluded from fine-tuning. We release our datasets, code, and leaderboard on our website.\footnote{\url{https://jn-huang.github.io/householdbench}}

\end{abstract}

\begin{figure}[!b]
\centering
\input{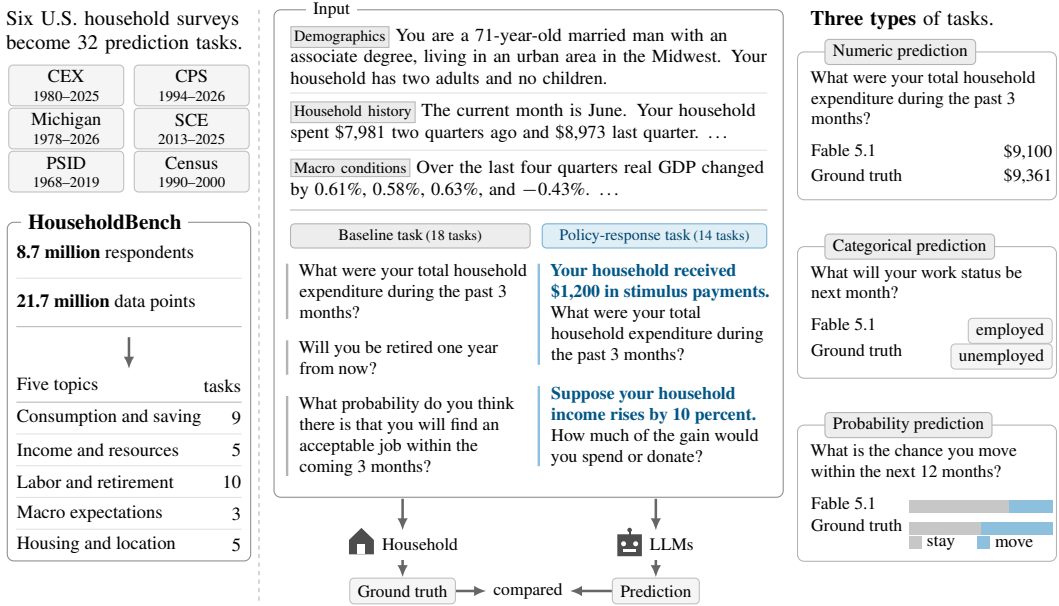}
\caption{\bench turns six household surveys into 32 prediction tasks across five topics.}
\label{fig:teaser}
\end{figure}
\section{Introduction}
\label{sec:introduction}

Large language models (LLMs) may be able to meet a key goal in economics---a quantitative model of how economic agents make decisions. To assess this promise, many papers evaluate whether LLMs can predict decision making in various settings, ranging from social science experiments ~\citep{horton2023large,ashokkumar2026predict}, to survey response ~\citep{argyle2023out,park2024llm}, to behavior in economic games ~\citep{BeFM-1,behaviorbench}. A core question in economics is how households make real-world decisions about labor supply and consumption. Household behavior matters because it is a fundamental driver of aggregate outcomes. For instance, how households consume or save after income shocks determines the aggregate effect of fiscal and monetary policies~\citep{kaplan2018hank,auclert2019redistribution}. If LLMs provide a realistic model of household decision making, then one can use them to carry out realistic simulations of the macroeconomy---as the literature in ``generative agent-based modelling'' has begun to explore ~\citep{li2024econagent,piao2025agentsociety,karten2025llmeconomist}. In particular, one could simulate how various policies---such as stimulus checks or unemployment insurance changes---affect the economy.

There is not yet a comprehensive evaluation of how well LLMs simulate household economic behavior. Recent work evaluates LLMs on household survey data, for instance predicting a respondent's occupation, employment, or retirement~\citep{athey2024laborllm,jia2026personas,garzon2026demographics}, and their income or homeownership~\citep{cruz2024evaluating,gao2026howwell}. However, previous work usually uses few surveys and focuses on few outcomes, meaning the results may be context specific rather than widely applicable. Moreover prior work does not ask whether LLMs can predict how households respond to real-world changes in economic policy---which is critical for realistic policy simulations.

We propose \bench: a comprehensive benchmark for evaluating whether LLMs can predict household economic behavior across a variety of settings.
\bench uses six public U.S. household surveys with rich longitudinal data on household economic behavior and has a total of 21.7M observations (Figure~\ref{fig:teaser}). We construct 32 tasks spanning five topics: consumption and saving, income and resources, labor and retirement, macroeconomic expectations, and housing and location. We cover three types of prediction tasks: numeric prediction, such as total household spending; categorical prediction, such as employment status; and probability prediction, such as households' subjective probability distributions for inflation. To evaluate if LLMs can predict how household behavior adjusts to policy, we include a set of \emph{policy-response tasks}. These tasks ask how households respond to an external, policy-related change such as a stimulus check or a job displacement. Using past behavior, demographics, and macroeconomic conditions, we ask how well each LLM can predict households' future behavior in each task.

We evaluate thirteen LLMs on \bench. Eight are proprietary models from the GPT and Claude families, and five are open-weight models. We compare them to two statistical references: a \emph{no-change baseline} carries forward the most recent value from the household's history; and a gradient-boosted tree ~\citep{chen2016xgboost} trained task-by-task on the same information that LLMs use.

Given the breadth of \bench, we can draw widely applicable conclusions about household behavior. Our main findings are as follows. First, we find that most LLMs outperform the no-change baseline, including on policy response tasks, at least for numerical and categorical outcomes.
Second, widespread across tasks, there is a clear ranking of models. The gradient-boosted tree predicts best (19\% better than the no-change baseline for numeric tasks). Proprietary LLMs from the Claude and GPT families approach this performance, but open-weight models perform worse.
The performance of the best proprietary LLMs is notable, given that gradient-boosted trees achieve good performance on tabular prediction tasks~\citep{holzmuller2024better}.
Third, again widespread across tasks, performance after LLM knowledge cut-offs is equally good, which suggests that the performance of LLMs is not because of data contamination (e.g., public survey microdata may appear in pretraining corpora~\citep{sarkar2024lookahead,ludwig2024large}). Fourth, we document systematic biases of various kinds, with LLMs systematically overestimating household outcomes on some tasks, and underestimating them on others. Fifth, we uncover conditions under which a small, open-weight model can approach the frontier performance. In particular fine tuning Qwen3.5-4B~\citep{qwen3_5}, and aggregating across multiple predictions, greatly improves performance across a range of tasks, for instance reducing the numeric error by 24\% and outperforming the strongest LLM. Sixth, these forecasting improvements generalise to policy response tasks, which are not used for fine tuning. This step is important because policymakers often contemplate new policies, for which there is no existing data for fine-tuning.

\section{\bench}

    \subsection{Data}

        Our main data sources are six leading surveys of U.S. households that offer rich self-reported information on households' socio-demographic background, their economic behavior, and their preferences and beliefs about the future. We choose these surveys because they are publicly available, extremely well-documented and widely used in social science research. Except for the Census, all of them enable longitudinal linking of households or respondents over several waves. Thus, \bench is able to track households over time and focus on predicting \textit{changes} in behavior, rather than on pure cross-sectional prediction.

        Each survey provides high-quality information about a few narrow topics. The \textbf{Consumer Expenditure Survey} (CEX) collects detailed information on household consumption expenditure. The \textbf{Current Population Survey} (CPS) contains detailed information on employment, unemployment, hours worked and labor earnings. The University of Michigan's \textbf{Surveys of Consumers} (Michigan) focus on households' expectations and attitudes about their own economic situation and about the U.S. economy as a whole. The New York Fed's \textbf{Survey of Consumer Expectations} (SCE) specializes in consumer beliefs about the economy, including measures of subjective uncertainty, and it also provides a rich set of special modules eliciting consumer preferences directly. The \textbf{Panel Study of Income Dynamics} (PSID) has followed a set of U.S. individuals and families continuously since 1968, providing detailed information on income, employment, wealth, and other aspects of economic behavior. Finally, \textbf{U.S. Decennial Census} extracts (Census) provide data on demographics and residential mobility. For a detailed discussion of each source, including exact provenance and data cleaning, see Appendix \ref{appendix:sources}.

    \subsection{Evaluation Tasks}

        We use micro-data from these surveys to build 32 separate evaluation tasks, grouped into five broad \textbf{topics}. They are (1) consumption and saving, (2) income dynamics and household resources, (3) labor supply, job search, and retirement, (4) macroeconomic expectations, and (5) housing and location. These topics cover important aspects of household behavior and expectations and correspond to different \textit{blocks} within a model of household choice. Individual tasks within the topics were designed to zoom in on particular margins or decisions. See Table \ref{tab:householdbench-tasks} for a summary of the full suite, and Appendix \ref{appendix:tasks} for detailed information on each task, including full prompt examples.

        \begin{table}[t]
\centering
\caption{The 32 \bench tasks by topic. Baseline tasks condition on the household's history and the macroeconomic environment. Policy-response tasks add a policy, shock, or scenario. Type: N numeric, C categorical, P probabilistic. Appendix~\ref{appendix:tasks} describes every task in full.}
\label{tab:householdbench-tasks}
\begingroup
\fontsize{7}{8.4}\selectfont
\renewcommand{\arraystretch}{1.0}
\setlength{\tabcolsep}{2pt}
\newcommand{\taskid}{\fontfamily{lmtt}\fontseries{lc}\fontsize{6.5}{8.4}\selectfont}
\begin{tabular}{@{}>{\taskid}p{1.72cm}>{\raggedright\arraybackslash}p{\dimexpr(\textwidth-4.52cm-26pt)/2\relax}>{\centering\arraybackslash}p{0.5cm}@{\hspace{10pt}}>{\taskid}p{1.80cm}>{\raggedright\arraybackslash}p{\dimexpr(\textwidth-4.52cm-26pt)/2\relax}>{\centering\arraybackslash}p{0.5cm}@{}}
\toprule
\multicolumn{3}{@{}l}{\textbf{Baseline tasks}} & \multicolumn{3}{l@{}}{\textbf{Policy-response tasks}} \\
\cmidrule(r{10pt}){1-3} \cmidrule(l{0pt}){4-6}
\normalfont\textbf{Task ID} & \textbf{Prediction target} & \textbf{Type} & \normalfont\textbf{Task ID} & \textbf{Prediction target} & \textbf{Type} \\
\midrule
\multicolumn{6}{@{}l}{\textbf{Consumption and saving}} \\
cons\_cex\_categories & Spending in 12 categories & N & cons\_cex\_rebate01 & Spending after the 2001 tax rebates & N \\
cons\_cex\_total & Total, nondurable, and durable spending & N & cons\_cex\_sspay & Daily spending around Social Security payments & N \\
cons\_psid\_wealth & Net worth in the next wave & N & cons\_cex\_stimulus08 & Spending after the 2008 stimulus payments & N \\
cons\_sce\_growth & Year-on-year growth in monthly spending & N & cons\_psid\_jobloss & Food spending the year after a job loss & N \\
 & &  & cons\_sce\_shock & Spending share of $\pm$10\% income changes & N \\
\addlinespace[2pt]
\multicolumn{6}{@{}l}{\textbf{Income dynamics and household resources}} \\
income\_mich\_finance & Expected 1-year family-income growth & N & income\_cps\_displace & Weekly earnings after job displacement & N \\
income\_psid\_earnings & Labor earnings 2, 4, and 10 years ahead & N & income\_sce\_policy & Stated effect of tax and benefit changes & C \\
income\_sce\_growth & Expected 1-year household-income growth & N &  & &  \\
\addlinespace[2pt]
\multicolumn{6}{@{}l}{\textbf{Labor supply, job search, and retirement}} \\
labor\_cps\_jobfind & Next-month status of the unemployed & C & labor\_cps\_displace & Labor-force status after job displacement & C \\
labor\_cps\_retire & Retirement within 12 months, aged 50--75 & C & labor\_cps\_ui & Next-month status under unemployment-insurance durations & C \\
labor\_cps\_separation & Next-month status of the employed & C & labor\_psid\_addedworker & Partner hours and earnings after spousal job loss & N \\
labor\_sce\_offer & Acceptance of the first listed job offer & C & labor\_sce\_reswage & Reservation wage and preferred hours & N \\
labor\_sce\_risk & Job-loss, quit, and reemployment chances & P &  & &  \\
labor\_sce\_search & 3- and 12-month job-finding probabilities & P &  & &  \\
\addlinespace[2pt]
\multicolumn{6}{@{}l}{\textbf{Macroeconomic expectations}} \\
macro\_mich\_outlook & Expected inflation at 1 and 5--10 years & N & macro\_sce\_revision & Expected inflation after an inflation surprise & N \\
macro\_sce\_uncertainty & Inflation and macro-outcome probabilities & P &  & &  \\
\addlinespace[2pt]
\multicolumn{6}{@{}l}{\textbf{Housing and location}} \\
house\_census\_move & Five-year residential-mobility status & C & house\_sce\_financing & Home price, down payment, 3 scenarios & N \\
house\_psid\_owner & Next-wave homeownership among renters & C & house\_sce\_lockin & Three-year moving probability with a portable mortgage rate & P \\
house\_sce\_move & One-year moving probability & P &  & &  \\
\bottomrule
\end{tabular}
\endgroup
\end{table}

        \textbf{Target Variables.} Each task features one or several targets (variables to be predicted), which generally are direct, untransformed survey answers.
        Targets can be \textit{numeric}, like total consumption expenditure in US Dollars, \textit{categorical}, like employment status in a month, or \textit{probabilities}, like beliefs about inflation over the next 12 months.

        \textbf{Baseline and Policy-Response Tasks.} We also classify the tasks into two groups based on the information provided in the prompt.
        \begin{itemize}[leftmargin=*,itemsep=4pt,topsep=2pt]
            \item \textit{Baseline tasks} measure whether models can predict household outcomes or beliefs using information already available about the household. The prompt gives socio-demographic background (sex, race, education, household composition, location), past economic behavior and outcomes, and the macroeconomic environment.\footnote{All tasks include a common set of core macroeconomic variables. All tasks include the last four quarters of GDP growth, CPI inflation, unemployment and the Federal Funds rate, plus their average values over the previous five years. Some tasks include additional variables if relevant.}
            \item \textit{Policy-response tasks} ask whether models can predict behavior or beliefs conditional on variation in the economic environment that is plausibly external to the household. That variation is a change in government policy, an observed economic shock, an information treatment, or a hypothetical scenario the interviewer provides. Policy-response tasks offer a sharper test of whether models understand how households adjust their behavior when the economic environment changes.
        \end{itemize}

        \textbf{Sample Construction.} We filter the raw survey entries using a careful protocol that is harmonised across datasets. For each task we drop records with a missing target or missing recent history, apply survey-specific quality filters, and trim outliers within each period. Each data point is one household in one period. The prompt gives that household's demographics, its own recent history, and current macroeconomic conditions, and policy-response tasks add the policy or shock. The answer is the household's actual survey response. Appendix \ref{appendix:tasks} gives the filters and a full prompt for each task.

        \textbf{Dataset Splits.} After constructing the samples for each task, we divide the dataset into a few splits. We first hold out every data point released after 16th February 2026 as a \textit{post-cutoff test set}. It postdates the knowledge cutoff of most LLMs we evaluate, so it gives valuable information about the contamination effect.
        Earlier samples are then grouped into calendar quarters, and we randomly split the quarters into training, validation, and pre-cutoff testing in an 80/10/10 ratio. Splitting on quarters means the test set contains macro conditions unseen in the training or validation set.
        From each test set, we evaluate at most 200 observations per task, sampled across calendar quarters in proportion to the number of eligible observations in each quarter.

    \subsection{Comparison with Existing Work}
        Table~\ref{tab:benchmark_comparison} compares \bench with related work evaluating LLM predictions of household economic outcomes and beliefs. Existing studies focus on specific topics at a time, and use at most four surveys. \bench is the only study covering a broad range of five topics (spending, income, jobs and retirement, macro expectations, housing) under a harmonised framework, while using as many as six survey datasets. This breadth is important for drawing robust and widely applicable conclusions about household behavior, which are not specific to a particular survey or outcome.

        Regarding policy response, there are some papers studying how households adjust to changing information or hypothetical policy treatments~\citep{anesti2025inflation,wu2025llmsurvey,park2025augmenting,lin2026simulating,liu2026mimic,zarifhonarvar2025generating}. However our policy-response tasks include not only changing information and hypotheticals, but also real-world episodes with policy changes and other related shocks, including: tax rebates and stimulus payments, unemployment-insurance variation, and household job loss. Asking whether LLMs can match actual household policy responses is critical---the answer tells us whether LLMs can be useful for realistic policy simulations.

        \providecommand{\cmark}{\ensuremath{\checkmark}}
\providecommand{\rothead}[1]{\rotatebox[origin=l]{90}{#1}}
\begin{table}[t]
\centering
\caption{Comparison between \bench and selected studies.}
\label{tab:benchmark_comparison}
\begingroup
\scriptsize
\setlength{\tabcolsep}{2.2pt}
\renewcommand{\arraystretch}{1.1}
\resizebox{\textwidth}{!}{%
\begin{tabular}{@{}l>{\RaggedRight\arraybackslash\hyphenpenalty=10000\exhyphenpenalty=10000}p{3.7cm}r|ccccc|c@{}}
\toprule
\textbf{Study} & \textbf{Data} & \textbf{Years} & \multicolumn{5}{c|}{\textbf{Outcomes predicted}} & \multicolumn{1}{c@{}}{\textbf{Design}} \\
\cmidrule(lr){4-8}\cmidrule(l){9-9}
 & &
 & \textbf{Spending}
 & \textbf{Income}
 & \textbf{Labor}
 & \textbf{Macro}
 & \textbf{Housing}
 & \textbf{Policy} \\
\midrule
\citet{athey2024laborllm} & US labor panels (PSID, NLSY) & 1979--2021 &  &  & \cmark &  &  &  \\
\citet{cruz2024evaluating} & American Community Survey (ACS) & 2018 &  & \cmark & \cmark &  & \cmark &  \\
\citet{brynjolfsson2025augmenting} & PSID, UK valuation surveys & 2017--2021 &  & \cmark &  &  &  &  \\
\citet{jia2026personas} & Dutch household panel (LISS) & 2023--2024 &  &  & \cmark &  & \cmark &  \\
\citet{wu2025llmsurvey} & US consumer surveys (SCE, Nielsen) & 2018--2023 &  &  &  & \cmark &  & \cmark \\
\citet{zarifhonarvar2025generating} & US consumer surveys (SCE, Nielsen) & 2019--2025 &  &  &  & \cmark &  & \cmark \\
\midrule
\textbf{\bench (ours)} & Six U.S. household surveys & \textbf{1968--2026} & \cmark & \cmark & \cmark & \cmark & \cmark & \cmark \\
\bottomrule
\end{tabular}}
\endgroup
\end{table}

\section{Experimental Setting}

\subsection{Model Suite}
\label{sec:models}

We benchmark two types of language models, open-weight LLMs and proprietary frontier LLMs. We also include a no-change baseline and XGBoost as statistical reference models.

\textbf{Open-Weight LLMs.} We include open-weight LLMs with various model sizes, including Qwen3.5-4B~\citep{qwen3_5}, Qwen3.6-27B and Qwen3.6-35B-A3B~\citep{qwen3_6}, DeepSeek-V4-Flash and DeepSeek-V4-Pro~\citep{deepseek-v4}. These span from small dense models to larger mixture-of-experts models.

\textbf{Proprietary LLMs.} We include two families of widely used frontier proprietary models. Within each family we include different capability tiers. For GPT, we include GPT-5.6 luna, GPT-5.6 terra, GPT-5.6 sol~\citep{gpt-5-6}, and GPT-6 astra~\citep{gpt-6-astra}. For Claude, we include Claude Opus 4.8~\citep{claude-opus-4-8}, Claude Sonnet 5~\citep{claude-sonnet-5}, Claude Opus 5~\citep{claude-opus-5}, and Fable 5.1~\citep{claude-fable-5-1}. We run every model under its default inference settings. Details are included in Appendix~\ref{appendix:inference}.

\textbf{Statistical and Reference Models.}
Two statistical models provide reference points for evaluating LLM performance.
\begin{itemize}[leftmargin=*,itemsep=0pt,topsep=2pt]
    \item \textit{No-change baseline.} This uses the household's most recent observed value as its prediction. For the policy-response tasks it predicts zero response to the policy or shock. This simple forecast is a standard benchmark in economics (e.g. \citealp{meese1983empirical}; \citealp{atkeson2001phillips}).
    \item \textit{XGBoost.} XGBoost is a gradient-boosted tree method and achieves good performance on tabular prediction tasks~\citep{holzmuller2024better}.
    XGBoost~\citep{chen2016xgboost} is trained, task-by-task, on the same household features the LLMs receive as text. We use the pre-tuned parameter settings of \citet{holzmuller2024better}, which outperform XGBoost's default parameters. Appendix~\ref{appendix:xgboost} gives the features and every parameter value.
 \end{itemize}

\subsection{Metrics}

We use the following metrics for the three task types.

\begin{itemize}[leftmargin=*,itemsep=4pt,topsep=2pt]
    \item \textit{Numeric targets} use relative mean absolute error (RelMAE), the model's absolute error over the absolute error of the no-change baseline on the same samples~\citep{hyndman2006another,hewamalage2023forecast}:
    \begin{equation*}
        \mathrm{RelMAE} = \frac{\sum_{i=1}^{n} \lvert \hat{y}_i - y_i \rvert}{\sum_{i=1}^{n} \lvert \tilde{y}_{i,\text{no-change}} - y_i \rvert}.
    \end{equation*}
    Here $y_i$ denotes the household's answer, $\hat{y}_i$ the model's prediction, and $\tilde{y}_{i,\text{no-change}}$ that of the no-change baseline of Section~\ref{sec:models}, over the $n$ samples of the task. A value below $1.0$ beats the no-change baseline. The tasks of the benchmark come on different numerical scales, and RelMAE is a natural way to normalize across them.
    \item \textit{Categorical targets} use macro-$F_1$, which computes an $F_1$ score for each outcome category and averages these scores with equal weight. This matters when some outcomes are much less common than others.
    \item \textit{Probability targets} use mean total variation (TV) distance between the predicted probability vector $\hat{p}_i$ and the household's own $p_i$ over the target's $B$ bins:
    \begin{equation*}
        \mathrm{TV} = \frac{1}{n} \sum_{i=1}^{n} \frac{1}{2} \sum_{b=1}^{B} \lvert \hat{p}_{ib} - p_{ib} \rvert.
    \end{equation*}
    It is $0$ when the two probability vectors agree perfectly, and $1$ when they put their mass on disjoint bins, with no overlap at all.
\end{itemize}

To obtain a unified score for each model that represents its performance on \bench, we report the \textbf{average rank} across all 32 tasks. Lower average ranks indicate better performance. Since we only evaluate models on relatively small fractions of our full testing samples, we assess sampling variation in all our final results and measures using a standard i.i.d. bootstrapping procedure (10,000 bootstrap samples for each task), paired across models.

\section{Results and Discussion}
\label{sec:results}

    \subsection{Main Results}
    \label{sec:results-main}

    \begin{table}[t]
\centering
\resizebox{\textwidth}{!}{%
\small
\begin{tabular}{rlccccc|c|ccc}
\toprule
 &  & \multicolumn{6}{c}{\textbf{Average rank within topic}} & \multicolumn{3}{c}{\textbf{Average score by target type}} \\
\cmidrule(lr){3-8} \cmidrule(lr){9-11}
\textbf{Rank} & \textbf{Model} & \textbf{Consumption} & \textbf{Housing} & \textbf{Income} & \textbf{Labor} & \textbf{Macroeconomic} & \textbf{Average rank} & \textbf{RelMAE $\downarrow$} & \textbf{Macro-F1 $\uparrow$} & \textbf{TV $\downarrow$} \\
 &  & \scriptsize (\# = 9) & \scriptsize (\# = 5) & \scriptsize (\# = 5) & \scriptsize (\# = 10) & \scriptsize (\# = 3) &  & \scriptsize (\# = 18) & \scriptsize (\# = 9) & \scriptsize (\# = 5) \\
\midrule
1 & XGBoost & \shortstack{2.3 \\ {\scriptsize (0.6)}} & \shortstack{5.8 \\ {\scriptsize (1.0)}} & \shortstack{2.6 \\ {\scriptsize (0.8)}} & \shortstack{4.4 \\ {\scriptsize (0.7)}} & \shortstack{2.0 \\ {\scriptsize (0.8)}} & \shortstack{3.4 \\ {\scriptsize (0.4)}} & \shortstack{0.811 \\ {\scriptsize (0.011)}} & \shortstack{0.432 \\ {\scriptsize (0.014)}} & \shortstack{0.141 \\ {\scriptsize (0.004)}} \\
\hdashline[1pt/2pt]
2 & Fable 5.1 & \shortstack{6.0 \\ {\scriptsize (0.5)}} & \shortstack{\textbf{4.4} \\ {\scriptsize (0.8)}} & \shortstack{\textbf{2.4} \\ {\scriptsize (0.9)}} & \shortstack{\textbf{4.2} \\ {\scriptsize (0.6)}} & \shortstack{\textbf{2.0} \\ {\scriptsize (0.9)}} & \shortstack{\textbf{3.8} \\ {\scriptsize (0.3)}} & \shortstack{\textbf{0.878} \\ {\scriptsize (0.013)}} & \shortstack{\textbf{0.413} \\ {\scriptsize (0.013)}} & \shortstack{\textbf{0.139} \\ {\scriptsize (0.005)}} \\
3 & GPT-5.6 sol & \shortstack{5.3 \\ {\scriptsize (0.5)}} & \shortstack{\underline{5.2} \\ {\scriptsize (0.7)}} & \shortstack{6.2 \\ {\scriptsize (1.0)}} & \shortstack{\underline{4.8} \\ {\scriptsize (0.5)}} & \shortstack{6.0 \\ {\scriptsize (0.7)}} & \shortstack{\underline{5.5} \\ {\scriptsize (0.3)}} & \shortstack{0.910 \\ {\scriptsize (0.013)}} & \shortstack{0.399 \\ {\scriptsize (0.011)}} & \shortstack{\textbf{0.139} \\ {\scriptsize (0.005)}} \\
4 & GPT-6 astra & \shortstack{\textbf{4.4} \\ {\scriptsize (0.4)}} & \shortstack{8.6 \\ {\scriptsize (0.7)}} & \shortstack{\underline{4.6} \\ {\scriptsize (0.7)}} & \shortstack{6.4 \\ {\scriptsize (0.4)}} & \shortstack{\underline{5.0} \\ {\scriptsize (0.8)}} & \shortstack{5.8 \\ {\scriptsize (0.3)}} & \shortstack{0.915 \\ {\scriptsize (0.014)}} & \shortstack{0.369 \\ {\scriptsize (0.006)}} & \shortstack{\textbf{0.139} \\ {\scriptsize (0.005)}} \\
5 & Claude Opus 5 & \shortstack{\underline{4.7} \\ {\scriptsize (0.4)}} & \shortstack{7.2 \\ {\scriptsize (0.7)}} & \shortstack{5.4 \\ {\scriptsize (1.0)}} & \shortstack{5.2 \\ {\scriptsize (0.6)}} & \shortstack{8.7 \\ {\scriptsize (1.0)}} & \shortstack{6.2 \\ {\scriptsize (0.3)}} & \shortstack{\underline{0.905} \\ {\scriptsize (0.014)}} & \shortstack{\underline{0.406} \\ {\scriptsize (0.013)}} & \shortstack{0.144 \\ {\scriptsize (0.005)}} \\
6 & Claude Opus 4.8 & \shortstack{5.4 \\ {\scriptsize (0.5)}} & \shortstack{6.4 \\ {\scriptsize (0.8)}} & \shortstack{6.8 \\ {\scriptsize (0.8)}} & \shortstack{5.4 \\ {\scriptsize (0.5)}} & \shortstack{8.7 \\ {\scriptsize (0.9)}} & \shortstack{6.5 \\ {\scriptsize (0.3)}} & \shortstack{0.919 \\ {\scriptsize (0.014)}} & \shortstack{0.402 \\ {\scriptsize (0.009)}} & \shortstack{0.146 \\ {\scriptsize (0.005)}} \\
7 & GPT-5.6 terra & \shortstack{7.2 \\ {\scriptsize (0.6)}} & \shortstack{7.4 \\ {\scriptsize (0.8)}} & \shortstack{8.4 \\ {\scriptsize (0.9)}} & \shortstack{6.9 \\ {\scriptsize (0.5)}} & \shortstack{6.3 \\ {\scriptsize (0.8)}} & \shortstack{7.3 \\ {\scriptsize (0.3)}} & \shortstack{0.950 \\ {\scriptsize (0.013)}} & \shortstack{0.385 \\ {\scriptsize (0.011)}} & \shortstack{0.147 \\ {\scriptsize (0.005)}} \\
8 & Claude Sonnet 5 & \shortstack{8.3 \\ {\scriptsize (0.6)}} & \shortstack{7.4 \\ {\scriptsize (0.9)}} & \shortstack{6.8 \\ {\scriptsize (0.8)}} & \shortstack{8.8 \\ {\scriptsize (0.5)}} & \shortstack{9.0 \\ {\scriptsize (0.9)}} & \shortstack{8.1 \\ {\scriptsize (0.3)}} & \shortstack{0.934 \\ {\scriptsize (0.014)}} & \shortstack{0.361 \\ {\scriptsize (0.008)}} & \shortstack{0.148 \\ {\scriptsize (0.005)}} \\
9 & Qwen3.6-27B & \shortstack{10.4 \\ {\scriptsize (0.5)}} & \shortstack{6.6 \\ {\scriptsize (1.0)}} & \shortstack{8.8 \\ {\scriptsize (0.9)}} & \shortstack{6.6 \\ {\scriptsize (0.5)}} & \shortstack{9.3 \\ {\scriptsize (1.0)}} & \shortstack{8.4 \\ {\scriptsize (0.4)}} & \shortstack{1.007 \\ {\scriptsize (0.019)}} & \shortstack{0.385 \\ {\scriptsize (0.011)}} & \shortstack{0.148 \\ {\scriptsize (0.005)}} \\
10 & GPT-5.6 luna & \shortstack{9.0 \\ {\scriptsize (0.6)}} & \shortstack{5.8 \\ {\scriptsize (1.0)}} & \shortstack{10.6 \\ {\scriptsize (0.8)}} & \shortstack{7.8 \\ {\scriptsize (0.6)}} & \shortstack{9.0 \\ {\scriptsize (0.7)}} & \shortstack{8.4 \\ {\scriptsize (0.3)}} & \shortstack{1.001 \\ {\scriptsize (0.018)}} & \shortstack{0.381 \\ {\scriptsize (0.012)}} & \shortstack{\underline{0.143} \\ {\scriptsize (0.005)}} \\
11 & DeepSeek-V4-Pro & \shortstack{9.6 \\ {\scriptsize (0.7)}} & \shortstack{10.8 \\ {\scriptsize (1.1)}} & \shortstack{7.6 \\ {\scriptsize (0.9)}} & \shortstack{8.7 \\ {\scriptsize (0.6)}} & \shortstack{7.3 \\ {\scriptsize (0.8)}} & \shortstack{8.8 \\ {\scriptsize (0.4)}} & \shortstack{1.002 \\ {\scriptsize (0.017)}} & \shortstack{0.375 \\ {\scriptsize (0.011)}} & \shortstack{0.151 \\ {\scriptsize (0.005)}} \\
12 & DeepSeek-V4-Flash & \shortstack{10.4 \\ {\scriptsize (0.6)}} & \shortstack{11.8 \\ {\scriptsize (0.9)}} & \shortstack{10.0 \\ {\scriptsize (0.9)}} & \shortstack{7.8 \\ {\scriptsize (0.5)}} & \shortstack{8.7 \\ {\scriptsize (1.0)}} & \shortstack{9.7 \\ {\scriptsize (0.4)}} & \shortstack{0.995 \\ {\scriptsize (0.019)}} & \shortstack{0.363 \\ {\scriptsize (0.010)}} & \shortstack{0.151 \\ {\scriptsize (0.005)}} \\
13 & No-change baseline & \shortstack{11.4 \\ {\scriptsize (0.5)}} & \shortstack{6.8 \\ {\scriptsize (0.8)}} & \shortstack{12.8 \\ {\scriptsize (0.5)}} & \shortstack{7.9 \\ {\scriptsize (0.4)}} & \shortstack{10.3 \\ {\scriptsize (0.4)}} & \shortstack{9.9 \\ {\scriptsize (0.2)}} & \shortstack{1.000 \\ {\scriptsize (0.000)}} & \shortstack{0.315 \\ {\scriptsize (0.003)}} & \shortstack{0.139 \\ {\scriptsize (0.005)}} \\
14 & Qwen3.6-35B-A3B & \shortstack{10.4 \\ {\scriptsize (0.5)}} & \shortstack{11.2 \\ {\scriptsize (1.0)}} & \shortstack{12.2 \\ {\scriptsize (0.8)}} & \shortstack{9.1 \\ {\scriptsize (0.6)}} & \shortstack{13.0 \\ {\scriptsize (0.6)}} & \shortstack{11.2 \\ {\scriptsize (0.3)}} & \shortstack{0.995 \\ {\scriptsize (0.016)}} & \shortstack{0.365 \\ {\scriptsize (0.011)}} & \shortstack{0.174 \\ {\scriptsize (0.006)}} \\
15 & Qwen3.5-4B & \shortstack{14.9 \\ {\scriptsize (0.3)}} & \shortstack{12.2 \\ {\scriptsize (1.0)}} & \shortstack{14.8 \\ {\scriptsize (0.1)}} & \shortstack{10.7 \\ {\scriptsize (0.5)}} & \shortstack{14.7 \\ {\scriptsize (0.2)}} & \shortstack{13.5 \\ {\scriptsize (0.2)}} & \shortstack{1.304 \\ {\scriptsize (0.033)}} & \shortstack{0.346 \\ {\scriptsize (0.012)}} & \shortstack{0.229 \\ {\scriptsize (0.008)}} \\
\bottomrule
\end{tabular}%
}
\caption{Average rank by topic and average score by target type, pre-cutoff test split. Best LLM in bold, second best underlined. Ranks are taken over the models shown. Brackets report bootstrap SEs.}
\label{tab:leaderboard_main}
\end{table}

    \textbf{Most LLMs outperform the no-change baseline on numerical and categorical prediction, but not probability prediction.} Table~\ref{tab:leaderboard_main} reports the pre-cutoff leaderboard, where we rank the models by their average rank over the five topics. Most LLMs are better than the no-change baseline, which ranks 13th.
    Among the LLMs, Fable 5.1 achieves the strongest overall LLM performance, with a RelMAE of 0.878 and a macro-F1 of 0.413.
    The proprietary models also rank above the open-weight models: the best open-weight model (Qwen3.6-27B) only ranks 9th.

    For probability tasks, no model improves on the no-change baseline.
    This is due to models underestimating the persistence of beliefs. On the four baseline probability tasks, 26.9\% of target probability vectors exactly repeat the previous household report. Appendix Figure \ref{fig:appendix-probability-persistence} separates performance on changed and unchanged reports. For GPT-5.6 sol and Claude Opus 5, gains on changed reports are more than offset by errors on unchanged reports.

\begin{figure}[!t]
        \centering
        \includegraphics[width=\linewidth]{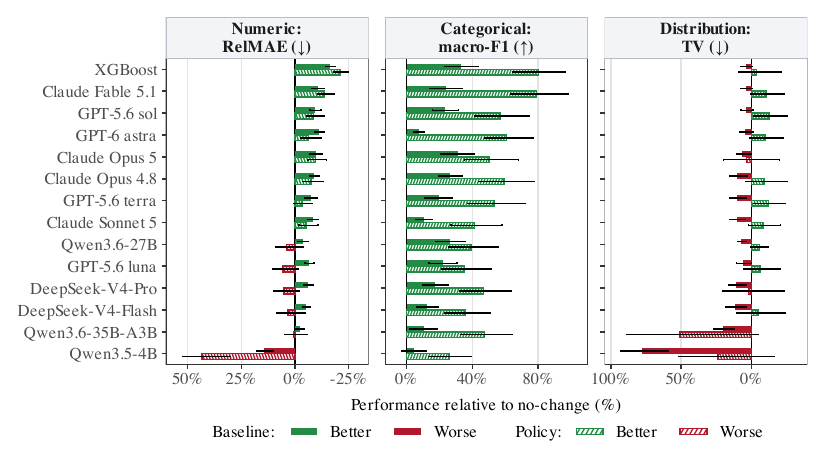}
        \caption{Model performance relative to the no-change benchmark on baseline and policy-response tasks. Solid bars report mean scores for baseline tasks, while hatched bars report mean scores for policy-response tasks. Whiskers are pointwise 95\% basic IID bootstrap intervals.
        }
        \label{fig:baseline-vs-policy}
    \end{figure}

    \textbf{Leading LLMs also improve over the no-change prediction on policy-response tasks.} Figure \ref{fig:baseline-vs-policy} considers performance improvements over the no-change model separately for baseline and policy-response tasks. Fable 5.1, Claude Opus 5, and GPT-5.6 sol reduce numeric error by 13.6\%, 9.5\%, and 8.7\% on policy-response tasks, even without task-specific training. Their baseline-task gains are 10.5\%, 9.6\%, and 9.4\%, respectively. These results illustrate the flexibility of LLMs in predicting household responses to policy changes.

    \textbf{Widespread across tasks, LLMs do not outperform XGBoost models trained per task, but Fable 5.1 is close; other proprietary models perform well, and open-weight models lag.} XGBoost ranks first on the leaderboard with an average rank of 3.4, against 3.8 for Fable 5.1, and their bootstrap intervals overlap. The two are closest on the categorical and probability tasks, where their intervals overlap: 0.413 against 0.432 in macro-F1, and 0.139 against 0.141 in TV, with Fable 5.1 marginally ahead. XGBoost keeps its lead on the numeric tasks, 0.811 against 0.878.
    This suggests that the strongest proprietary LLM, without any task-specific training, can predict household behavior nearly as accurately as an XGBoost model trained on each task.
    Open-weight models perform less well, occupying the bottom ranks of the leaderboard. These results are widespread across tasks.

    \textbf{LLMs' performance on \bench does not drop after or close to their knowledge cutoff.} Panels (a) and (b) of Appendix Figure \ref{fig:appendix-post-minus-pre-cutoff-all-models} compare model performance on the matched pre-cutoff and post-cutoff task samples. Across all tasks, we do not see large systematic changes in performance for tasks with numerical targets, suggesting limited look-ahead bias from pre-training. This may reflect the lower likelihood that pre-training data contain individual household outcomes.
    Appendix Figure \ref{fig:appendix-performance-over-time-all-models} performs an additional analysis, looking at average performance over time for a subset of tasks where we have a long sample period. Again, we do not see any systematic degradation of performance as we approach knowledge cutoffs either.

    \subsection{LLMs' bias on predicting household behavior}
    \label{sec:results-failure-modes}

    Although leading frontier LLMs predict household behavior well overall, they exhibit systematic biases.

\begin{figure}[t]
    \centering
    \begin{minipage}[t]{0.4695\textwidth}
    \vspace{0pt}
    \centering
    \includegraphics[width=\linewidth]{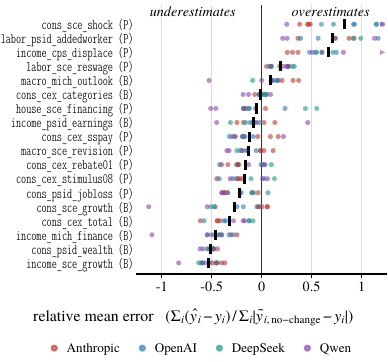}
    \end{minipage}\hfill
    \begin{minipage}[t]{0.5184\textwidth}
    \vspace{0pt}
    \centering
    \scriptsize
\setlength{\tabcolsep}{2pt}
\renewcommand{\arraystretch}{1.15}
\newcommand{\taskid}{\fontfamily{lmtt}\fontseries{lc}\selectfont}
\begin{tabular}{@{}l r;{1pt/2pt}rrrr@{}}
\toprule
\shortstack[l]{\bfseries Task\\[-1.2pt](majority class)} & \multicolumn{1}{c}{\shortstack[c]{\bfseries Ground\\[-1.2pt]truth}} & \multicolumn{1}{c}{\shortstack[c]{\bfseries Claude\\[-1.2pt]Fable 5.1}} & \multicolumn{1}{c}{\shortstack[c]{\bfseries GPT-6\\[-1.2pt]astra}} & \multicolumn{1}{c}{\shortstack[c]{\bfseries GPT-5.6\\[-1.2pt]sol}} & \multicolumn{1}{c}{\shortstack[c]{\bfseries XGBoost}} \\
\midrule
\shortstack[l]{{\taskid labor\_cps\_retire (B)}\\[-1.2pt]({\itshape not retired})} & \shortstack[r]{96\%\\[-1.2pt]{}} & \shortstack[r]{\textbf{100\%}\\[-1.2pt]{}} & \shortstack[r]{\textbf{100\%}\\[-1.2pt]{}} & \shortstack[r]{96\%\\[-1.2pt]{}} & \shortstack[r]{\textbf{98\%}\\[-1.2pt]{}} \\
\hdashline[1pt/2pt]
\shortstack[l]{{\taskid labor\_cps\_separation (B)}\\[-1.2pt]({\itshape still employed})} & \shortstack[r]{94\%\\[-1.2pt]{}} & \shortstack[r]{\textbf{100\%}\\[-1.2pt]{}} & \shortstack[r]{\textbf{100\%}\\[-1.2pt]{}} & \shortstack[r]{\textbf{100\%}\\[-1.2pt]{}} & \shortstack[r]{\textbf{100\%}\\[-1.2pt]{}} \\
\hdashline[1pt/2pt]
\shortstack[l]{{\taskid house\_psid\_owner (B)}\\[-1.2pt]({\itshape still renting})} & \shortstack[r]{89\%\\[-1.2pt]{}} & \shortstack[r]{\textbf{98\%}\\[-1.2pt]{}} & \shortstack[r]{\textbf{98\%}\\[-1.2pt]{}} & \shortstack[r]{\textbf{96\%}\\[-1.2pt]{}} & \shortstack[r]{\textbf{97\%}\\[-1.2pt]{}} \\
\hdashline[1pt/2pt]
\shortstack[l]{{\taskid labor\_sce\_offer (B)}\\[-1.2pt]({\itshape rejected the offer})} & \shortstack[r]{69\%\\[-1.2pt]{}} & \shortstack[r]{\textbf{76\%}\\[-1.2pt]{}} & \shortstack[r]{\textbf{76\%}\\[-1.2pt]{}} & \shortstack[r]{\textbf{72\%}\\[-1.2pt]{}} & \shortstack[r]{\textbf{80\%}\\[-1.2pt]{}} \\
\hdashline[1pt/2pt]
\shortstack[l]{{\taskid labor\_cps\_ui (P)}\\[-1.2pt]({\itshape still unemployed})} & \shortstack[r]{69\%\\[-1.2pt]{}} & \shortstack[r]{\textbf{94\%}\\[-1.2pt]{}} & \shortstack[r]{\textbf{100\%}\\[-1.2pt]{}} & \shortstack[r]{\textbf{100\%}\\[-1.2pt]{}} & \shortstack[r]{\textbf{84\%}\\[-1.2pt]{}} \\
\hdashline[1pt/2pt]
\shortstack[l]{{\taskid house\_census\_move (B)}\\[-1.2pt]({\itshape did not move})} & \shortstack[r]{67\%\\[-1.2pt]{}} & \shortstack[r]{\textbf{83\%}\\[-1.2pt]{}} & \shortstack[r]{\textbf{100\%}\\[-1.2pt]{}} & \shortstack[r]{\textbf{88\%}\\[-1.2pt]{}} & \shortstack[r]{\textbf{94\%}\\[-1.2pt]{}} \\
\hdashline[1pt/2pt]
\shortstack[l]{{\taskid labor\_cps\_displace (P)}\\[-1.2pt]({\itshape re-employed})} & \shortstack[r]{62\%\\[-1.2pt]{}} & \shortstack[r]{\textbf{80\%}\\[-1.2pt]{}} & \shortstack[r]{62\%\\[-1.2pt]{}} & \shortstack[r]{22\%\\[-1.2pt]{}} & \shortstack[r]{\textbf{87\%}\\[-1.2pt]{}} \\
\hdashline[1pt/2pt]
\shortstack[l]{{\taskid income\_sce\_policy (P)}\\[-1.2pt]({\itshape average of 4 targets})} & \shortstack[r]{55\%\\[-1.2pt]{}} & \shortstack[r]{\textbf{57\%}\\[-1.2pt]{}} & \shortstack[r]{\textbf{64\%}\\[-1.2pt]{}} & \shortstack[r]{52\%\\[-1.2pt]{}} & \shortstack[r]{\textbf{72\%}\\[-1.2pt]{}} \\
\hdashline[1pt/2pt]
\shortstack[l]{{\taskid labor\_cps\_jobfind (B)}\\[-1.2pt]({\itshape still unemployed})} & \shortstack[r]{50\%\\[-1.2pt]{}} & \shortstack[r]{\textbf{94\%}\\[-1.2pt]{}} & \shortstack[r]{\textbf{100\%}\\[-1.2pt]{}} & \shortstack[r]{\textbf{96\%}\\[-1.2pt]{}} & \shortstack[r]{\textbf{76\%}\\[-1.2pt]{}} \\
\bottomrule
\end{tabular}

    \par\vspace{3pt}
    {\footnotesize Share of majority-class answers.}
    \end{minipage}
    \caption{LLMs exhibit systematic bias in predicting household behavior. \emph{Left.} LLMs consistently overestimate household behavior on some tasks and underestimate it on others. \emph{Right.} LLMs overpredict the majority class on categorical tasks. Bold means that a model predicts the majority class more often than the ground truth. (B) and (P) denote baseline and policy-response tasks.}
    \label{fig:failure-modes}
    \end{figure}

    \textbf{LLMs systematically overestimate household outcomes on some tasks and underestimate them on others.} Figure~\ref{fig:failure-modes} (left) reports each LLM's relative mean error on the 18 numeric tasks, which is RelMAE with the signed error $\hat{y}_i - y_i$ in the numerator. Overestimation is largest on policy-response tasks, where the LLMs predict stronger responses than households report. For example, \texttt{cons\_sce\_shock} asks what share of a hypothetical permanent 10\% income gain the household would spend or donate. The median household answers 5\%, while the three frontier LLMs (Fable 5.1, GPT-5.6 sol, and GPT-6 astra) predict 30\% to 40\%.

    \textbf{On categorical prediction tasks, LLMs predict the most common outcome of a household decision more often than households choose it.} Figure~\ref{fig:failure-modes} (right) shows that the three frontier LLMs over-predict the majority class, usually the status quo. The gap is largest on \texttt{labor\_cps\_jobfind} and \texttt{labor\_cps\_ui}, which predict the next-month labor-force status of unemployed workers. On \texttt{labor\_cps\_jobfind}, only half of the workers remain unemployed, but the LLMs predict this for 94\% to 100\% of them. The LLMs also miss every worker who leaves the labor force on both tasks. Among the three models, GPT-6 astra shows this pattern most strongly: it predicts the majority class for every data point on 5 of the 9 categorical tasks.
    These biases suggest that more progress is needed before LLMs are fully reliable as simulations of household behavior.

\section{Improving LLMs' Predictive Power}
\label{sec:improving}

We have found that open-weight models lag proprietary models for predicting household behavior. This section presents a method that enables a small open-weight model to match leading proprietary LLMs. In particular, we show that combining supervised fine-tuning on household behavior data with aggregation over draws can close the gap.

    \subsection{Fine-Tuning Setup}
    \label{sec:ft}

    To test the hypothesis that fine-tuning an LLM on household training data could improve its performance on \bench, we use supervised fine-tuning (SFT), which is commonly used to adapt a general-purpose language model to a specific domain~\citep{FLAN, li2025scilitllm}.
    We fine-tune models on the training split of the 18 baseline tasks. For each task, we sample at most 6,000 data points from its training split, stratified by quarters. The training set contains 100,702 data points.
    We fine-tune two widely-used open-weight LLMs, Qwen3.5-4B and Qwen3.6-27B, with LoRA~\citep{hu2022lora} for one epoch.

    We do not fine tune on the policy response tasks, and instead treat them as a hold out sample, which allows us to test whether fine-tuning on baseline tasks generalizes to policy-response tasks. This step is important because data for policy response tasks is often sparse. Moreover often policymakers contemplate new policies, for which there is no existing data---say, a new form of unemployment insurance. One would like the LLM to make good predictions even for these kinds of policies.

    \subsection{Fine-Tuning Results}
    \label{sec:sft}

    \begingroup
        \let\HouseholdBenchTable\table
        \let\endHouseholdBenchTable\endtable
        \renewenvironment{table}[1][]{\HouseholdBenchTable[t]}{\endHouseholdBenchTable}
        \providecommand{\hbbetter}[1]{{\setlength{\fboxsep}{1.5pt}\colorbox{green!20}{#1}}}
\providecommand{\hbworse}[1]{{\setlength{\fboxsep}{1.5pt}\colorbox{red!20}{#1}}}
\begin{table}[t]
\centering
\resizebox{\textwidth}{!}{%
\small
\begin{tabular}{llcccc|ccc}
\toprule
 &  &  & \multicolumn{3}{c}{\textbf{Baseline tasks} \scriptsize (seen in fine-tuning)} & \multicolumn{3}{c}{\textbf{Policy-response tasks} \scriptsize (held out)} \\
\cmidrule(lr){4-6} \cmidrule(lr){7-9}
 & \textbf{Model} & \textbf{$K$} & \textbf{RelMAE} $\downarrow$ & \textbf{Macro-F1} $\uparrow$ & \textbf{TV} $\downarrow$ & \textbf{RelMAE} $\downarrow$ & \textbf{Macro-F1} $\uparrow$ & \textbf{TV} $\downarrow$ \\
 &  &  & \scriptsize (\# = 8) & \scriptsize (\# = 6) & \scriptsize (\# = 4) & \scriptsize (\# = 10) & \scriptsize (\# = 3) & \scriptsize (\# = 1) \\
\midrule
\multirow{4}{*}{\scriptsize Reference} & XGBoost & 1 & 0.84 & 0.46 & 0.15 & 0.79 & 0.36 & 0.12 \\
\hdashline[1pt/2pt]
 & GPT-6 astra & 1 & \underline{0.89} & 0.39 & \textbf{0.15} & 0.94 & 0.32 & \textbf{0.11} \\
 & Claude Opus 5 & 1 & 0.90 & \textbf{0.47} & \textbf{0.15} & \underline{0.91} & 0.28 & 0.13 \\
 & Fable 5.1 & 1 & \underline{0.89} & 0.44 & \textbf{0.15} & \textbf{0.86} & \textbf{0.35} & \textbf{0.11} \\
\midrule
\multirow{4}{*}{\scriptsize SFT} & Qwen3.5-4B & 1 & 1.14 & 0.39 & 0.25 & 1.43 & 0.25 & 0.15 \\
 & \quad + SFT & 1 & 1.00\,\hbbetter{\scriptsize ($-$0.14)} & 0.45\,\hbbetter{\scriptsize ($+$0.06)} & \underline{0.16}\,\hbbetter{\scriptsize ($-$0.09)} & 1.30\,\hbbetter{\scriptsize ($-$0.14)} & \underline{0.33}\,\hbbetter{\scriptsize ($+$0.08)} & 0.17\,\hbworse{\scriptsize ($+$0.01)} \\
 & Qwen3.6-27B & 1 & 0.96 & 0.45 & \underline{0.16} & 1.04 & 0.26 & \underline{0.12} \\
 & \quad + SFT & 1 & 1.00\,\hbworse{\scriptsize ($+$0.03)} & \underline{0.46}\,\hbbetter{\scriptsize ($+$0.01)} & \underline{0.16}\,\hbworse{\scriptsize ($+$0.01)} & 1.18\,\hbworse{\scriptsize ($+$0.14)} & 0.30\,\hbbetter{\scriptsize ($+$0.04)} & 0.15\,\hbworse{\scriptsize ($+$0.04)} \\
\midrule
\multirow{4}{*}{\scriptsize \shortstack[l]{SFT\\+ aggregation}} & Qwen3.5-4B & 16 & 1.11 & 0.38 & 0.24 & 1.31 & 0.24 & 0.15 \\
 & \quad + SFT & 16 & \textbf{0.87}\,\hbbetter{\scriptsize ($-$0.24)} & 0.42\,\hbbetter{\scriptsize ($+$0.05)} & \textbf{0.15}\,\hbbetter{\scriptsize ($-$0.09)} & 1.13\,\hbbetter{\scriptsize ($-$0.18)} & 0.31\,\hbbetter{\scriptsize ($+$0.06)} & 0.16\,\hbworse{\scriptsize ($+$0.01)} \\
 & Qwen3.6-27B & 16 & 0.95 & 0.43 & \textbf{0.15} & 1.01 & 0.25 & \underline{0.12} \\
 & \quad + SFT & 16 & \textbf{0.87}\,\hbbetter{\scriptsize ($-$0.08)} & 0.44\,\hbbetter{\scriptsize ($+$0.01)} & \textbf{0.15}\,\hbbetter{\scriptsize ($-$0.01)} & 1.03\,\hbworse{\scriptsize ($+$0.02)} & 0.27\,\hbbetter{\scriptsize ($+$0.02)} & 0.14\,\hbworse{\scriptsize ($+$0.02)} \\
\bottomrule
\end{tabular}%
}
\caption{Results of SFT. The $K$ column is the number of aggregated predictions. Best among the language models in bold, second best underlined. XGBoost is refitted on each task it is scored on, so its policy-response scores are not held out.}
\label{tab:sft_scores}
\end{table}

        \endgroup

    We fine-tune the two models with the data and configuration described above. We also aggregate $K=16$ draws per model by averaging numerical and probability predictions and taking majority votes for categorical predictions. Table~\ref{tab:sft_scores} shows the main results.

    First, SFT improves Qwen3.5-4B on all three types of prediction tasks, and the gains generalize to the unseen policy-response tasks. On the baseline tasks, SFT improves Qwen3.5-4B's RelMAE by 13\%, macro-$F_1$ by 15\%, and TV by 34\%. The gains generalize to the unseen policy-response tasks, where RelMAE improves by 10\% and macro-$F_1$ by 30\%. Qwen3.6-27B's results are more mixed: SFT improves its macro-$F_1$ but worsens its RelMAE and TV. This suggests that a model that already performs well on \bench may not always benefit from additional fine-tuning.

    Second, \textbf{the fine-tuned models with prediction aggregation can beat the best LLMs on numeric tasks.}
    We find that the fine-tuned models perform better with prediction aggregation.
    On the numeric baseline tasks, aggregating $K=16$ draws lowers the backbone models' RelMAE by only 3\% (Qwen3.5-4B) and 1\% (Qwen3.6-27B), but the fine-tuned models' by 10\% to 13\%. With aggregation, the fine-tuned Qwen3.5-4B reaches the best performance on numerical and probability prediction tasks on the baseline tasks, outperforming Fable 5.1. We also note that aggregating for proprietary models does not improve their performance (Appendix~\ref{appendix:two_draws}).
    In addition to SFT, we explore fine-tuning with a number-token loss~\citep{zausinger2025regress} and report the results in Appendix~\ref{appendix:ntl}.

\section{Conclusion}

We introduce HouseholdBench, a benchmark that evaluates LLM predictions of household outcomes and beliefs across five areas of economic behavior.
Our evaluation shows that LLMs still do not outperform XGBoost overall, although Fable 5.1 comes close. With fine-tuning and aggregation of 16 predictions, small open-weight models outperform leading frontier LLMs evaluated on numeric baseline tasks.
By testing these capabilities against household survey data, we provide a comprehensive way to assess LLMs as predictors of household behavior.

\bibliographystyle{iclr2027_conference}
\bibliography{references}

\clearpage
\appendix

\counterwithin{figure}{section}
\counterwithin{table}{section}
\renewcommand{\thefigure}{\thesection\arabic{figure}}
\renewcommand{\thetable}{\thesection\arabic{table}}

\section{Related Work}
\label{appendix:related_work}

In the main text and Table~\ref{tab:benchmark_comparison}, we focus on a few representative studies from the rapidly growing literature that evaluates or fine-tunes LLMs using household survey microdata, but there are others that deserve discussion. Within the topics covered by \bench, further evaluations use PSID data to predict wages and homeownership \citep{gao2026howwell}, SOEP data to reconstruct held-out survey answers \citep{kinzinger2026synthetic}, and SHARE data to predict retirement-related attitudes \citep{garzon2026demographics}. Other studies fine-tune LLMs to predict pension enrollment in Chinese household surveys \citep{li2026pension}, or combine individual model adaptation with longitudinal life histories to predict survey responses \citep{wang2026lifetrajectories}. The expectations literature includes comparisons with UK inflation beliefs \citep{anesti2025inflation}, Italian household expectations and responses to information \citep{park2025augmenting}, and human survey responses to hypothetical macroeconomic shocks and housing information \citep{lin2026simulating}. \citet{liu2026mimic} compare LLM spending expectations and job-acceptance intentions with Survey of Consumer Expectations responses before using the agents in an economic simulation. Related survey-based evaluations also examine decisions outside our specific tasks, including investment perceptions \citep{fedyk2024investment}, product choices and willingness to pay \citep{brand2026market}, and electric-vehicle preferences and responses to information \citep{namikoshi2024targeted}. These studies provide valuable evidence within particular economic domains, but, as we argued in the main text, \bench provides a more systematic and comprehensive evaluation of LLMs across a broad set of economic outcomes and settings. Also, policy-response evaluations in the cited studies are limited to stated responses to information or hypothetical scenarios, while \bench additionally evaluates predictions against realized household outcomes after real economic events.

\bench also sits within a broader literature simulating human behavior with LLMs. LLMs can simulate results from social science experiments~\citep{horton2023large,ashokkumar2026predict} and approximate the survey response distributions of demographic subgroups when conditioned on respondent demographics~\citep{argyle2023out}. Yet these simulations remain inaccurate: LLM opinions stay misaligned with those of demographic groups even after steering~\citep{opinionqa}, synthetic responses vary less than real ones~\citep{bisbee2024synthetic}, and predictions for specific individuals correlate weakly with their actual responses~\citep{peng2026digital}. Fine-tuning on human behavioral data can narrow this gap. For example, \citet{centaur} fine-tune an LLM on choices from 160 psychology experiments and outperform traditional cognitive models. \citet{socsci210} fine-tune on responses from social science experiments, \citet{BeFM-1} on economic games and personality surveys, and \citet{subpop} on public opinion surveys to predict response distributions across demographic subgroups. \bench extends this broader literature to fine-tune LLMs with household economic decisions.

\section{Data Sources}\label{appendix:sources}

This appendix describes each survey and the preparation of the common data files used to construct \bench tasks.

\subsection{Consumer Expenditure Survey: Diary Survey}\label{cex-diary}

\paragraph{Additional Information.} The U.S. Bureau of Labor Statistics designs and publishes the Consumer Expenditure Surveys, which the U.S. Census Bureau administers. The current continuous program began in 1980. In the Diary Survey, each sampled consumer unit records all expenditures for two consecutive weeks. The survey focuses on frequent purchases and also records demographic characteristics, income, and benefit receipt. A recent annual sample selects roughly 18,000 addresses and yields about 6,700 usable pairs of diaries \citep{bls2022cexhandbook,census2026cex}. BLS publishes the public-use files annually. Exact historical posting dates are unavailable for the years used here. We assign each release to March 31 of the following year.

\paragraph{Downloading and Cleaning the Data.} We download the 1982--1989 fixed-width files from ICPSR and the 1990--2011 comma-delimited files from BLS \citep{icpsr2026cexdiary,bls2026cex}. We combine the consumer-unit, individual, diary-week, and expenditure files. We use the accompanying documentation to harmonize identifiers, demographics, resources, benefit receipt, diary dates, and expenditure categories across releases. We assign each expenditure to its recorded date and sum expenditures within consumer unit, date, and category. We add days with no expenditure so that each valid diary week contains seven rows. We leave the date and daily expenditures missing when the source date cannot be recovered. For the Social Security task, we reproduce the recipient and expenditure definitions in \citet{stephens2003third}. The cleaned file contains 2,720,802 consumer-unit-day rows for 205,003 consumer units, with each consumer unit contributing either seven or fourteen rows.

\subsection{Consumer Expenditure Survey: Interview Survey}\label{cex-interview}

\paragraph{Additional Information.} The Interview Survey is the second component of the Consumer Expenditure Surveys. It follows sampled addresses quarterly and generally interviews a consumer unit four times. Interviews record large and regularly incurred expenditures over the preceding three months. They also record household composition, employment, income, assets, and liabilities. The current design contacts roughly 13,000 addresses and obtains about 5,000 usable interviews per quarter \citep{bls2022cexhandbook,census2026cex}. BLS publishes annual public-use packages. We use official release dates for 2015--2024. Exact posting dates are unavailable for 1980--2014, so we assign those releases to March 31 of the following year.

\paragraph{Downloading and Cleaning the Data.} We download the BLS annual public-use packages for 1980, 1981, and 1984--2024 \citep{bls2026cex}. We take the FMLI consumer-unit summary files and construct one row per interview. Annual packages overlap, so some interviews appear twice. We identify an interview by NEWID, year, and month and keep the copy from the later package. We harmonize demographic, geographic, income, and expenditure fields across questionnaire changes. We create separate household identifiers for each survey design because raw identifiers can repeat after a redesign. We calculate quarterly expenditure as the sum of its three monthly components. Before-tax income covers the preceding twelve months. The cleaned file contains 1,040,182 consumer-unit interviews. These interviews form 375,509 linked series, each containing one to four quarterly interviews.

\subsection{Current Population Survey: Basic Monthly Survey}\label{cps-basic}

\paragraph{Additional Information.} The U.S. Census Bureau conducts the Current Population Survey for the U.S. Bureau of Labor Statistics. It has run monthly since 1940 and represents the civilian noninstitutional population. The current design includes about 62,000 eligible housing units and 54,000 completed interviews each month. These interviews describe roughly 105,000 people aged 16 or older \citep{bls2026cpsdesign,bls2026cpshistory}. The Basic Monthly Survey records household composition, demographics, employment, unemployment, and labor-force participation. Households enter for four months, leave for eight months, and return for four months. Public-use files are usually available 30--45 days after collection ends \citep{census2026cpsfaq}. We use exact Census release dates from January 2020 onward. For earlier months, we use 30 days after the Saturday ending the interview week that contains the nineteenth of the month.

\paragraph{Downloading and Cleaning the Data.} We download one IPUMS CPS Version 13.0 extract and its codebook \citep{flood2025ipumscps}. We interpret each variable as specified in the codebook. We recode documented missing and not-in-universe values as missing and record whether each variable exists in each month. We add a release date for every survey month. For longitudinal tasks, we link records only when CPSIDP is positive and unique in both months. We require the records to occupy the expected positions in the survey rotation and to report consistent sex, race, and age. The cleaned Basic file contains 79,330,448 person-month rows from January 1976 through June 2026. A person can appear in as many as eight monthly interviews under the 4-8-4 design.

\subsection{Current Population Survey: Displaced Worker Supplement}\label{cps-dws}

\paragraph{Additional Information.} The Census Bureau administers the Displaced Worker Supplement for BLS to CPS respondents aged 20 or older. It has generally run every two years, in January or February, since 1984. The supplement records involuntary job loss, the lost job, unemployment, benefits, relocation, subsequent work, current earnings, and health insurance. Its recall window covered five years in 1984--1992 and three years from 1994 onward. Since 1998, it has excluded self-employed workers and workers expecting recall \citep{ipums2026dwsnotes,census2024dwstech}. DWS files are published separately from Basic CPS files. We use the concurrent Basic release date when an exact DWS date is unavailable \citep{census2024releases}.

\paragraph{Downloading and Cleaning the Data.} We obtain the DWS variables in the same IPUMS CPS extract used for the Basic survey \citep{flood2025ipumscps}. We apply the same definitions of missing values and the same release dates. The cleaned file contains 1,840,305 person-wave records with a DWS status across 21 supplements from 1984 through 2024. Each person identifier appears once in these data. Of these records, 95,904 identify a displaced worker.

\subsection{University of Michigan Surveys of Consumers}\label{michigan-soc}

\paragraph{Additional Information.} The University of Michigan Survey Research Center has conducted the Surveys of Consumers since 1946. The monthly survey covers the contiguous United States and the District of Columbia. It records household finances, buying conditions, and expectations about inflation, unemployment, interest rates, and business conditions. The final sample for each month now contains about 1,000 households. Each month combines new respondents with repeat interviews about six and twelve months after the first interview \citep{michigan2026soc}. Public respondent files generally appear four weeks after the final monthly release \citep{michigan2026faq}. From 1991 onward, we use the final release date plus 28 days. For earlier months, we use the last day of the survey month because the official release table does not cover those years.

\paragraph{Downloading and Cleaning the Data.} We download the Cross-Section Archive data file, Stata dictionary, and codebook \citep{michigan2026soc}. We use the dictionary and codebook to assign variable labels, value labels, and missing values. We apply the historical revisions supplied with the archive. We link repeat interviews using the identifiers for the interviews about six and twelve months earlier. The cleaned file contains 342,345 respondent-month rows from January 1978 through February 2026. These rows form 217,641 linked respondent panels with one, two, or three interviews.

\subsection{Survey of Consumer Expectations}\label{sce}

\paragraph{Additional Information.} The Federal Reserve Bank of New York has conducted the Survey of Consumer Expectations monthly since June 2013. This internet panel follows roughly 1,200--1,300 adult household heads for as long as twelve months. It records probabilistic expectations about inflation, labor markets, income, spending, credit, housing, and household finances. Topical modules cover related subjects \citep{armantier2017sce,nyfed2026scefaq}. Micro-data is released with a nine-month lag for Core and Credit Access data, and an 18-month lag for Household Spending, Public Policy, Housing, and Labor Market data. The New York Fed does not list separate lags for Household Finance, Informal Work, or Job Search, so we use nine months for those modules. The Public Policy workbook available on May 27, 2026 contained April 2025 samples, despite the stated 18-month lag, so we therefore use nine months for Public Policy as well.

\paragraph{Downloading and Cleaning the Data.} We download the Core and topical workbooks from the New York Fed Data Bank \citep{nyfed2026scedata}. We import each workbook separately and use common names and formats for respondent identifiers, dates, weights, and variable labels. The Core file contains 186,660 respondent-month rows for 24,592 respondent identifiers. The median respondent appears in nine months. Although the stated panel duration is twelve months, 774 respondent identifiers appear in more than twelve months and the observed maximum is sixteen. The eight topical files contain between 6,809 and 38,105 rows each. We recover the February 2014 home-financing experiment, which is not included in the standard release, from the replication package of \citet{fuster2021sensitivity,fusterzafar2022data}.

\subsection{Panel Study of Income Dynamics}\label{psid}

\paragraph{Additional Information.} The University of Michigan Survey Research Center began the Panel Study of Income Dynamics in 1968. Its initial nationally representative sample included more than 18,000 people in about 5,000 families. The survey follows original sample families, their descendants, and later immigrant refresher samples. One adult generally answers the family interview. The survey covers employment, income, wealth, expenditure, health, education, and family structure. Interviews were annual through 1997 and have been biennial since 1999. PSID does not follow a fixed publication lag, so we use December 31 of the survey year for waves through 2019. The 2021 wave is dated June 30, 2023, the month of the official user guide.

\paragraph{Downloading and Cleaning the Data.} We rely on PSID-SHELF \citep{pfeffer2025psidshelf,daumler2025psidshelf}, an unofficial harmonization effort that covers 42 waves through 2021 and was published on February 24, 2025. We download the long person-wave PSID-SHELF file and the Complete Main Study file from its V2 release. We standardize variable names, sample status, family roles, survey years, income years, occupations, value labels, and release dates. We also select 306 fields from the raw Complete Main Study file (also provided in the release) to add to PSID-SHELF, covering spouse roles, annual hours, food expenditure, and reasons for job endings. The final file contains 3,533,082 person-wave rows for 84,121 permanent person identifiers and 42 survey waves. It contains one row for every person and wave, including years when the person was not interviewed or was not active in the sample.

\subsection{U.S. Decennial Census}\label{decennial-census}

\paragraph{Additional Information.} The Census Bureau has conducted a national population census every ten years since 1790. Sampling for the detailed long form began in 1940 and ended after 2000, when the American Community Survey replaced it. The 2000 long form contained 52 questions and went to about one in six households. It covered demographics, education, employment, income, migration, and housing \citep{census2026decennial,census2026census2000}. We use two IPUMS USA 1-percent samples: 1990 and 2000 \citep{ruggles2025ipumsusa,ipums2026usasamples}. We use December 31 of each census year as the release date because a consistent set of exact dates is unavailable.

\paragraph{Downloading and Cleaning the Data.} We download the six samples in one IPUMS USA Version 16.0 Stata extract with its codebook \citep{ruggles2025ipumsusa}. We retain the six sample codes and apply the IPUMS definitions of missing values. We keep the two 1970 forms separate because they asked different questions. We harmonize migration variables across years and identify topcoded and censored income values. The cleaned file contains 13,445,846 person rows across six independent cross-sections. Each person appears once in a given cross-section.

\subsection{Additional Data Sources}\label{appendix:additional-data-sources}

\paragraph{National Macroeconomic and Financial Series.} We construct the quarterly national variables from the July 2026 FRED-QD file \citep{mccracken2021fredqd,fred2026qd}. We use real GDP, unemployment, all-items CPI, the federal funds rate, house prices, the S\&P 500, and the 30-year mortgage rate. The corresponding FRED-QD columns are GDPC1, UNRATE, CPIAUCSL, FEDFUNDS, USSTHPI, S\&P 500, and MORTGAGE30US. For each survey sample, we use only quarters completed before the survey date. We include the preceding four quarterly values and one five-year summary. We calculate quarterly percentage changes and 20-quarter compound growth for quantities and prices. Rate variables enter as quarterly levels and 20-quarter averages. These series include revisions present in the July 2026 file. They are not historical vintages available on each survey date.

\paragraph{Price and State Macroeconomic Series.} We download monthly all-items CPI-U and the eight major-group indexes from BLS \citep{bls2026cpi}. The groups cover food and beverages, housing, apparel, transportation, medical care, recreation, education and communication, and other goods and services. We average the monthly indexes within each quarter and calculate quarterly percentage changes. For the Census migration task, we also use three additional sources: annual state unemployment from BLS Local Area Unemployment Statistics \citep{bls2026laus}, from which we use the unadjusted M13 annual average; state and national per-capita personal income from BEA SAINC1 line 3, where we calculate annual growth \citep{bea2026sainc}; and the FHFA traditional all-transactions quarterly house-price index for each state and the nation \citep{fhfa2026hpi}, which we average within year to calculate annual growth.

\hypertarget{xgboost-models}{%
\section{XGBoost Models}\label{appendix:xgboost}}

\paragraph{Specification.} All models are estimated in Python 3.13.5 using XGBoost 3.3.0 \citep{chen2016xgboost}. For each task, the model receives the same predictor variables represented in the household profile supplied to the language models. We adopt the pre-tuned parameter values proposed by \citet{holzmuller2024better} and select the number of boosting rounds using the relevant \bench evaluation metric. Training stops after 300 consecutive rounds without improvement. Predictions use the iteration with the best validation criterion; the model is not subsequently refit on the combined training and validation samples. We think this specification balances performance and training cost in terms of time and compute. Experiments with package-default parameters and no early stopping were dominated by this specification, while full random search over hyperparameter values yielded small performance gains, on the order of 1 to 2\%, and increased training time by a factor of 8. Results are available on request.

\paragraph{Hyperparameter values.} Here are the exact settings for each target type:

\textit{Numeric targets.} We train one regression for each outcome with the following parameter values: at most 1,000 trees, learning rate 0.05, maximum depth 9, row sampling 0.70, full column sampling, minimum child weight 2, and zero L1, L2, and split-gain penalties. The loss criterion option is \texttt{reg:absoluteerror}. Direct early stopping minimizes validation mean absolute error.

\textit{Categorical targets.} We train one classifier for each outcome, with the following parameter values: at most 1,000 trees, learning rate 0.08, maximum depth 6, row sampling 0.65, column sampling by level 0.90, minimum child weight 0.000005, and the same zero penalties. Binary models use \texttt{binary:\allowbreak{}logistic}; multiclass models use \texttt{multi:\allowbreak{}softprob}. Class predictions select the category with the highest predicted probability. Early stopping in the literature-informed specification minimizes one minus Macro-F1, calculated over the complete set of categories defined for the task.

\textit{Probability targets.} We train one regression for each entry of the probability vector with the following parameter values: at most 2,000 trees, learning rate 0.05, maximum depth 9, row sampling 0.70, full column sampling, minimum child weight 2, and zero L1, L2, and split-gain penalties. The loss criterion option is \texttt{reg:absoluteerror}. A binary distribution is estimated through its event probability, with the complement restored during reconstruction. For distributions with more than two outcomes, XGBoost trains one tree per output probability (\texttt{one\_output\_per\_tree}) and then predictions are projected onto the unit simplex. Early stopping minimizes validation total-variation distance after projection.

\paragraph{Training samples.} For each task, we use at most 400,000 training rows and 50,000 validation rows. Tasks with less data use all of theirs. Across the 32 tasks, the training split holds 3,544,247 rows and the validation split 447,819.

\section{Additional Results and Discussion}
\label{appendix:additional_results}

\subsection{Persistence in Household Probability Reports}
\label{appendix:probability_persistence}

\begin{figure}[H]
    \centering
    \includegraphics[page=1,width=0.95\linewidth]{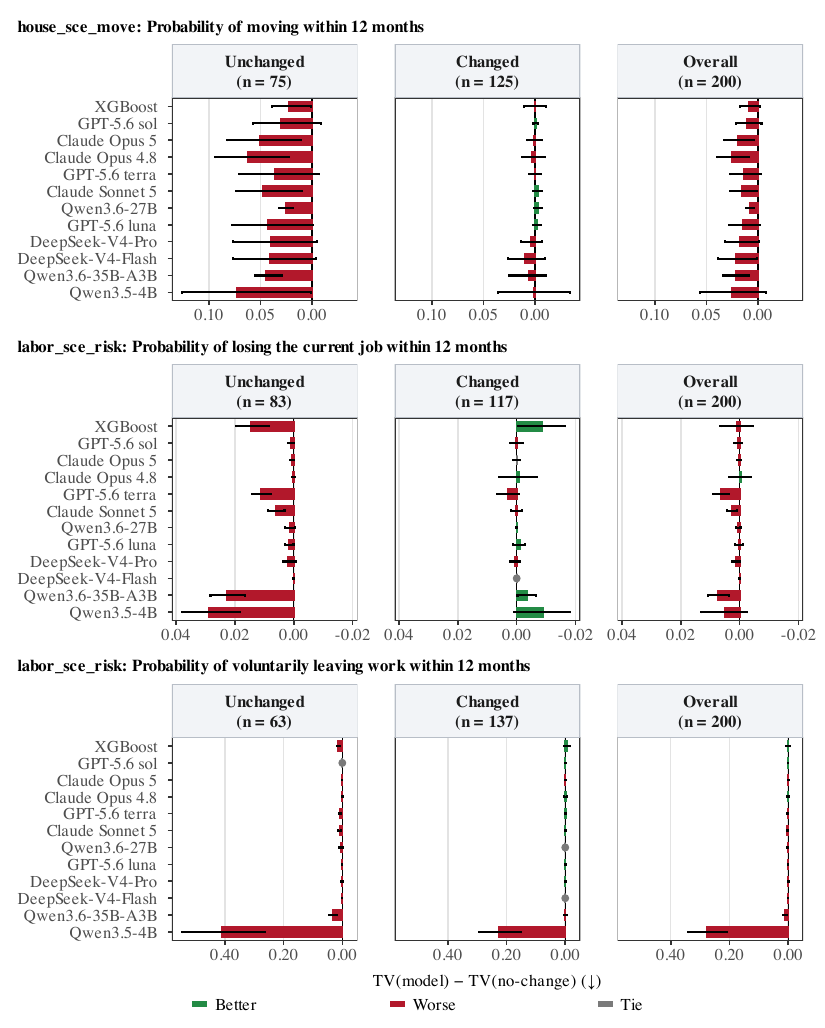}
    \caption{Difference in mean total variation by target, separately for unchanged, changed and all target probability vectors. Bars show $\mathrm{TV}(\mathrm{model})-\mathrm{TV}(\mathrm{no\mbox{-}change})$ on the same valid observations, without division. Negative values favor the model; zero denotes equal performance. Each row reports one target; multi-target tasks are not pooled. The figure shows twelve prediction models using pre-cutoff valid responses; the no-change row is omitted. Facet titles count evaluation rows for that target before excluding invalid model responses. Unchanged and Changed partition those rows. Green denotes a negative difference, red a positive difference and grey a tie. Black bars are 95\% basic paired IID bootstrap intervals for the difference, resampling rows within task and omitting draws without a valid subgroup observation. Each target uses one scale across its three panels; scales differ across targets. Continued on the next page.}
    \label{fig:appendix-probability-persistence}
\end{figure}
\clearpage
\addtocounter{figure}{-1}
\begingroup
\renewcommand{\theHfigure}{\thesection.\arabic{figure}.continued2}
\begin{figure}[H]
    \centering
    \includegraphics[page=2,width=0.95\linewidth]{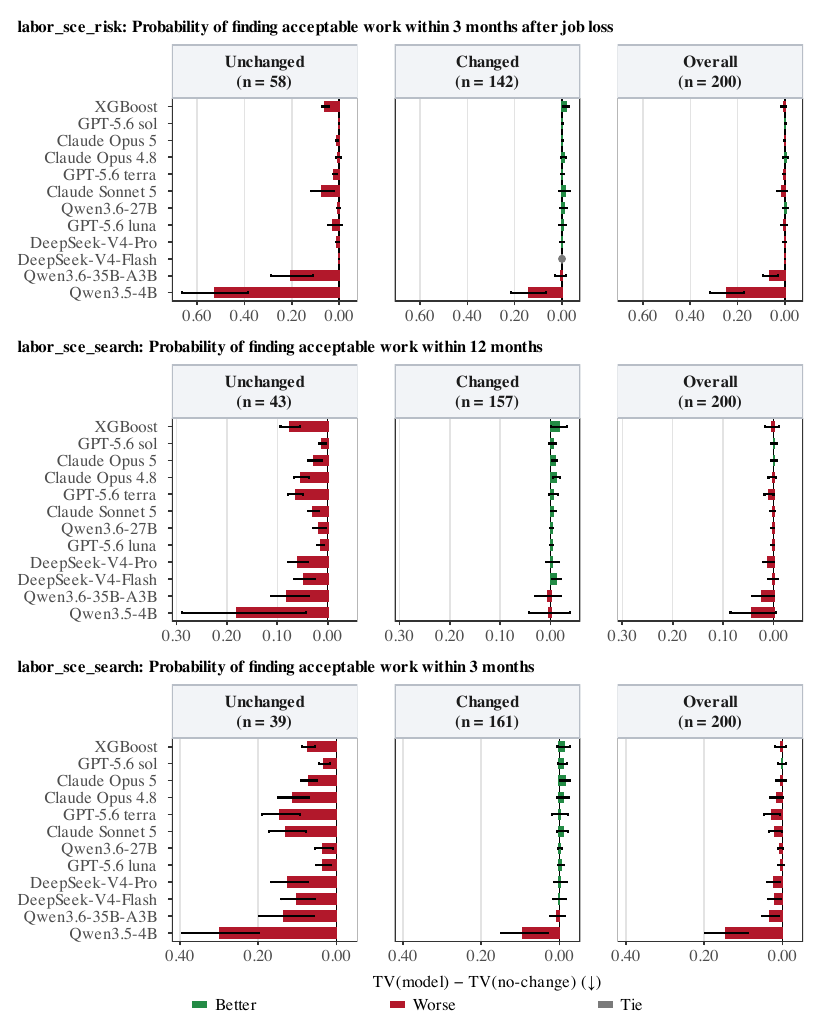}
    \caption{Difference in mean total variation by target (continued). The model set, sample definitions and interval construction are the same as on the first page. Continued on the next page.}
\end{figure}
\endgroup

\clearpage
\addtocounter{figure}{-1}
\begingroup
\renewcommand{\theHfigure}{\thesection.\arabic{figure}.continued3}
\begin{figure}[H]
    \centering
    \includegraphics[page=3,width=0.95\linewidth]{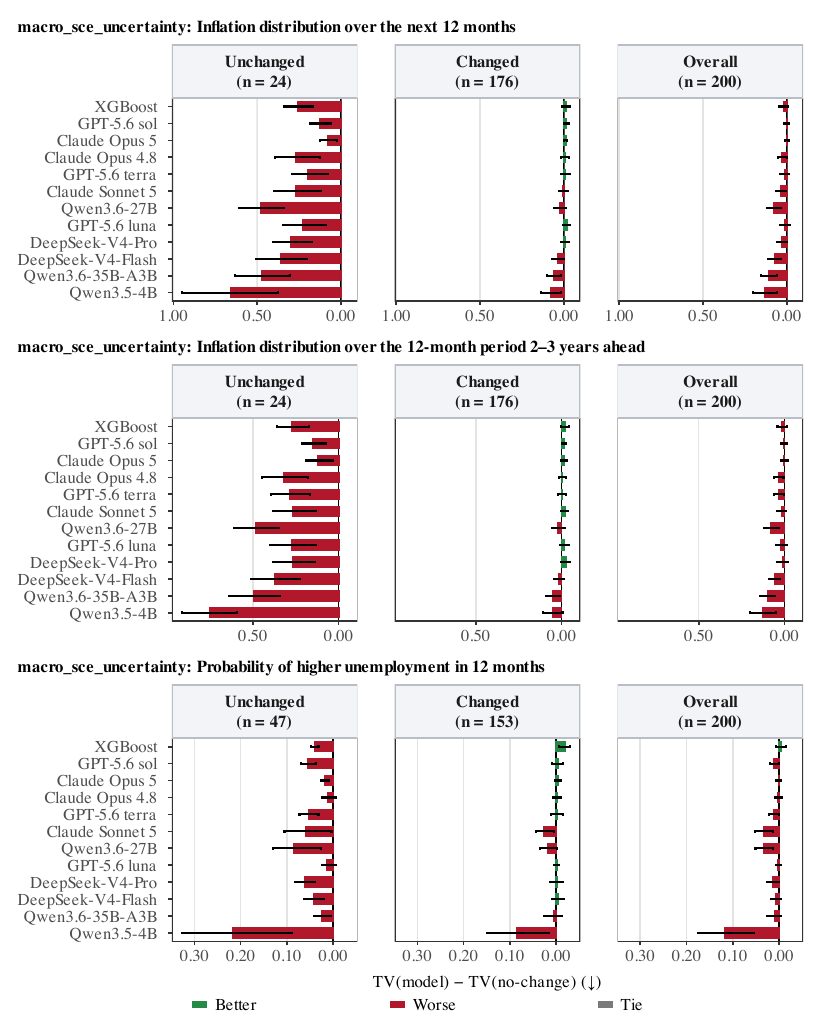}
    \caption{Difference in mean total variation by target (continued). The model set, sample definitions and interval construction are the same as on the first page. Continued on the next page.}
\end{figure}
\endgroup

\clearpage
\addtocounter{figure}{-1}
\begingroup
\renewcommand{\theHfigure}{\thesection.\arabic{figure}.continued4}
\begin{figure}[H]
    \centering
    \includegraphics[page=4,width=0.95\linewidth]{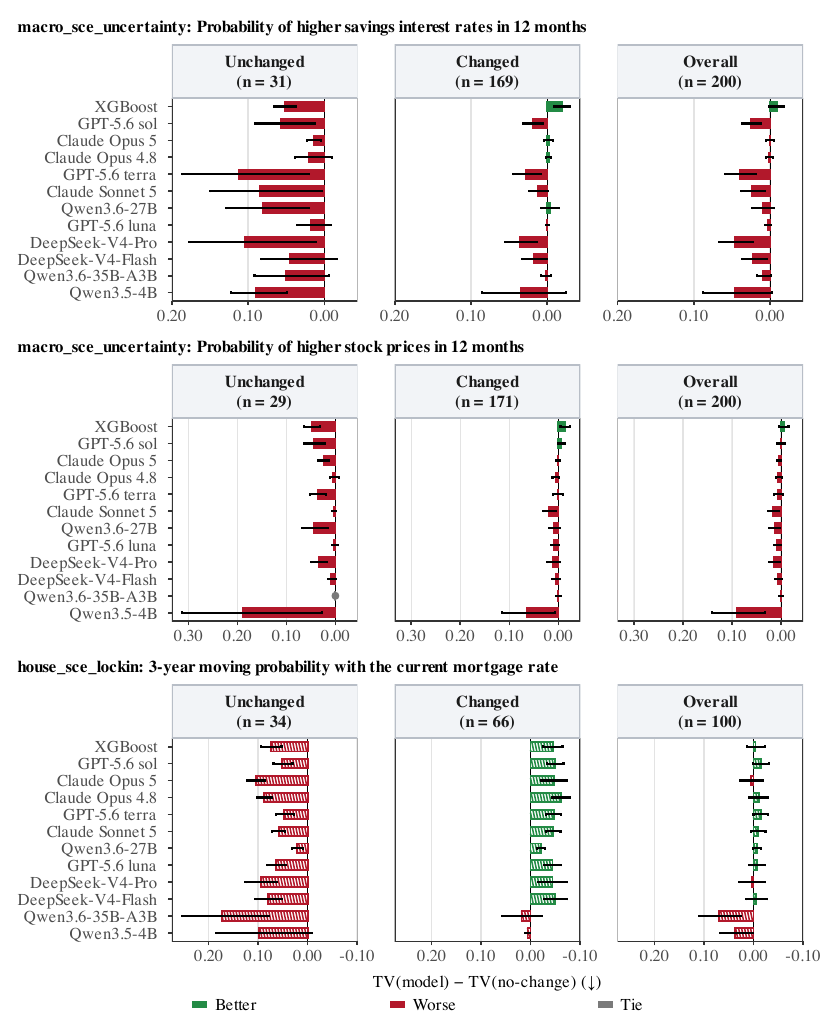}
    \caption{Difference in mean total variation by target (continued). Solid bars denote baseline tasks and hatched bars denote the policy task. Each model and its no-change comparator use the same valid observations in both the point estimate and every bootstrap draw.}
\end{figure}
\endgroup

\FloatBarrier
\clearpage

\subsection{Benchmark Results}
\label{appendix:benchmark_results_all_models}

\FloatBarrier

\begingroup
\let\HouseholdBenchTable\table
\let\endHouseholdBenchTable\endtable
\renewenvironment{table}[1][]{\HouseholdBenchTable[H]}{\endHouseholdBenchTable}

\begin{figure}[H]
    \centering
    {\small\textbf{(a) Full post-cutoff sample}\par}
    \includegraphics[width=\linewidth]{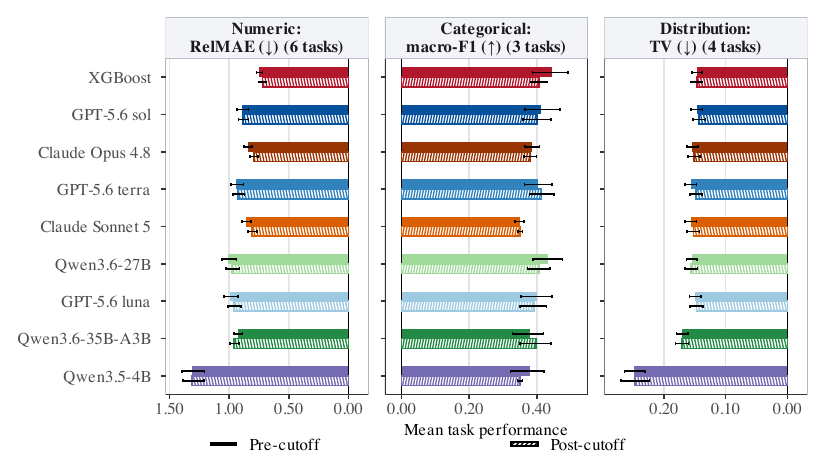}
    \par\medskip
    {\small\textbf{(b) Rows released from July 2026}\par}
    \includegraphics[width=\linewidth]{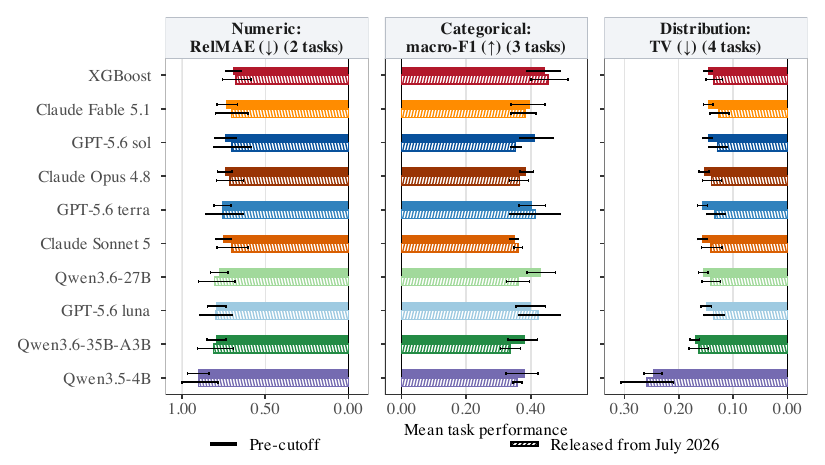}
    \caption{Mean performance before and after the cutoff. Panel (a) compares matched pre-cutoff and post-cutoff task samples for the main-leaderboard models evaluated in both samples, excluding the no-change baseline. Panel (b) compares pre-cutoff performance with performance on the 454 rows across nine tasks released on or after 1 July 2026, the first release wave after Claude Fable 5.1's June 2026 knowledge cutoff; its pre-cutoff sample covers the same nine tasks. Models follow the main-leaderboard order. Solid bars report equal-task means of the raw task metric in each panel's pre-cutoff sample; hatched bars report the corresponding later-sample means. Whiskers are pointwise 95\% basic IID bootstrap intervals. The RelMAE and total-variation axes are reversed because lower values indicate better performance.}
    \label{fig:appendix-post-minus-pre-cutoff-all-models}
\end{figure}

\FloatBarrier

\endgroup

\FloatBarrier
\clearpage

\subsection{Performance Over Time}
\label{appendix:performance_over_time}

\begin{figure}[H]
    \centering
    \includegraphics[width=\linewidth,height=0.88\textheight,keepaspectratio]{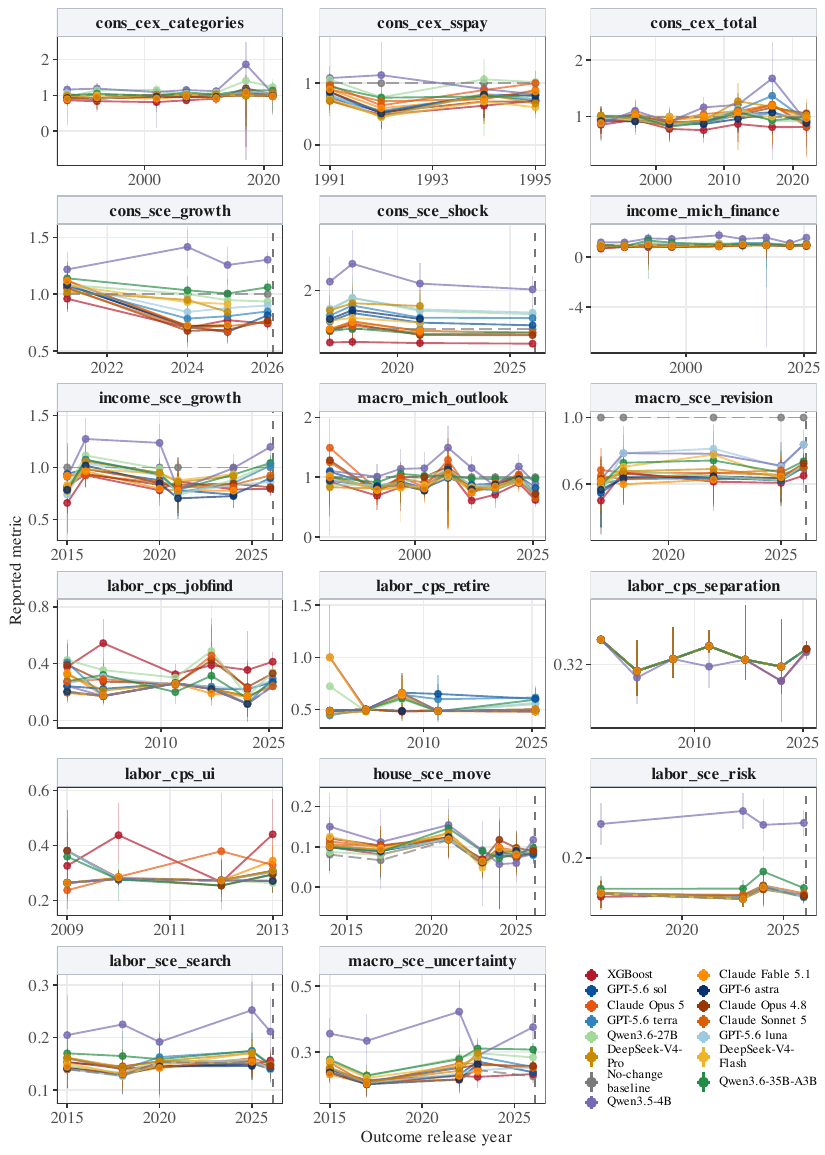}
    \caption{Model performance over outcome-release time for the main-leaderboard models. The legend follows the main-leaderboard order. Vertical bars are pointwise 95\% basic IID bootstrap intervals.}
    \label{fig:appendix-performance-over-time-all-models}
\end{figure}

\FloatBarrier
\clearpage
\flushbottom

\section{Fine-Tuning With a Number-Token Loss}
\label{appendix:ntl}

Besides supervised fine-tuning (Section~\ref{sec:ft}), we test SFT with a \emph{number-token loss} (SFT w/ NTL)~\citep{zausinger2025regress}. Our motivation is that two of our three target types (numeric and probability prediction) ask the model to predict numbers, and that SFT performs suboptimally on numerical prediction. The intuition is that the cross-entropy loss used in SFT treats the digit tokens as unordered labels and cannot express the closeness of predicted numbers.
Motivated by this, \citet{zausinger2025regress} propose adding a number-token loss, which minimizes the Wasserstein distance between the predicted and ground-truth numerical numbers. We expect that SFT with NTL brings better performance on the numerical and probability prediction tasks compared to SFT.
We note that NTL loss only adds about 1\% computation overhead.

\providecommand{\hbbetter}[1]{{\setlength{\fboxsep}{1.5pt}\colorbox{green!20}{#1}}}
\providecommand{\hbworse}[1]{{\setlength{\fboxsep}{1.5pt}\colorbox{red!20}{#1}}}
\begin{table}[H]
\centering
\resizebox{\textwidth}{!}{%
\small
\begin{tabular}{llcccc|ccc}
\toprule
 &  &  & \multicolumn{3}{c}{\textbf{Baseline tasks} \scriptsize (seen in fine-tuning)} & \multicolumn{3}{c}{\textbf{Policy-response tasks} \scriptsize (held out)} \\
\cmidrule(lr){4-6} \cmidrule(lr){7-9}
 & \textbf{Model} & \textbf{$K$} & \textbf{RelMAE} $\downarrow$ & \textbf{Macro-F1} $\uparrow$ & \textbf{TV} $\downarrow$ & \textbf{RelMAE} $\downarrow$ & \textbf{Macro-F1} $\uparrow$ & \textbf{TV} $\downarrow$ \\
 &  &  & \scriptsize (\# = 8) & \scriptsize (\# = 6) & \scriptsize (\# = 4) & \scriptsize (\# = 10) & \scriptsize (\# = 3) & \scriptsize (\# = 1) \\
\midrule
\multirow{4}{*}{\scriptsize Reference} & XGBoost & 1 & 0.84 & 0.46 & 0.15 & 0.79 & 0.36 & 0.12 \\
\hdashline[1pt/2pt]
 & GPT-6 astra & 1 & \underline{0.89} & 0.39 & \underline{0.15} & 0.94 & 0.32 & \textbf{0.11} \\
 & Claude Opus 5 & 1 & 0.90 & \textbf{0.47} & \underline{0.15} & \underline{0.91} & 0.28 & 0.13 \\
 & Fable 5.1 & 1 & \underline{0.89} & 0.44 & \underline{0.15} & \textbf{0.86} & \textbf{0.35} & \textbf{0.11} \\
\midrule
\multirow{6}{*}{\scriptsize SFT} & Qwen3.5-4B & 1 & 1.14 & 0.39 & 0.25 & 1.43 & 0.25 & 0.15 \\
 & \quad + SFT & 1 & 1.00\,\hbbetter{\scriptsize ($-$0.14)} & 0.45\,\hbbetter{\scriptsize ($+$0.06)} & 0.16\,\hbbetter{\scriptsize ($-$0.09)} & 1.30\,\hbbetter{\scriptsize ($-$0.14)} & \underline{0.33}\,\hbbetter{\scriptsize ($+$0.08)} & 0.17\,\hbworse{\scriptsize ($+$0.01)} \\
 & \cellcolor{gray!20}\quad + SFT w/ NTL & \cellcolor{gray!20}1 & \cellcolor{gray!20}0.97\,\hbbetter{\scriptsize ($-$0.17)} & \cellcolor{gray!20}\underline{0.46}\,\hbbetter{\scriptsize ($+$0.07)} & \cellcolor{gray!20}\underline{0.15}\,\hbbetter{\scriptsize ($-$0.10)} & \cellcolor{gray!20}1.23\,\hbbetter{\scriptsize ($-$0.20)} & \cellcolor{gray!20}\underline{0.33}\,\hbbetter{\scriptsize ($+$0.07)} & \cellcolor{gray!20}0.14\,\hbbetter{\scriptsize ($-$0.02)} \\
 & Qwen3.6-27B & 1 & 0.96 & 0.45 & 0.16 & 1.04 & 0.26 & \underline{0.12} \\
 & \quad + SFT & 1 & 1.00\,\hbworse{\scriptsize ($+$0.03)} & \underline{0.46}\,\hbbetter{\scriptsize ($+$0.01)} & 0.16\,\hbworse{\scriptsize ($+$0.01)} & 1.18\,\hbworse{\scriptsize ($+$0.14)} & 0.30\,\hbbetter{\scriptsize ($+$0.04)} & 0.15\,\hbworse{\scriptsize ($+$0.04)} \\
 & \cellcolor{gray!20}\quad + SFT w/ NTL & \cellcolor{gray!20}1 & \cellcolor{gray!20}0.97\,\hbworse{\scriptsize ($+$0.01)} & \cellcolor{gray!20}0.44\,{\scriptsize (0.00)} & \cellcolor{gray!20}\underline{0.15}\,{\scriptsize (0.00)} & \cellcolor{gray!20}1.11\,\hbworse{\scriptsize ($+$0.07)} & \cellcolor{gray!20}0.27\,\hbbetter{\scriptsize ($+$0.01)} & \cellcolor{gray!20}0.13\,\hbworse{\scriptsize ($+$0.02)} \\
\midrule
\multirow{6}{*}{\scriptsize \shortstack[l]{SFT\\+ aggregation}} & Qwen3.5-4B & 16 & 1.11 & 0.38 & 0.24 & 1.31 & 0.24 & 0.15 \\
 & \quad + SFT & 16 & \textbf{0.87}\,\hbbetter{\scriptsize ($-$0.24)} & 0.42\,\hbbetter{\scriptsize ($+$0.05)} & \underline{0.15}\,\hbbetter{\scriptsize ($-$0.09)} & 1.13\,\hbbetter{\scriptsize ($-$0.18)} & 0.31\,\hbbetter{\scriptsize ($+$0.06)} & 0.16\,\hbworse{\scriptsize ($+$0.01)} \\
 & \cellcolor{gray!20}\quad + SFT w/ NTL & \cellcolor{gray!20}16 & \cellcolor{gray!20}\textbf{0.87}\,\hbbetter{\scriptsize ($-$0.24)} & \cellcolor{gray!20}0.43\,\hbbetter{\scriptsize ($+$0.06)} & \cellcolor{gray!20}\textbf{0.14}\,\hbbetter{\scriptsize ($-$0.10)} & \cellcolor{gray!20}1.12\,\hbbetter{\scriptsize ($-$0.19)} & \cellcolor{gray!20}0.32\,\hbbetter{\scriptsize ($+$0.08)} & \cellcolor{gray!20}0.14\,\hbbetter{\scriptsize ($-$0.01)} \\
 & Qwen3.6-27B & 16 & 0.95 & 0.43 & \underline{0.15} & 1.01 & 0.25 & \underline{0.12} \\
 & \quad + SFT & 16 & \textbf{0.87}\,\hbbetter{\scriptsize ($-$0.08)} & 0.44\,\hbbetter{\scriptsize ($+$0.01)} & \underline{0.15}\,\hbbetter{\scriptsize ($-$0.01)} & 1.03\,\hbworse{\scriptsize ($+$0.02)} & 0.27\,\hbbetter{\scriptsize ($+$0.02)} & 0.14\,\hbworse{\scriptsize ($+$0.02)} \\
 & \cellcolor{gray!20}\quad + SFT w/ NTL & \cellcolor{gray!20}16 & \cellcolor{gray!20}\textbf{0.87}\,\hbbetter{\scriptsize ($-$0.08)} & \cellcolor{gray!20}0.45\,\hbbetter{\scriptsize ($+$0.02)} & \cellcolor{gray!20}\textbf{0.14}\,\hbbetter{\scriptsize ($-$0.01)} & \cellcolor{gray!20}1.01\,{\scriptsize (0.00)} & \cellcolor{gray!20}0.26\,\hbbetter{\scriptsize ($+$0.01)} & \cellcolor{gray!20}0.13\,\hbworse{\scriptsize ($+$0.01)} \\
\bottomrule
\end{tabular}%
}
\caption{Results of SFT and SFT with the number-token loss. The $K$ column is the number of aggregated predictions. Best among the language models in bold, second best underlined. XGBoost is refitted on each task it is scored on, so its policy-response scores are fitted rather than held out.}
\label{tab:sft_scores_ntl}
\end{table}

SFT with the number-token loss achieves better performance on numeric and probability tasks than SFT alone (Table~\ref{tab:sft_scores_ntl}). Compared with SFT alone, adding the NTL lowers RelMAE by a further 3\% for Qwen3.5-4B and 2\% for Qwen3.6-27B on the baseline tasks. On the policy-response tasks, the additional improvement is 5\% and 6\%. It also lowers TV by 7\% to 19\% across the two models. However, the additional gains from NTL diminish after prediction aggregation.

\clearpage

\section{Model Ensembles}
\label{sec:ensembles}

\subsection{Aggregating Model Predictions}
\label{appendix:two_draws}

In Table~\ref{tab:k2_draws}, we show the detailed results of aggregating a model's predictions. We note that aggregating the two predictions of proprietary models (i.e., Fable 5.1, Claude Opus 5, and GPT-5.6 sol) does not change their prediction performance: the relative changes on numeric and probability tasks are minimal (less than 0.5\%), and the changes on categorical tasks are mostly negative. This implies that aggregating more predictions from proprietary models would not improve the performance either. On the contrary, on the numeric and probability prediction tasks, the fine-tuned Qwen3.5-4B and Qwen3.6-27B improve from the aggregation with only two predictions ($K=2$), and the improvement becomes larger at $K=16$.

\begin{table}[H]
\centering
\resizebox{\textwidth}{!}{%
\small
\begin{tabular}{llccccc}
\toprule
 & \textbf{Model} & \textbf{$K=1$} & \textbf{$K=2$} & \textbf{\shortstack{Improvement of\\$K=2$ from $K=1$ (\%)}} & \textbf{$K=16$} & \textbf{\shortstack{Improvement of\\$K=16$ from $K=1$ (\%)}} \\
\midrule
\multicolumn{7}{l}{\textbf{Numeric}, RelMAE $\downarrow$ \scriptsize (\# = 18)} \\
\midrule
\multirow{3}{*}{\scriptsize Proprietary} & Fable 5.1 & 0.878 & 0.879 & $-$0.2 & -- & -- \\
 & Claude Opus 5 & 0.905 & 0.902 & 0.3 & -- & -- \\
 & GPT-5.6 sol & 0.910 & 0.910 & 0.0 & -- & -- \\
\hdashline[1pt/2pt]
\multirow{6}{*}{\scriptsize Open-weight} & Qwen3.5-4B & 1.304 & 1.272 & 2.5 & 1.223 & 6.3 \\
 & \quad + SFT & 1.164 & 1.087 & 6.6 & 1.017 & 12.6 \\
 & \quad + SFT w/ NTL & 1.117 & 1.059 & 5.2 & 1.008 & 9.8 \\
 & Qwen3.6-27B & 1.007 & 0.999 & 0.8 & 0.984 & 2.2 \\
 & \quad + SFT & 1.101 & 1.037 & 5.8 & 0.956 & 13.1 \\
 & \quad + SFT w/ NTL & 1.050 & 1.001 & 4.6 & 0.950 & 9.5 \\
\midrule
\multicolumn{7}{l}{\textbf{Categorical}, Macro-F1 $\uparrow$ \scriptsize (\# = 9)} \\
\midrule
\multirow{3}{*}{\scriptsize Proprietary} & Fable 5.1 & 0.413 & 0.400 & $-$3.1 & -- & -- \\
 & Claude Opus 5 & 0.406 & 0.401 & $-$1.2 & -- & -- \\
 & GPT-5.6 sol & 0.399 & 0.400 & 0.2 & -- & -- \\
\hdashline[1pt/2pt]
\multirow{6}{*}{\scriptsize Open-weight} & Qwen3.5-4B & 0.346 & 0.332 & $-$4.1 & 0.332 & $-$4.0 \\
 & \quad + SFT & 0.410 & 0.409 & $-$0.2 & 0.384 & $-$6.4 \\
 & \quad + SFT w/ NTL & 0.416 & 0.401 & $-$3.4 & 0.396 & $-$4.8 \\
 & Qwen3.6-27B & 0.385 & 0.381 & $-$1.1 & 0.374 & $-$2.9 \\
 & \quad + SFT & 0.406 & 0.397 & $-$2.2 & 0.386 & $-$4.9 \\
 & \quad + SFT w/ NTL & 0.388 & 0.397 & 2.4 & 0.388 & 0.0 \\
\midrule
\multicolumn{7}{l}{\textbf{Probability}, TV $\downarrow$ \scriptsize (\# = 5)} \\
\midrule
\multirow{3}{*}{\scriptsize Proprietary} & Fable 5.1 & 0.139 & 0.139 & 0.0 & -- & -- \\
 & Claude Opus 5 & 0.144 & 0.144 & 0.2 & -- & -- \\
 & GPT-5.6 sol & 0.139 & 0.140 & $-$0.4 & -- & -- \\
\hdashline[1pt/2pt]
\multirow{6}{*}{\scriptsize Open-weight} & Qwen3.5-4B & 0.229 & 0.233 & $-$1.6 & 0.221 & 3.4 \\
 & \quad + SFT & 0.164 & 0.156 & 4.5 & 0.148 & 9.3 \\
 & \quad + SFT w/ NTL & 0.148 & 0.146 & 1.5 & 0.142 & 4.4 \\
 & Qwen3.6-27B & 0.148 & 0.148 & $-$0.1 & 0.146 & 0.8 \\
 & \quad + SFT & 0.161 & 0.155 & 4.0 & 0.147 & 9.0 \\
 & \quad + SFT w/ NTL & 0.147 & 0.145 & 1.1 & 0.140 & 4.7 \\
\bottomrule
\end{tabular}%
}
\caption{Aggregating predictions of one model, pre-cutoff test split. $K=1$ is the score of one draw, and $K=2$ and $K=16$ are the scores of the aggregate of two and sixteen draws. Each improvement is the change from $K=1$ as a percentage of the $K=1$ score, positive where the metric gets better.}
\label{tab:k2_draws}
\end{table}

\clearpage

\section{Inference and Fine-Tuning Settings}\label{appendix:inference}

\subsection{Inference Settings}

We prompt every model at its provider's default sampling settings.
The Qwen models are run with non-thinking mode and use the recommended non-thinking inference parameters for general tasks (temperature=$0.7$, top-$p$=$0.8$, top-$k$=$20$)\footnote{\url{https://huggingface.co/Qwen/Qwen3.5-4B}}.
The GPT models and Claude models run at their default reasoning effort.
The DeepSeek models run at medium reasoning effort.

\subsection{Fine-Tuning Settings}
\label{appendix:ft_settings}

We fine-tune Qwen3.5-4B and Qwen3.6-27B with LoRA~\citep{hu2022lora} on the 100,702 training data points described in Section~\ref{sec:ft}.
Both models share the hyperparameters in Table~\ref{tab:ft_settings}.
We train each model on four NVIDIA A100 80GB GPUs.

\begin{table}[H]
\centering
\small
\begin{tabular}{ll}
\toprule
\textbf{Hyperparameter} & \textbf{Value} \\
\midrule
LoRA rank & 8 \\
LoRA alpha & 32 \\
LoRA dropout & 0.05 \\
LoRA target modules & all linear layers of the language model \\
Optimizer & AdamW ($\beta_1=0.9$, $\beta_2=0.95$, weight decay $0.1$) \\
Peak learning rate & $5\times10^{-5}$ \\
Learning-rate schedule & linear warmup over the first 10\% of steps, then cosine decay \\
Gradient clipping & 1.0 \\
Epochs & 1 \\
Effective batch size & 128 \\
Maximum sequence length & 2,048 tokens \\
Precision & bfloat16 \\
Random seed & 42 \\
\bottomrule
\end{tabular}
\caption{Fine-tuning hyperparameters, shared by Qwen3.5-4B and Qwen3.6-27B and by both training losses.}
\label{tab:ft_settings}
\end{table}

\section{Evaluation Tasks}\label{appendix:tasks}

This appendix describes each of the 32 tasks in \bench in detail, including information about the target, the predictors, sample construction and splits, and a full prompt example.

\hypertarget{task:cons_cex_categories}{}
\subsection{cons\_\allowbreak{}cex\_\allowbreak{}categories}

\textbf{Target.} Household expenditure over the last three months, in current U.S. Dollars, for twelve categories: food, alcoholic beverages, housing, apparel and services, transportation, health care, entertainment, personal care, reading, education, tobacco, and miscellaneous. These aggregate to total expenditure, as used in other CEX tasks such as \texttt{cons\_cex\_total}. We take these expenditure variables directly from the raw CEX data, where they are constructed as aggregates of more detailed item-level expenditures reported directly by respondents.\par

\textbf{Predictors.} Spending history (up to three prior three-month periods of category expenditure), before-tax household income over the past 12 months as recorded on the current interview row, household composition (household size, number of adults, number of children), respondent demographics and geography (age, sex, race, education, marital status, region, urban or rural status), core macro context (real GDP growth, unemployment rate, headline CPI inflation, and the effective federal funds rate over the four most recently completed calendar quarters, plus five-year quarterly references), and disaggregated CPI inflation (major-group inflation over the same four calendar quarters, plus five-year quarterly averages).\par

\textbf{No-change baseline.} For each expenditure category, the no-change prediction is the household's expenditure in that category reported in the immediately preceding interview, conducted three months earlier.\par

\textbf{Sample construction.} Starting from our clean CEX Interview panel, we merge core macro variables from FRED-QD and disaggregated CPI from the BLS's CPI database. We then construct up to three lags of income and expenditure variables. These need to form a continuous history: if a household missed an interview, we set deeper lags to missing as well. Next, we drop rows that are missing one or more of the target variables, and also rows with current or first-history predictors missing (further lags are allowed to remain missing). Then, we require current annual before-tax household income to be at least \$1,000, current and lagged total expenditure to be at least \$250 minimum, and require every displayed expenditure category to be nonnegative. Finally, we trim the top and bottom percentiles of current annual income, current total expenditure and lagged total expenditure strictly, with the trimming bounds computed within each year and calendar quarter.\par

\begin{tcolorbox}[
  breakable,
  lines before break=8,
  colback=white,
  colframe=black!55,
  boxrule=0.45pt,
  arc=0pt,
  left=7pt,
  right=7pt,
  top=7pt,
  bottom=7pt
]
\begingroup
\setlength{\parindent}{0pt}
\setlength{\parskip}{5pt}
\hyphenpenalty=1800
\exhyphenpenalty=1800
\emergencystretch=0.8em
{\centering\fontsize{7.7}{9.0}\selectfont\bfseries Prompt example\par
\texttt{cons\_cex\_categories}\par}
\medskip
{\color{black!25}\hrule height 0.35pt}
\vspace{5pt}
\fontsize{7.2}{8.55}\selectfont\rmfamily
\paragraph{\fontsize{7.2}{8.55}\selectfont System.} You are the head of an American household making decisions and forming expectations about income, spending, and saving.\par
\paragraph{\fontsize{7.2}{8.55}\selectfont User.}
Here is some background information about yourself and your household. You are a 32-year-old white man, are currently married, and have a graduate or professional degree. You currently reside in an urban area in the West. Your household has 2 adults and 1 child, for 3 people in total.\par

Here are your household's recent spending and income. Note that the current month is February, and that \$ denotes amounts in U.S. dollars. Between nine and twelve months ago, your household spent \$1,170 on food; \$15 on alcoholic beverages; \$4,369 on housing; \$424 on apparel and services; \$2,739 on transportation; \$1,172 on health care; \$496 on entertainment; \$248 on personal care; \$64 on reading; \$0 on education; \$0 on tobacco; and \$750 on miscellaneous. Between six and nine months ago, your household spent \$1,352 on food; \$155 on alcoholic beverages; \$2,494 on housing; \$369 on apparel and services; \$3,403 on transportation; \$644 on health care; \$600 on entertainment; \$108 on personal care; \$20 on reading; \$0 on education; \$0 on tobacco; and \$0 on miscellaneous. Between three and six months ago, your household spent \$1,830 on food; \$0 on alcoholic beverages; \$2,436 on housing; \$1,171 on apparel and services; \$1,886 on transportation; \$1,449 on health care; \$572 on entertainment; \$90 on personal care; \$63 on reading; \$0 on education; \$0 on tobacco; and \$30 on miscellaneous. Your before-tax household income over the past 12 months was \$47,050.\par

You are also aware of recent national economic conditions. Over the four most recently completed calendar quarters, listed from oldest to most recent, U.S. real GDP changed by 0.94\%, 0.83\%, 1.33\%, and 1.64\% quarter over quarter; the unemployment rate was 4.30\%, 4.27\%, 4.23\%, and 4.07\%; overall consumer prices changed by 0.37\%, 0.75\%, 0.74\%, and 0.74\% quarter over quarter; and the effective federal funds rate was 4.73\%, 4.75\%, 5.09\%, and 5.31\%. Over the five years ending in the most recently completed quarter, compound average quarterly real GDP growth was 1.02\%, the average unemployment rate was 4.93\%, compound average quarterly CPI inflation was 0.59\%, and the average effective federal funds rate was 5.38\%.\par

You are also aware of changes in prices for specific spending categories. Over the same four quarters, listed from oldest to most recent, prices changed by 0.61\%, 0.26\%, 0.41\%, and 0.63\% quarter over quarter for food and beverages; 0.35\%, 0.55\%, 0.65\%, and 0.63\% quarter over quarter for housing; -1.18\%, 0.61\%, -0.38\%, and 0.03\% quarter over quarter for apparel; -0.50\%, 2.02\%, 1.74\%, and 1.46\% quarter over quarter for transportation; 0.77\%, 0.85\%, 0.95\%, and 0.90\% quarter over quarter for medical care; 0.33\%, 0.26\%, 0.03\%, and 0.07\% quarter over quarter for recreation; 0.20\%, 0.26\%, 0.33\%, and 0.46\% quarter over quarter for education and communication; and 4.29\%, 0.48\%, 1.50\%, and 1.39\% quarter over quarter for other goods and services. Over the same five-year period, compound average quarterly inflation was 0.62\% for food and beverages, 0.63\% for housing, -0.05\% for apparel, 0.40\% for transportation, 0.84\% for medical care, 0.46\% for recreation, 0.63\% for education and communication, and 1.34\% for other goods and services.\par

Given this information, how much did your household spend in each of the categories during the past three months, in current U.S. dollars?\par

Return only a valid JSON array with 12 numeric elements, in this order: food, alcoholic beverages, housing, apparel and services, transportation, health care, entertainment, personal care, reading, education, tobacco, and miscellaneous. Your output must follow this exact array structure: [v\_\allowbreak{}1, v\_\allowbreak{}2, ..., v\_\allowbreak{}12]. Replace every v placeholder with a single number. Do not include keys or explanatory text.\par
\paragraph{\fontsize{7.2}{8.55}\selectfont Assistant.} [2750, 130, 3184, 535, 3655, 572, 1171, 72, 66, 0, 0, 14]\par
\endgroup
\end{tcolorbox}

\hypertarget{task:cons_cex_total}{}
\subsection{cons\_\allowbreak{}cex\_\allowbreak{}total}

\textbf{Target.} Total, nondurable, and durable household expenditure over the last three months, in current U.S. dollars. We construct these measures from the expenditure summaries in the CEX, following the definitions in \citet{parker2013stimulus}. Total expenditure equals the survey's total expenditure measure less cash contributions, life and other personal insurance, and contributions to retirement plans, pensions, and Social Security. Nondurable expenditure is the sum of spending on food and alcoholic beverages, utilities, household operations, public transportation, gasoline and motor oil, personal care, tobacco, miscellaneous items, apparel, health care, and reading. Durable expenditure equals total expenditure less nondurable expenditure.\par

\textbf{Predictors.} Spending history (up to three prior three-month periods of total, nondurable, and durable expenditure), before-tax household income over the past 12 months, household composition (household size, number of adults, number of children), respondent demographics and geography (age, sex, race, education, marital status, region, urban or rural status), and core macro context (real GDP growth, unemployment rate, headline CPI inflation, and the effective federal funds rate over the four most recently completed calendar quarters, plus five-year quarterly references).\par

\textbf{No-change baseline.} For each expenditure measure, the no-change prediction is the household's expenditure reported in the immediately preceding interview, conducted three months earlier.\par

\textbf{Sample construction.} Starting from our clean CEX Interview panel, we merge core macro variables from FRED-QD and construct up to three lags of income and expenditure variables. These need to form a continuous history: if a household missed an interview, we set deeper lags to missing as well. Next, we drop rows that are missing one or more of the target variables, and also rows with current or first-history predictors missing (further lags are allowed to remain missing). Next, we require current annual before-tax household income to be at least \$1,000, and current and available lagged total expenditure to be at least \$250. We also require current and available lagged values of total, nondurable, and durable expenditure to be nonnegative. Finally, within each year and calendar quarter, we trim the top and bottom percentiles of current annual income and current and available lagged total expenditure.\par

\begin{tcolorbox}[
  breakable,
  lines before break=8,
  colback=white,
  colframe=black!55,
  boxrule=0.45pt,
  arc=0pt,
  left=7pt,
  right=7pt,
  top=7pt,
  bottom=7pt
]
\begingroup
\setlength{\parindent}{0pt}
\setlength{\parskip}{5pt}
\hyphenpenalty=1800
\exhyphenpenalty=1800
\emergencystretch=0.8em
{\centering\fontsize{7.7}{9.0}\selectfont\bfseries Prompt example\par
\texttt{cons\_cex\_total}\par}
\medskip
{\color{black!25}\hrule height 0.35pt}
\vspace{5pt}
\fontsize{7.2}{8.55}\selectfont\rmfamily
\paragraph{\fontsize{7.2}{8.55}\selectfont System.} You are the head of an American household making decisions and forming expectations about income, spending, and saving.\par
\paragraph{\fontsize{7.2}{8.55}\selectfont User.}
Here is some background information about yourself and your household. You are a 72-year-old white woman, are currently divorced, and have a high school diploma. You currently reside in an urban area in the West. Your household has 1 adult and no children.\par

Here are your household's recent spending and income. Note that the current month is March, and that \$ denotes amounts in U.S. dollars. Between nine and twelve months ago, your total household expenditure was \$2,135, non-durable expenditure was \$1,290, and durable expenditure was \$846. Between six and nine months ago, your total household expenditure was \$22,364, non-durable expenditure was \$1,334, and durable expenditure was \$21,030. Between three and six months ago, your total household expenditure was \$4,527, non-durable expenditure was \$1,360, and durable expenditure was \$3,166. Your before-tax household income over the past 12 months was \$5,826.\par

You are also aware of recent national economic conditions. Over the four most recently completed calendar quarters, listed from oldest to most recent, U.S. real GDP changed by 0.94\%, 0.83\%, 1.33\%, and 1.64\% quarter over quarter; the unemployment rate was 4.30\%, 4.27\%, 4.23\%, and 4.07\%; overall consumer prices changed by 0.37\%, 0.75\%, 0.74\%, and 0.74\% quarter over quarter; and the effective federal funds rate was 4.73\%, 4.75\%, 5.09\%, and 5.31\%. Over the five years ending in the most recently completed quarter, compound average quarterly real GDP growth was 1.02\%, the average unemployment rate was 4.93\%, compound average quarterly CPI inflation was 0.59\%, and the average effective federal funds rate was 5.38\%.\par

Given this information, what were your total household expenditure, non-durable expenditure, and durable expenditure during the past three months, in current U.S. dollars?\par

Return only a valid JSON array with exactly 3 numeric elements, in this order: total\_\allowbreak{}expenditure, non\_\allowbreak{}durable\_\allowbreak{}expenditure, durable\_\allowbreak{}expenditure. Your output must follow this exact array structure: [v\_\allowbreak{}1, v\_\allowbreak{}2, v\_\allowbreak{}3]. Replace every v placeholder with a single number. Do not include keys or explanatory text.\par
\paragraph{\fontsize{7.2}{8.55}\selectfont Assistant.} [4824, 1379, 3446]\par
\endgroup
\end{tcolorbox}

\hypertarget{task:cons_psid_wealth}{}
\subsection{cons\_\allowbreak{}psid\_\allowbreak{}wealth}

\textbf{Target.} Household net worth in the next PSID interview, usually two years after the current interview and sometimes one year later in the early part of the sample. We use the PSID's summary measure of net worth, which combines the household's reported assets, including savings, financial investments, home equity, and other assets, and subtracts its reported debts.\par

\textbf{Predictors.} Household resources (income, net worth, savings, financial investments, home equity, other debt), household history (up to three prior samples of those resource measures and housing tenure), housing tenure (own, rent, or another arrangement), household composition (presence of a spouse or partner, marital status, household size, number of children), respondent demographics and geography (age, sex, race and ethnicity, region, metropolitan status when available), core macro context (real GDP growth, unemployment rate, headline CPI inflation, and the effective federal funds rate over the four most recently completed quarters, plus five-year quarterly references), and financial and housing-market context (house-price growth, S\&P 500 growth, and the 30-year mortgage rate over the same periods).\par

\textbf{No-change baseline.} The no-change prediction is the household's net worth in the current PSID interview.\par

\textbf{Sample construction.} Starting from our clean PSID household panel, we merge national economic, financial-market, and housing-market variables. We link each household to its next consecutive PSID interview. Next, we drop rows that are missing the next-interview net-worth target or any required current predictor: age, sex, race or ethnicity, region, housing tenure, spouse or partner status, marital status, family income, net worth, or the economic, financial-market, and housing-market variables. Current household size, number of children, metropolitan status, the individual wealth components, and all prior household histories are allowed to remain missing. Finally, within the relevant survey year, we trim the top and bottom percentiles of current and available lagged family income and of current, available lagged, and next-interview net worth.\par

\begin{tcolorbox}[
  breakable,
  lines before break=8,
  colback=white,
  colframe=black!55,
  boxrule=0.45pt,
  arc=0pt,
  left=7pt,
  right=7pt,
  top=7pt,
  bottom=7pt
]
\begingroup
\setlength{\parindent}{0pt}
\setlength{\parskip}{5pt}
\hyphenpenalty=1800
\exhyphenpenalty=1800
\emergencystretch=0.8em
{\centering\fontsize{7.7}{9.0}\selectfont\bfseries Prompt example\par
\texttt{cons\_psid\_wealth}\par}
\medskip
{\color{black!25}\hrule height 0.35pt}
\vspace{5pt}
\fontsize{7.2}{8.55}\selectfont\rmfamily
\paragraph{\fontsize{7.2}{8.55}\selectfont System.} You are the head of an American household making decisions about income, saving, housing, and household wealth.\par
\paragraph{\fontsize{7.2}{8.55}\selectfont User.}
Here is some background information about yourself and your household. You are a 57-year-old White woman. You live in the South, in a metropolitan area. You do not have a spouse or partner in your household and are not legally married. Your household has 1 person, no children.\par

Here is your household's financial situation now and in past years. Note that \$ denotes amounts in U.S. dollars. Your current household income is \$60,704, and your total net worth is \$1,000. You have \$0 in savings, \$0 in financial investments, \$0 in home equity, and \$0 in other debt. You currently rent your home. Six years ago, your household income was \$52,656, your total net worth was \$17,000, your savings were \$4,000, your financial investments were \$0, your home equity was \$0, your other debt was \$0, and your household rented its home. Four years ago, your household income was \$18,428, your total net worth was \$10,000, your savings were \$0, your financial investments were \$0, your home equity was \$0, your other debt was \$0, and your household rented its home. Two years ago, your household income was \$55,080, your total net worth was \$4,000, your savings were \$4,000, your financial investments were \$0, your home equity was \$0, your other debt was \$3,000, and your household rented its home.\par

You are also aware of recent national economic conditions. Over the four most recently completed calendar quarters, listed from oldest to most recent, U.S. real GDP changed by -0.35\%, 1.29\%, 1.22\%, and 0.51\% quarter over quarter; the unemployment rate was 6.67\%, 6.20\%, 6.07\%, and 5.70\%; overall consumer prices changed by 0.62\%, 0.53\%, 0.26\%, and -0.25\% quarter over quarter; and the effective federal funds rate was 0.07\%, 0.09\%, 0.09\%, and 0.10\%. Over the five years ending in the most recently completed quarter, compound average quarterly real GDP growth was 0.57\%, the average unemployment rate was 8.03\%, compound average quarterly CPI inflation was 0.44\%, and the average effective federal funds rate was 0.12\%. You are also aware of recent financial and housing-market conditions. Over the same four completed quarters, listed from oldest to most recent, U.S. house prices changed by 0.49\%, 1.50\%, 1.06\%, and 0.54\% quarter over quarter; the S\&P 500 changed by 3.61\%, 3.60\%, 3.98\%, and 1.83\% quarter over quarter; and the average 30-year mortgage rate was 4.36\%, 4.23\%, 4.14\%, and 3.97\%. Over the same five-year period, compound average quarterly house-price growth was -0.15\%, compound average quarterly S\&P 500 growth was 3.12\%, and the average 30-year mortgage rate was 4.19\%.\par

Given this information, in two years, what will your household's total net worth be, in current U.S. dollars?\par

Return only a valid JSON array containing exactly one integer, with no explanatory text. Your output must follow this exact array structure: [v\_\allowbreak{}1]. Replace every v placeholder with a single number.\par
\paragraph{\fontsize{7.2}{8.55}\selectfont Assistant.} [2000]\par
\endgroup
\end{tcolorbox}

\hypertarget{task:cons_sce_growth}{}
\subsection{cons\_\allowbreak{}sce\_\allowbreak{}growth}

\textbf{Target.} The percentage change in the household's current monthly spending relative to 12 months earlier. The SCE first asks whether monthly spending is higher, lower, or unchanged and then asks by what percentage it changed. We combine the direction and size of the reported change so that increases are positive and decreases are negative.\par

\textbf{Predictors.} Household finances and expectations (current financial situation, expected financial situation one year ahead, one-year inflation expectation, longer-run inflation expectation, spending relative to budget when available), work situation (current work-status indicators), respondent demographics, resources, and geography (age, numeracy, region, education, household before-tax income bracket), and core macro context (real GDP growth, unemployment rate, headline CPI inflation, and the effective federal funds rate over the four most recently completed quarters, plus five-year quarterly references).\par

\textbf{No-change baseline.} The no-change prediction is no change in the household's monthly spending relative to 12 months earlier.\par

\textbf{Sample construction.} We merge the SCE Household Spending module with respondent information from the SCE Core survey and national economic variables from FRED-QD. Next, we drop rows that are missing either part of the spending-growth target or any current predictor. We then require respondents to be between 18 and 100 years old. Finally, within each survey month, we trim the top and bottom percentiles of the reported spending change.\par

\begin{tcolorbox}[
  breakable,
  lines before break=8,
  colback=white,
  colframe=black!55,
  boxrule=0.45pt,
  arc=0pt,
  left=7pt,
  right=7pt,
  top=7pt,
  bottom=7pt
]
\begingroup
\setlength{\parindent}{0pt}
\setlength{\parskip}{5pt}
\hyphenpenalty=1800
\exhyphenpenalty=1800
\emergencystretch=0.8em
{\centering\fontsize{7.7}{9.0}\selectfont\bfseries Prompt example\par
\texttt{cons\_sce\_growth}\par}
\medskip
{\color{black!25}\hrule height 0.35pt}
\vspace{5pt}
\fontsize{7.2}{8.55}\selectfont\rmfamily
\paragraph{\fontsize{7.2}{8.55}\selectfont System.} You are the head of an American household making decisions and forming expectations about income, spending, and saving.\par
\paragraph{\fontsize{7.2}{8.55}\selectfont User.}
Here is some background information about yourself and your household. You are a 36-year-old, have a college education or more, live in the West, and have a high numeracy score.\par

Here are your household finances, work situation, and expectations. Note that the current month is August, and that \$ denotes amounts in U.S. dollars. Your household's total pre-tax income over the past 12 months was between \$30,000 and \$39,999. Your household is somewhat better off than it was 12 months ago, and you expect your household finances to be about the same 12 months from now. You expect inflation to be 1.1\% over the next 12 months and 1.0\% over the next three years. You are working full-time, and your household currently uses a budget.\par

You are also aware of recent national economic conditions. Over the four most recently completed calendar quarters, listed from oldest to most recent, U.S. real GDP changed by 1.22\%, 0.51\%, 0.90\%, and 0.62\% quarter over quarter; the unemployment rate was 6.07\%, 5.70\%, 5.53\%, and 5.43\%; overall consumer prices changed by 0.26\%, -0.25\%, -0.65\%, and 0.68\% quarter over quarter; and the effective federal funds rate was 0.09\%, 0.10\%, 0.11\%, and 0.12\%. Over the five years ending in the most recently completed quarter, compound average quarterly real GDP growth was 0.58\%, the average unemployment rate was 7.60\%, compound average quarterly CPI inflation was 0.43\%, and the average effective federal funds rate was 0.12\%.\par

Compared with 12 months ago, by what percentage has your household's current monthly spending changed? Use a positive number if spending is higher now and a negative number if spending is lower now.\par

Return only a valid JSON array containing exactly one number. Your output must follow this exact array structure: [v\_\allowbreak{}1]. Replace every v placeholder with a single number. Do not include explanatory text.\par
\paragraph{\fontsize{7.2}{8.55}\selectfont Assistant.} [-2.0]\par
\endgroup
\end{tcolorbox}

\hypertarget{task:cons_cex_rebate01}{}
\subsection{cons\_\allowbreak{}cex\_\allowbreak{}rebate01}

This task is inspired by \citet{johnson2006rebates}.\par

\textbf{Target.} Total, durable, and nondurable household expenditure over the last three months, in current U.S. dollars. We construct these measures from the expenditure summaries in the CEX, following the definitions in \citet{parker2013stimulus}. Total expenditure equals the survey's total expenditure measure less cash contributions, life and other personal insurance, and contributions to retirement plans, pensions, and Social Security. Nondurable expenditure sums food and alcoholic beverages, utilities, household operations, public transportation, gasoline and motor oil, personal care, tobacco, miscellaneous items, apparel, health care, and reading. Durable expenditure equals total expenditure less nondurable expenditure.\par

\textbf{Predictors.} Spending history (up to three prior three-month periods of total, nondurable, and durable expenditure), before-tax household income over the past 12 months, household composition (household size, number of adults, number of children), respondent demographics and geography (age, sex, race, education, marital status, region, urban or rural status), core macro context (real GDP growth, unemployment rate, headline CPI inflation, and the effective federal funds rate over the four most recently completed quarters, plus five-year quarterly references), and amount received under the 2001 federal income-tax rebates (the amount received during the current three-month period and during prior three-month periods, including zero amounts).\par

\textbf{No-change baseline.} For each expenditure measure, the no-change prediction is the household's expenditure in the immediately preceding three-month period.\par

\textbf{Sample construction.} Starting from our clean CEX Interview panel, we merge the 2001 tax-rebate supplement and national economic variables from FRED-QD. We keep interviews conducted from January 2001 through March 2002 while the rebate questions were fielded, and construct up to three continuous lags of income, expenditure, and assigned rebate amounts. Next, we drop rows that are missing one or more target variables, any current demographic, geographic, income, or macroeconomic predictor, any first-history expenditure, or the first-history assigned rebate. The current assigned rebate and the second and third histories of expenditures and rebates are allowed to remain missing. We then require current annual before-tax household income to be at least \$1,000, current and available lagged total expenditure to be at least \$250, and current and available lagged values of total, nondurable, and durable expenditure to be nonnegative. Finally, within each year and calendar quarter, we trim the top and bottom percentiles of current annual income and current and available lagged total expenditure.\par

\begin{tcolorbox}[
  breakable,
  lines before break=8,
  colback=white,
  colframe=black!55,
  boxrule=0.45pt,
  arc=0pt,
  left=7pt,
  right=7pt,
  top=7pt,
  bottom=7pt
]
\begingroup
\setlength{\parindent}{0pt}
\setlength{\parskip}{5pt}
\hyphenpenalty=1800
\exhyphenpenalty=1800
\emergencystretch=0.8em
{\centering\fontsize{7.7}{9.0}\selectfont\bfseries Prompt example\par
\texttt{cons\_cex\_rebate01}\par}
\medskip
{\color{black!25}\hrule height 0.35pt}
\vspace{5pt}
\fontsize{7.2}{8.55}\selectfont\rmfamily
\paragraph{\fontsize{7.2}{8.55}\selectfont System.} You are the head of an American household making decisions and forming expectations about income, spending, and saving.\par
\paragraph{\fontsize{7.2}{8.55}\selectfont User.}
Here is some background information about yourself and your household. You are a 48-year-old white woman, are currently widowed, and have a bachelor's degree. You currently reside in an urban area in the South. Your household has 1 adult and no children.\par

Here are your household's recent spending and income. Note that the current month is December, and that \$ denotes amounts in U.S. dollars. Between nine and twelve months ago, your total household expenditure was \$4,797, non-durable expenditure was \$2,407, and durable expenditure was \$2,390. Between six and nine months ago, your total household expenditure was \$3,966, non-durable expenditure was \$2,151, and durable expenditure was \$1,815. Between three and six months ago, your total household expenditure was \$4,046, non-durable expenditure was \$2,229, and durable expenditure was \$1,817. Your before-tax household income over the past 12 months was \$28,500.\par

You are also aware of recent national economic conditions. Over the four most recently completed calendar quarters, listed from oldest to most recent, U.S. real GDP changed by 0.60\%, -0.33\%, 0.62\%, and -0.40\% quarter over quarter; the unemployment rate was 3.90\%, 4.23\%, 4.40\%, and 4.83\%; overall consumer prices changed by 0.71\%, 0.96\%, 0.70\%, and 0.28\% quarter over quarter; and the effective federal funds rate was 6.47\%, 5.59\%, 4.33\%, and 3.50\%. Over the five years ending in the most recently completed quarter, compound average quarterly real GDP growth was 0.89\%, the average unemployment rate was 4.46\%, compound average quarterly CPI inflation was 0.61\%, and the average effective federal funds rate was 5.34\%.\par

The federal government has recently implemented a policy of sending tax rebates to households who satisfy certain criteria. Over the past three months, your household received a total of \$300 as part of the program. Before this, the tax-rebate amounts recorded for your household were \$300 between three and six months ago, \$0 between six and nine months ago, and \$0 between nine and twelve months ago.\par

Given this information, what were your total household expenditure, non-durable expenditure, and durable expenditure during the past three months, in current U.S. dollars?\par

Return only a valid JSON array with exactly 3 numeric elements, in this order: total\_\allowbreak{}expenditure, non\_\allowbreak{}durable\_\allowbreak{}expenditure, durable\_\allowbreak{}expenditure. Your output must follow this exact array structure: [v\_\allowbreak{}1, v\_\allowbreak{}2, v\_\allowbreak{}3]. Replace every v placeholder with a single number. Do not include keys or explanatory text.\par
\paragraph{\fontsize{7.2}{8.55}\selectfont Assistant.} [4643, 2768, 1875]\par
\endgroup
\end{tcolorbox}

\hypertarget{task:cons_cex_sspay}{}
\subsection{cons\_\allowbreak{}cex\_\allowbreak{}sspay}

This task is inspired by \citet{stephens2003third}.\par

\textbf{Target.} Total, food-at-home, and food-away-from-home household expenditure on a given day, in current U.S. dollars. We construct these measures by assigning each purchase recorded in the CEX Diary to its reported date and summing purchases within the three categories, as defined in \citet{stephens2003third}, for that day. A day with no recorded spending in a category is assigned zero.\par

\textbf{Predictors.} Spending history (total, food-at-home, and food-away-from-home expenditure on each of the previous seven calendar days), household resources (annual household income and Social Security or Railroad Retirement income), household composition and respondent demographics, calendar timing (weekday, day of month, month, and the household's position relative to its scheduled Social Security payment date), and core macro context (national economic conditions over the most recently completed quarters).\par

\textbf{No-change baseline.} For each expenditure measure, the no-change prediction is the household's expenditure on the immediately preceding calendar day.\par

\textbf{Sample construction.} Starting from our clean daily CEX Diary panel for 1986--1996, we add scheduled Social Security payment dates and national economic variables, and construct spending histories for the previous seven calendar days. We keep non-January days in a seven-day window around a scheduled payment for households in which the reference person or spouse reports Social Security retirement or disability income. We require both weeks of the household's two-week diary to be complete, with valid nonoverlapping dates and usable dated purchases. Next, we drop rows that are missing one or more target variables, any current predictor, or any of the seven daily spending histories. No target or predictor shown to the model is allowed to remain missing. We then require positive household income and Social Security income, Social Security income to account for between 0 and 100 percent of household income, all displayed spending to be nonnegative, and food-at-home plus food-away-from-home spending not to exceed total spending on any displayed day. Finally, within each calendar year, we remove samples above the 99th percentile of annual household income or of total daily spending on the current or any of the previous seven days.\par

\begin{tcolorbox}[
  breakable,
  lines before break=8,
  colback=white,
  colframe=black!55,
  boxrule=0.45pt,
  arc=0pt,
  left=7pt,
  right=7pt,
  top=7pt,
  bottom=7pt
]
\begingroup
\setlength{\parindent}{0pt}
\setlength{\parskip}{5pt}
\hyphenpenalty=1800
\exhyphenpenalty=1800
\emergencystretch=0.8em
{\centering\fontsize{7.7}{9.0}\selectfont\bfseries Prompt example\par
\texttt{cons\_cex\_sspay}\par}
\medskip
{\color{black!25}\hrule height 0.35pt}
\vspace{5pt}
\fontsize{7.2}{8.55}\selectfont\rmfamily
\paragraph{\fontsize{7.2}{8.55}\selectfont System.} You are the head of an American household making decisions and forming expectations about income, spending, and saving.\par
\paragraph{\fontsize{7.2}{8.55}\selectfont User.}
Here is some background information about yourself and your household. You are an 82-year-old white man, are widowed, have an elementary school education, and are not currently working. You live in an urban area in the South. Your household consists of 1 adult and no children. Your household owns its home without a mortgage.\par

Your household reported about \$9,886 in income before tax over the past 12 months. Of this, about \$6,186, or 62.6\%, came from Social Security and Railroad Retirement. These are annual figures and do not tell you the size of any particular monthly payment.\par

Here is your household's spending over the last seven days, listed from oldest to most recent. Note that today is Monday, 27th January, and that \$ denotes amounts in U.S. dollars.\par

- 7 days ago: \$7.41 in total, including \$7.41 on food at home and \$0.00 on food away from home.\par
- 6 days ago: \$194.20 in total, including \$20.45 on food at home and \$0.00 on food away from home.\par
- 5 days ago: \$0.00 in total, including \$0.00 on food at home and \$0.00 on food away from home.\par
- 4 days ago: \$5.85 in total, including \$4.39 on food at home and \$0.00 on food away from home.\par
- 3 days ago: \$9.42 in total, including \$9.42 on food at home and \$0.00 on food away from home.\par
- 2 days ago: \$0.00 in total, including \$0.00 on food at home and \$0.00 on food away from home.\par
- 1 day ago: \$54.74 in total, including \$10.74 on food at home and \$0.00 on food away from home.\par

You are also aware of recent national economic conditions. Over the four most recently completed calendar quarters, listed from oldest to most recent, U.S. real GDP changed by 0.97\%, 0.88\%, 1.53\%, and 0.74\% quarter over quarter; the unemployment rate was 7.23\%, 7.30\%, 7.20\%, and 7.03\%; overall consumer prices changed by 0.92\%, 0.91\%, 0.62\%, and 1.02\% quarter over quarter; and the effective federal funds rate was 8.48\%, 7.92\%, 7.90\%, and 8.10\%. Over the five years ending in the most recently completed quarter, compound average quarterly real GDP growth was 0.85\%, the average unemployment rate was 8.32\%, compound average quarterly CPI inflation was 1.22\%, and the average effective federal funds rate was 11.21\%.\par

Social Security benefits are normally paid on the third day of each month. If the third falls on a weekend or federal holiday, payment is made on the preceding business day.\par

Given this information, how much will your household spend today in total, on food at home, and on food away from home?\par

Return only a valid JSON array with exactly 3 numeric elements, in this order: total\_\allowbreak{}expenditure, food\_\allowbreak{}at\_\allowbreak{}home, food\_\allowbreak{}away\_\allowbreak{}from\_\allowbreak{}home. Use 0 when there is no spending in a category. Your output must follow this exact array structure: [v\_\allowbreak{}1, v\_\allowbreak{}2, v\_\allowbreak{}3]. Replace every v placeholder with a single number. Do not include keys or explanatory text.\par
\paragraph{\fontsize{7.2}{8.55}\selectfont Assistant.} [0.0, 0.0, 0.0]\par
\endgroup
\end{tcolorbox}

\hypertarget{task:cons_cex_stimulus08}{}
\subsection{cons\_\allowbreak{}cex\_\allowbreak{}stimulus08}

This task is inspired by \citet{parker2013stimulus}.\par

\textbf{Target.} Total, durable, and nondurable household expenditure over the last three months, in current U.S. dollars. We construct these measures from the expenditure summaries in the CEX, following the definitions in \citet{parker2013stimulus}. Total expenditure equals the survey's total expenditure measure less cash contributions, life and other personal insurance, and contributions to retirement plans, pensions, and Social Security. Nondurable expenditure sums food and alcoholic beverages, utilities, household operations, public transportation, gasoline and motor oil, personal care, tobacco, miscellaneous items, apparel, health care, and reading. Durable expenditure equals total expenditure less nondurable expenditure. \par

\textbf{Predictors.} Spending history (up to three prior three-month periods of total, nondurable, and durable expenditure), before-tax household income over the past 12 months, household composition (household size, number of adults, number of children), respondent demographics and geography (age, sex, race, education, marital status, region, urban or rural status), core macro context (real GDP growth, unemployment rate, headline CPI inflation, and the effective federal funds rate over the four most recently completed quarters, plus five-year quarterly references), and amount received under the 2008 economic stimulus payment program (the amount received during the current three-month period and during prior periods, including zero amounts).\par

\textbf{No-change baseline.} For each expenditure measure, the no-change prediction is the household's expenditure in the immediately preceding three-month period.\par

\textbf{Sample construction.} Starting from our clean CEX Interview panel, we merge the 2008 stimulus-payment supplement and national economic variables from FRED-QD. We keep interviews conducted from September 2007 through March 2009 and require the household to have been interviewed while the stimulus questions were fielded, from June 2008 through March 2009. We construct up to three continuous lags of income, expenditure, and assigned payment amounts. Next, we drop rows that are missing one or more target variables, any current demographic, geographic, income, or macroeconomic predictor, or the first lag of the expenditure fields. The current and first lag of the assigned payment amounts are set to zero when no payment is assigned. Second and third lags of expenditures and assigned payments are allowed to remain missing. We then require current annual before-tax household income to be at least \$1,000, current and available lagged total expenditure to be at least \$250, and current and available lagged values of total, nondurable, and durable expenditure to be nonnegative. Finally, within each year and calendar quarter, we trim the top and bottom percentiles of current annual income and current and available lagged total expenditure.\par

\begin{tcolorbox}[
  breakable,
  lines before break=8,
  colback=white,
  colframe=black!55,
  boxrule=0.45pt,
  arc=0pt,
  left=7pt,
  right=7pt,
  top=7pt,
  bottom=7pt
]
\begingroup
\setlength{\parindent}{0pt}
\setlength{\parskip}{5pt}
\hyphenpenalty=1800
\exhyphenpenalty=1800
\emergencystretch=0.8em
{\centering\fontsize{7.7}{9.0}\selectfont\bfseries Prompt example\par
\texttt{cons\_cex\_stimulus08}\par}
\medskip
{\color{black!25}\hrule height 0.35pt}
\vspace{5pt}
\fontsize{7.2}{8.55}\selectfont\rmfamily
\paragraph{\fontsize{7.2}{8.55}\selectfont System.} You are the head of an American household making decisions and forming expectations about income, spending, and saving.\par
\paragraph{\fontsize{7.2}{8.55}\selectfont User.}
Here is some background information about yourself and your household. You are a 66-year-old white woman, are currently divorced, and have a graduate or professional degree. You currently reside in a rural area in the Midwest. Your household has 1 adult and no children.\par

Here are your household's recent spending and income. Note that the current month is September, and that \$ denotes amounts in U.S. dollars. Between nine and twelve months ago, your total household expenditure was \$3,896, non-durable expenditure was \$1,616, and durable expenditure was \$2,280. Between six and nine months ago, your total household expenditure was \$5,464, non-durable expenditure was \$3,404, and durable expenditure was \$2,060. Between three and six months ago, your total household expenditure was \$4,994, non-durable expenditure was \$2,326, and durable expenditure was \$2,668. Your before-tax household income over the past 12 months was \$41,273.\par

You are also aware of recent national economic conditions. Over the four most recently completed calendar quarters, listed from oldest to most recent, U.S. real GDP changed by 0.58\%, 0.63\%, -0.43\%, and 0.60\% quarter over quarter; the unemployment rate was 4.67\%, 4.80\%, 5.00\%, and 5.33\%; overall consumer prices changed by 0.63\%, 1.23\%, 1.08\%, and 1.30\% quarter over quarter; and the effective federal funds rate was 5.07\%, 4.50\%, 3.18\%, and 2.09\%. Over the five years ending in the most recently completed quarter, compound average quarterly real GDP growth was 0.70\%, the average unemployment rate was 5.08\%, compound average quarterly CPI inflation was 0.82\%, and the average effective federal funds rate was 3.27\%.\par

The federal government has recently implemented a policy of sending stimulus payments to households who satisfy certain criteria. Over the past three months, your household received a total of \$600 as part of the program. Before this, the stimulus-payment amounts recorded for your household were \$600 between three and six months ago, \$0 between six and nine months ago, and \$0 between nine and twelve months ago.\par

Given this information, what were your total household expenditure, non-durable expenditure, and durable expenditure during the past three months, in current U.S. dollars?\par

Return only a valid JSON array with exactly 3 numeric elements, in this order: total\_\allowbreak{}expenditure, non\_\allowbreak{}durable\_\allowbreak{}expenditure, durable\_\allowbreak{}expenditure. Your output must follow this exact array structure: [v\_\allowbreak{}1, v\_\allowbreak{}2, v\_\allowbreak{}3]. Replace every v placeholder with a single number. Do not include keys or explanatory text.\par
\paragraph{\fontsize{7.2}{8.55}\selectfont Assistant.} [3802, 2193, 1609]\par
\endgroup
\end{tcolorbox}

\hypertarget{task:cons_psid_jobloss}{}
\subsection{cons\_\allowbreak{}psid\_\allowbreak{}jobloss}

\textbf{Target.} The household's annual cash expenditure on food in the calendar year after a job loss, in current U.S. dollars. We construct this measure from the PSID questions on food purchased for use at home and food purchased away from home, excluding food paid for with public assistance.\par

\textbf{Predictors.} Household food spending history (up to three prior annual samples), household resources (family income and each adult's labor earnings, annual hours, and employment status), job-loss circumstances (which adult lost a job and whether the loss was caused by a layoff, firing, plant closure, or employer move), household composition and respondent demographics (age, sex, education, race, marital status, presence of a spouse or partner, household size, number of children), and core macro context (national economic conditions before and during the job-loss year). \par

\textbf{No-change baseline.} The no-change prediction is the household's most recently observed annual cash food expenditure before the job-loss year.\par

\textbf{Sample construction.} Starting from PSID household records for 1968--1992, we combine the reports of the two adults in each household, construct cash food expenditure, and link each household to its next interview and to as many as three earlier interviews. A job-loss sample requires a layoff, firing, plant closure, or employer move after the affected adult had been working; multiple losses reported by the same household for the same period are combined. A control sample requires every current adult to have answered the job-loss questions and no adult to report a current or earlier qualifying loss, and we retain at most one control sample per household. Next, we drop rows that are missing the positive cash-food target, current predictors or the first lag of spending before the job loss event. Current education, race or ethnicity, region, housing tenure, prior job-loss details, and history records beyond the mandatory one are allowed to remain missing. We keep households whose adults are between 25 and 65 years old and whose spouse or partner structure is unchanged across the relevant interviews. Finally, within each survey year, we trim the top and bottom percentiles of every family-income value shown in the household's history.\par

\begin{tcolorbox}[
  breakable,
  lines before break=8,
  colback=white,
  colframe=black!55,
  boxrule=0.45pt,
  arc=0pt,
  left=7pt,
  right=7pt,
  top=7pt,
  bottom=7pt
]
\begingroup
\setlength{\parindent}{0pt}
\setlength{\parskip}{5pt}
\hyphenpenalty=1800
\exhyphenpenalty=1800
\emergencystretch=0.8em
{\centering\fontsize{7.7}{9.0}\selectfont\bfseries Prompt example\par
\texttt{cons\_psid\_jobloss}\par}
\medskip
{\color{black!25}\hrule height 0.35pt}
\vspace{5pt}
\fontsize{7.2}{8.55}\selectfont\rmfamily
\paragraph{\fontsize{7.2}{8.55}\selectfont System.} You are the head of an American household making decisions about food spending, work, income, and household finances.\par
\paragraph{\fontsize{7.2}{8.55}\selectfont User.}
Here is some background information about yourself and your household. You are a 52-year-old Black woman. You live in the South. You do not have a spouse or partner in your household. Your household has 1 person, no children. You completed 14 years of schooling. You currently own your home.\par

Here is your household's work, income, and food-spending history. Note that \$ denotes amounts in U.S. dollars. Three years ago, you worked 540 hours in total and had \$650 in annual labor earnings. Your total family income was \$1,250. Two years ago, you worked 233 hours in total and had \$350 in annual labor earnings. Your total family income was \$1,350. Your household's annual food spending was \$520, including food at home and food away from home.\par

You are also aware of recent national economic conditions. Over the four most recently completed calendar quarters, listed from oldest to most recent, U.S. real GDP changed by -0.15\%, 0.14\%, 0.92\%, and -1.07\% quarter over quarter; the unemployment rate was 4.17\%, 4.77\%, 5.17\%, and 5.83\%; overall consumer prices changed by 1.60\%, 1.40\%, 1.04\%, and 1.45\% quarter over quarter; and the effective federal funds rate was 8.57\%, 7.89\%, 6.71\%, and 5.57\%. Over the five years ending in the most recently completed quarter, compound average quarterly real GDP growth was 0.69\%, the average unemployment rate was 3.93\%, compound average quarterly CPI inflation was 1.11\%, and the average effective federal funds rate was 6.08\%.\par

You did not lose your job during the current calendar year.\par

Given this information, in one year, what will your household's annual food spending be, including food at home and food away from home?\par

Return only a valid JSON array with exactly one number in current U.S. dollars. Your output must follow this exact array structure: [v\_\allowbreak{}1]. Replace v\_\allowbreak{}1 with one number. Do not include a dollar sign, commas, keys, or explanatory text.\par
\paragraph{\fontsize{7.2}{8.55}\selectfont Assistant.} [312.0]\par
\endgroup
\end{tcolorbox}

\hypertarget{task:cons_sce_shock}{}
\subsection{cons\_\allowbreak{}sce\_\allowbreak{}shock}

\textbf{Target.} How the household says it would adjust spending under two hypothetical income changes: a permanent 10 percent increase in household income and a permanent 10 percent decrease. For the increase, the SCE asks the respondent to allocate the gain among spending or donating, saving or investing, and paying down debt. For the decrease, it asks the respondent to allocate the loss among reducing spending, reducing saving, and increasing borrowing. The task predicts the share of the gain allocated to spending or donating and the share of the loss covered by reducing spending.\par

\textbf{Predictors.} Spending and budgeting (current spending growth relative to 12 months earlier, spending relative to budget when available), household finances and expectations (current financial situation, expected financial situation one year ahead, one-year and longer-run inflation expectations), work situation (current work-status indicators), respondent demographics, resources, and geography (age, education, household before-tax income bracket, region, numeracy), and core macro context (real GDP growth, unemployment rate, headline CPI inflation, and the effective federal funds rate over the four most recently completed quarters, plus five-year quarterly references). Every respondent considers the same permanent 10 percent income increase and decrease.\par

\textbf{No-change baseline.} The no-change prediction is that the household assigns none of either income change to spending.\par

\textbf{Sample construction.} We merge the SCE Household Spending module with respondent information from the SCE Core survey and national economic variables from FRED-QD. Next, we drop rows that are missing one or more of the six allocation answers or any required demographic, income, or macroeconomic predictor. Current spending and budgeting, household-finance, inflation-expectation, and work-status predictors are allowed to remain missing. We then require each allocation to lie between 0 and 100 percent, the three allocations within each scenario to sum to between 99 and 101 percent, and the respondent to be between 18 and 100 years old.\par

\begin{tcolorbox}[
  breakable,
  lines before break=8,
  colback=white,
  colframe=black!55,
  boxrule=0.45pt,
  arc=0pt,
  left=7pt,
  right=7pt,
  top=7pt,
  bottom=7pt
]
\begingroup
\setlength{\parindent}{0pt}
\setlength{\parskip}{5pt}
\hyphenpenalty=1800
\exhyphenpenalty=1800
\emergencystretch=0.8em
{\centering\fontsize{7.7}{9.0}\selectfont\bfseries Prompt example\par
\texttt{cons\_sce\_shock}\par}
\medskip
{\color{black!25}\hrule height 0.35pt}
\vspace{5pt}
\fontsize{7.2}{8.55}\selectfont\rmfamily
\paragraph{\fontsize{7.2}{8.55}\selectfont System.} You are the head of an American household making decisions and forming expectations about income, spending, and saving.\par
\paragraph{\fontsize{7.2}{8.55}\selectfont User.}
Here is some background information about yourself and your household. You are a 36-year-old, have a college education or more, live in the West, and have a high numeracy score.\par

Here are your household finances, work situation, and expectations. Note that the current month is August, and that \$ denotes amounts in U.S. dollars. Your household's total pre-tax income over the past 12 months was between \$30,000 and \$39,999. Your household is somewhat better off than it was 12 months ago, and you expect your household finances to be about the same 12 months from now. You expect inflation to be 1.1\% over the next 12 months and 1.0\% over the next three years. Your monthly household spending is 2.0 percent lower than 12 months ago, your household currently uses a budget, and you are working full-time.\par

You are also aware of recent national economic conditions. Over the four most recently completed calendar quarters, listed from oldest to most recent, U.S. real GDP changed by 1.22\%, 0.51\%, 0.90\%, and 0.62\% quarter over quarter; the unemployment rate was 6.07\%, 5.70\%, 5.53\%, and 5.43\%; overall consumer prices changed by 0.26\%, -0.25\%, -0.65\%, and 0.68\% quarter over quarter; and the effective federal funds rate was 0.09\%, 0.10\%, 0.11\%, and 0.12\%. Over the five years ending in the most recently completed quarter, compound average quarterly real GDP growth was 0.58\%, the average unemployment rate was 7.60\%, compound average quarterly CPI inflation was 0.43\%, and the average effective federal funds rate was 0.12\%.\par

For this question, suppose your household income changes unexpectedly next year. In one case, your household income rises by 10 percent. In the other case, your household income falls by 10 percent.\par

How much of the income gain would you spend or donate, and how much of the income loss would you cover by reducing spending? Give each answer as a number from 0 to 100, not as a fraction.\par

Return only a valid JSON array with exactly two numeric elements, in this order: percent of the income gain spent or donated; percent of the income loss covered by reducing spending. Your output must follow this exact array structure: [v\_\allowbreak{}1, v\_\allowbreak{}2]. Replace every v placeholder with a single number. Do not include keys or explanatory text.\par
\paragraph{\fontsize{7.2}{8.55}\selectfont Assistant.} [1.0, 70.0]\par
\endgroup
\end{tcolorbox}

\hypertarget{task:income_mich_finance}{}
\subsection{income\_\allowbreak{}mich\_\allowbreak{}finance}

\textbf{Target.} The expected percentage change in the household's income over the next 12 months. The Michigan Survey first asks whether family income is expected to increase, decrease, or remain unchanged and then asks by what percentage it is expected to change. We combine the direction and size so that expected increases are positive and expected decreases are negative.\par

\textbf{Predictors.} Household and national assessments (personal finances compared with one year earlier, national business conditions compared with one year earlier), household resources (current annual income), household composition and respondent demographics (age, sex, education, marital or partner status, number of adults, number of children), geography (region), and core macro context (real GDP growth, unemployment rate, headline CPI inflation, and the effective federal funds rate over the four most recently completed quarters, plus five-year quarterly references).\par

\textbf{No-change baseline.} The no-change prediction is no change in household income over the next 12 months.\par

\textbf{Sample construction.} Starting from the Michigan Survey, we merge national economic variables from FRED-QD. Next, we drop rows that are missing either part of the income-growth target or any current predictor. No target or predictor shown to the model is allowed to remain missing. Finally, within each survey month, we trim the top and bottom percentiles of expected income growth and annual household income.\par

\begin{tcolorbox}[
  breakable,
  lines before break=8,
  colback=white,
  colframe=black!55,
  boxrule=0.45pt,
  arc=0pt,
  left=7pt,
  right=7pt,
  top=7pt,
  bottom=7pt
]
\begingroup
\setlength{\parindent}{0pt}
\setlength{\parskip}{5pt}
\hyphenpenalty=1800
\exhyphenpenalty=1800
\emergencystretch=0.8em
{\centering\fontsize{7.7}{9.0}\selectfont\bfseries Prompt example\par
\texttt{income\_mich\_finance}\par}
\medskip
{\color{black!25}\hrule height 0.35pt}
\vspace{5pt}
\fontsize{7.2}{8.55}\selectfont\rmfamily
\paragraph{\fontsize{7.2}{8.55}\selectfont System.} You are an American adult forming expectations about economic conditions and your own finances.\par
\paragraph{\fontsize{7.2}{8.55}\selectfont User.}
Here is some background information about yourself and your household. You are an 18-year-old man living in the North Central United States. You have never been married, and your highest completed education is a high school diploma. Your household has 2 adults and no children.\par

Here are your current views on your finances and the economy. Note that the current month is February, and that \$ denotes amounts in U.S. dollars. Your current household income is \$22,500 per year. You report that your personal finances are better than they were a year ago, while national business conditions are worse than they were a year ago.\par

You are also aware of recent national economic conditions. Over the four most recently completed calendar quarters, listed from oldest to most recent, U.S. real GDP changed by 1.19\%, 1.94\%, 1.80\%, and 0.00\% quarter over quarter; the unemployment rate was 7.50\%, 7.13\%, 6.90\%, and 6.67\%; overall consumer prices changed by 1.83\%, 1.75\%, 1.38\%, and 1.47\% quarter over quarter; and the effective federal funds rate was 4.66\%, 5.16\%, 5.82\%, and 6.51\%. Over the five years ending in the most recently completed quarter, compound average quarterly real GDP growth was 0.68\%, the average unemployment rate was 6.74\%, compound average quarterly CPI inflation was 1.92\%, and the average effective federal funds rate was 7.13\%.\par

By what percentage do you expect your income to change over the next 12 months? Use a negative number if you expect your income to be lower, a positive number if you expect it to be higher, and 0 if you expect it to be about the same.\par

Return only a valid JSON array containing exactly one number. Your output must follow this exact array structure: [v\_\allowbreak{}1]. Replace every v placeholder with a single number. Do not include explanatory text.\par
\paragraph{\fontsize{7.2}{8.55}\selectfont Assistant.} [3]\par
\endgroup
\end{tcolorbox}

\hypertarget{task:income_psid_earnings}{}
\subsection{income\_\allowbreak{}psid\_\allowbreak{}earnings}

\textbf{Target.} The respondent's annual labor earnings two, four, and ten years after the current PSID interview, in current U.S. dollars.\par

\textbf{Predictors.} Current work and earnings (annual labor earnings, employment status, occupation when available), household resources and composition (household income, presence of a spouse or partner, household size, number of children), respondent demographics (age, sex, race and ethnicity, education), geography (region), and core macro context (real GDP growth, unemployment rate, headline CPI inflation, and the effective federal funds rate over the four quarters preceding the survey year, plus five-year quarterly references).\par

\textbf{No-change baseline.} At each horizon, the no-change prediction is the respondent's annual labor earnings in the current interview.\par

\textbf{Sample construction.} Starting from our clean PSID panel, we keep respondents between 20 and 60 years old, harmonize occupation information, and merge national economic variables. We link each current interview to the same respondent's record exactly two, four, and ten years later. Next, we drop rows that are missing one or more of the three earnings targets or any required current predictor: age, sex, race or ethnicity, region, education, employment status, labor earnings, family income, spouse or partner status, or the macroeconomic variables. Current household size, number of children, metropolitan status, and occupation are allowed to remain missing. Finally, within each survey year, we trim the top and bottom percentiles of current, future, and available prior labor earnings and of current and available prior family income.\par

\begin{tcolorbox}[
  breakable,
  lines before break=8,
  colback=white,
  colframe=black!55,
  boxrule=0.45pt,
  arc=0pt,
  left=7pt,
  right=7pt,
  top=7pt,
  bottom=7pt
]
\begingroup
\setlength{\parindent}{0pt}
\setlength{\parskip}{5pt}
\hyphenpenalty=1800
\exhyphenpenalty=1800
\emergencystretch=0.8em
{\centering\fontsize{7.7}{9.0}\selectfont\bfseries Prompt example\par
\texttt{income\_psid\_earnings}\par}
\medskip
{\color{black!25}\hrule height 0.35pt}
\vspace{5pt}
\fontsize{7.2}{8.55}\selectfont\rmfamily
\paragraph{\fontsize{7.2}{8.55}\selectfont System.} You are an American individual or family making education, work, housing, and long-run financial decisions.\par
\paragraph{\fontsize{7.2}{8.55}\selectfont User.}
Here is some background information about yourself and your household. You are a 25-year-old White man. You live in the South. You completed 11 years of schooling and did not complete high school. You have a spouse or partner in your household. Your household has 5 people, including 3 children.\par

Here is your current work situation. Note that \$ denotes amounts in U.S. dollars. You are currently working. Your labor earnings are \$5,800, your household income is \$7,900, and your broad occupation is production, transport, or labor.\par

You are also aware of recent national economic conditions. Over the four most recently completed calendar quarters, listed from oldest to most recent, U.S. real GDP changed by 0.89\%, 0.06\%, 0.95\%, and 0.75\% quarter over quarter; the unemployment rate was 3.83\%, 3.83\%, 3.80\%, and 3.90\%; overall consumer prices changed by 0.25\%, 0.61\%, 1.00\%, and 1.09\% quarter over quarter; and the effective federal funds rate was 4.82\%, 3.99\%, 3.89\%, and 4.17\%. Over the five years ending in the most recently completed quarter, compound average quarterly real GDP growth was 1.27\%, the average unemployment rate was 4.59\%, compound average quarterly CPI inflation was 0.54\%, and the average effective federal funds rate was 4.02\%.\par

What will your labor earnings be in current dollars two years from now, four years from now, and ten years from now?\par

Return only a valid JSON array with exactly 3 numeric elements, in this order: earn\_\allowbreak{}2y, earn\_\allowbreak{}4y, earn\_\allowbreak{}10y. Your output must follow this exact array structure: [v\_\allowbreak{}1, v\_\allowbreak{}2, v\_\allowbreak{}3]. Replace every v placeholder with a single number. Do not include keys or explanatory text.\par
\paragraph{\fontsize{7.2}{8.55}\selectfont Assistant.} [6100, 5700, 13000]\par
\endgroup
\end{tcolorbox}

\hypertarget{task:income_sce_growth}{}
\subsection{income\_\allowbreak{}sce\_\allowbreak{}growth}

\textbf{Target.} The expected percentage change in the household's total income over the next 12 months. The SCE first asks whether total household income is expected to increase, decrease, or remain unchanged and then asks by what percentage it is expected to change. We combine the direction and size so that expected increases are positive and expected decreases are negative.\par

\textbf{Predictors.} Household finances and health (finances compared with one year earlier, expected finances one year ahead, health when available), work and job-search situation (current work-status indicators, number of jobs, type of employment arrangement, whether the respondent is looking for work, and duration of unemployment or time out of work when available), respondent demographics, resources, and geography (age, education, household before-tax income bracket, region, numeracy), and core macro context (real GDP growth, unemployment rate, headline CPI inflation, and the effective federal funds rate over the four most recently completed quarters, plus five-year quarterly references).\par

\textbf{No-change baseline.} The no-change prediction is no change in total household income over the next 12 months.\par

\textbf{Sample construction.} We merge national economic variables from FRED-QD with the SCE Core survey. Next, we drop rows that are missing either part of the income-growth target or any required demographic, income, or macroeconomic predictor. Current household-finance, health, work, and job-search predictors are allowed to remain missing. We then require respondents to be between 18 and 100 years old. Finally, within each survey month, we trim the top and bottom percentiles of expected income growth.\par

\begin{tcolorbox}[
  breakable,
  lines before break=8,
  colback=white,
  colframe=black!55,
  boxrule=0.45pt,
  arc=0pt,
  left=7pt,
  right=7pt,
  top=7pt,
  bottom=7pt
]
\begingroup
\setlength{\parindent}{0pt}
\setlength{\parskip}{5pt}
\hyphenpenalty=1800
\exhyphenpenalty=1800
\emergencystretch=0.8em
{\centering\fontsize{7.7}{9.0}\selectfont\bfseries Prompt example\par
\texttt{income\_sce\_growth}\par}
\medskip
{\color{black!25}\hrule height 0.35pt}
\vspace{5pt}
\fontsize{7.2}{8.55}\selectfont\rmfamily
\paragraph{\fontsize{7.2}{8.55}\selectfont System.} You are an American adult making household decisions and forming expectations about your finances, work, and the economy.\par
\paragraph{\fontsize{7.2}{8.55}\selectfont User.}
Here is some background information about yourself and your household. You are a 54-year-old, have a high-school education, live in the Midwest, and have a high numeracy score.\par

Here are your household finances, work situation, and expectations. Note that the current month is June, and that \$ denotes amounts in U.S. dollars. Your household's total pre-tax income over the past 12 months was less than \$10,000. Your household is much worse off than it was 12 months ago, but you expect your household finances to be about the same 12 months from now.\par

You are also aware of recent national economic conditions. Over the four most recently completed calendar quarters, listed from oldest to most recent, U.S. real GDP changed by 0.45\%, 0.14\%, 0.12\%, and 0.99\% quarter over quarter; the unemployment rate was 8.20\%, 8.03\%, 7.80\%, and 7.73\%; overall consumer prices changed by 0.21\%, 0.45\%, 0.66\%, and 0.40\% quarter over quarter; and the effective federal funds rate was 0.15\%, 0.14\%, 0.16\%, and 0.14\%. Over the five years ending in the most recently completed quarter, compound average quarterly real GDP growth was 0.24\%, the average unemployment rate was 8.48\%, compound average quarterly CPI inflation was 0.44\%, and the average effective federal funds rate was 0.35\%.\par

By what percentage do you expect total household income to change over the next 12 months? Use a negative number for a decline, a positive number for an increase, and 0 for no change.\par

Return only a valid JSON array containing one number. Your output must follow this exact array structure: [v\_\allowbreak{}1]. Replace every v placeholder with a single number. Do not include explanatory text.\par
\paragraph{\fontsize{7.2}{8.55}\selectfont Assistant.} [0.0]\par
\endgroup
\end{tcolorbox}

\hypertarget{task:income_cps_displace}{}
\subsection{income\_\allowbreak{}cps\_\allowbreak{}displace}

\textbf{Target.} Current weekly earnings among workers who lost a job and subsequently found another one. We use the Displaced Worker Supplement's current weekly earnings measure, which records weekly pay directly or standardizes pay reported at another frequency to a weekly amount.\par

\textbf{Predictors.} Respondent demographics and background (age, sex, race and ethnicity, education, marital status, Hispanic origin, nativity, citizenship, veteran status), geography (state, region, metropolitan status), household composition (household role, household size, own children), displacement circumstances (reason and timing of job loss, occupation and industry of the lost job), lost-job history (weekly earnings, tenure), reemployment history (number of jobs held since displacement, weeks without work before the next job), and core macro context (national economic conditions over the most recently completed quarters). The task predicts current earnings conditional on the respondent having lost a job and found another one.\par

\textbf{No-change baseline.} The no-change prediction is the worker's weekly earnings on the job that was lost.\par

\textbf{Sample construction.} Starting from the Displaced Worker Supplement, we keep respondents who report a qualifying job loss and are employed at the time of the supplement. We convert current and lost-job pay to weekly earnings. Next, we drop rows that are missing the current earnings target; any required current demographic, household, geographic, or macroeconomic predictor; lost-job weekly earnings or tenure; or either reemployment-history measure. The reported reason and timing of displacement, occupation and industry of the lost job, advance notice, full-time status, union status, employer type, and health-insurance provision are allowed to remain missing. Finally, within each survey wave, we trim the top and bottom percentiles of current and lost-job weekly earnings.\par

\begin{tcolorbox}[
  breakable,
  lines before break=8,
  colback=white,
  colframe=black!55,
  boxrule=0.45pt,
  arc=0pt,
  left=7pt,
  right=7pt,
  top=7pt,
  bottom=7pt
]
\begingroup
\setlength{\parindent}{0pt}
\setlength{\parskip}{5pt}
\hyphenpenalty=1800
\exhyphenpenalty=1800
\emergencystretch=0.8em
{\centering\fontsize{7.7}{9.0}\selectfont\bfseries Prompt example\par
\texttt{income\_cps\_displace}\par}
\medskip
{\color{black!25}\hrule height 0.35pt}
\vspace{5pt}
\fontsize{7.2}{8.55}\selectfont\rmfamily
\paragraph{\fontsize{7.2}{8.55}\selectfont System.} You are an American person answering questions about your income, work, and labor-force situation.\par
\paragraph{\fontsize{7.2}{8.55}\selectfont User.}
Here is some background information about yourself and your household. You are a 40-year-old White woman with a bachelor's degree. You are divorced and live in Alabama, in a metropolitan area in the South. You live alone and have no own children in your household. You are not Hispanic, were born in the United States, are a U.S. citizen, and are not a veteran.\par

You are also aware of recent national economic conditions. Over the four most recently completed calendar quarters, listed from oldest to most recent, U.S. real GDP changed by 1.40\%, 1.70\%, 0.83\%, and 1.72\% quarter over quarter; the unemployment rate was 6.23\%, 5.93\%, 5.07\%, and 4.17\%; overall consumer prices changed by 1.02\%, 1.82\%, 1.59\%, and 2.18\% quarter over quarter; and the effective federal funds rate was 0.08\%, 0.07\%, 0.09\%, and 0.08\%. Over the five years ending in the most recently completed quarter, compound average quarterly real GDP growth was 0.65\%, the average unemployment rate was 5.08\%, compound average quarterly CPI inflation was 0.71\%, and the average effective federal funds rate was 1.09\%.\par

Note that the current month is January, and that \$ denotes amounts in U.S. dollars. During the past three years, you lost or left a full-time job at a private for-profit employer because there was insufficient work. You last worked at that job two years ago and received less than 1 month of notice before the job loss. You were not a labor-union member at that job, and the job provided health insurance. You worked at the lost job for about 1.5 years and earned approximately \$1,096.16 per week.\par

Since losing or leaving that job, you have held 1 job. You went 48 weeks without work between the end of the lost job and the start of your next job. You are currently employed.\par

Approximately how much do you earn per week at your current job?\par

Return exactly one valid JSON array containing one number in current U.S. dollars per week. Your output must follow this exact array structure: [v\_\allowbreak{}1]. Replace every v placeholder with a single number. Do not include a dollar sign, commas, keys, or explanatory text.\par
\paragraph{\fontsize{7.2}{8.55}\selectfont Assistant.} [1307.7]\par
\endgroup
\end{tcolorbox}

\hypertarget{task:income_sce_policy}{}
\subsection{income\_\allowbreak{}sce\_\allowbreak{}policy}

\textbf{Target.} The respondent's stated effect on their own household under four hypothetical policy changes over the next 12 months: changes in federal welfare benefits, unemployment benefits, the payroll tax rate, and the average income tax rate. For each policy, the survey asks whether the effect would be very negative, somewhat negative, neutral, somewhat positive, or very positive. The task predicts one of these five answers for each policy.\par

\textbf{Predictors.} Household finances and expectations (current financial situation, expected financial situation one year ahead, one-year and longer-run inflation expectations), work situation (current work-status indicators), respondent demographics, resources, and geography (age, education, household before-tax income bracket, region, numeracy), and core macro context (real GDP growth, unemployment rate, headline CPI inflation, and the effective federal funds rate over the four most recently completed quarters, plus five-year quarterly references). The prompt states whether federal welfare benefits, unemployment benefits, the payroll tax rate, and the average income tax rate increase or decrease.\par

\textbf{No-change baseline.} The no-change prediction is that each of the four policy changes has no effect on the household.\par

\textbf{Sample construction.} We merge the SCE Public Policy module with respondent information from the SCE Core survey and national economic variables from FRED-QD. Next, we drop rows that are missing one or more of the four household-impact targets, any of the four corresponding policy directions, or any required demographic, income, or macroeconomic predictor. Current household-finance, inflation-expectation, and work-status predictors are allowed to remain missing. We keep answers that map to the five target categories and require respondents to be between 18 and 100 years old.\par

\begin{tcolorbox}[
  breakable,
  lines before break=8,
  colback=white,
  colframe=black!55,
  boxrule=0.45pt,
  arc=0pt,
  left=7pt,
  right=7pt,
  top=7pt,
  bottom=7pt
]
\begingroup
\setlength{\parindent}{0pt}
\setlength{\parskip}{5pt}
\hyphenpenalty=1800
\exhyphenpenalty=1800
\emergencystretch=0.8em
{\centering\fontsize{7.7}{9.0}\selectfont\bfseries Prompt example\par
\texttt{income\_sce\_policy}\par}
\medskip
{\color{black!25}\hrule height 0.35pt}
\vspace{5pt}
\fontsize{7.2}{8.55}\selectfont\rmfamily
\paragraph{\fontsize{7.2}{8.55}\selectfont System.} You are an American adult making household decisions and forming expectations about spending, work, policy, and the economy.\par
\paragraph{\fontsize{7.2}{8.55}\selectfont User.}
Here is some background information about yourself and your household. You are a 54-year-old, have some college education, live in the South, and have a low numeracy score.\par

Here are your household finances, work situation, and expectations. Note that the current month is April, and that \$ denotes amounts in U.S. dollars. Your household's total pre-tax income over the past 12 months was between \$75,000 and \$99,999. Your household is somewhat better off than it was 12 months ago, and you expect it to be somewhat better off 12 months from now. You expect inflation to be -1.0\% over the next 12 months and 6.0\% over the next three years. You work full-time.\par

You are also aware of recent national economic conditions. Over the four most recently completed calendar quarters, listed from oldest to most recent, U.S. real GDP changed by 0.62\%, 0.40\%, 0.18\%, and 0.58\% quarter over quarter; the unemployment rate was 5.43\%, 5.10\%, 5.03\%, and 4.90\%; overall consumer prices changed by 0.68\%, 0.38\%, -0.01\%, and -0.06\% quarter over quarter; and the effective federal funds rate was 0.12\%, 0.14\%, 0.16\%, and 0.36\%. Over the five years ending in the most recently completed quarter, compound average quarterly real GDP growth was 0.58\%, the average unemployment rate was 6.95\%, compound average quarterly CPI inflation was 0.34\%, and the average effective federal funds rate was 0.12\%.\par

For this question, suppose the following policy changes happened. Federal welfare benefits decrease, unemployment benefits increase, the payroll tax rate increases, and the average income tax rate increases.\par

For each policy change, what impact would it have on your own household? Use one of "very\_\allowbreak{}negative", "somewhat\_\allowbreak{}negative", "no\_\allowbreak{}impact", "somewhat\_\allowbreak{}positive", or "very\_\allowbreak{}positive" for each item.\par

Return only a valid JSON array with exactly four string elements, in this order: welfare benefits; unemployment benefits; payroll tax rate; average income tax rate. Each element must be one of "very\_\allowbreak{}negative", "somewhat\_\allowbreak{}negative", "no\_\allowbreak{}impact", "somewhat\_\allowbreak{}positive", or "very\_\allowbreak{}positive". Your output must follow this exact array structure: ["s\_\allowbreak{}1", "s\_\allowbreak{}2", "s\_\allowbreak{}3", "s\_\allowbreak{}4"]. Replace every s placeholder with one of the permitted strings. Do not include keys or explanatory text.\par
\paragraph{\fontsize{7.2}{8.55}\selectfont Assistant.} ["somewhat\_\allowbreak{}negative", "somewhat\_\allowbreak{}negative", "somewhat\_\allowbreak{}negative", "somewhat\_\allowbreak{}negative"]\par
\endgroup
\end{tcolorbox}

\hypertarget{task:labor_cps_jobfind}{}
\subsection{labor\_\allowbreak{}cps\_\allowbreak{}jobfind}

\textbf{Target.} Whether a person who is unemployed in the current CPS interview is employed, unemployed, or out of the labor force one month later. We obtain the outcome by matching the person to the next monthly CPS interview and use the survey's standard classification based on reported work, job search, availability, and reasons for not working.\par

\textbf{Predictors.} Current unemployment (duration, search or temporary-layoff status, timing of last full-time work, and worker class on the most recent job when available), respondent demographics (age, sex, race, education, marital status, Hispanic origin, nativity, citizenship, veteran status), geography (state, region, metropolitan status), household composition (household role, household size, own children), and core macro context (national economic conditions over the most recently completed quarters).\par

\textbf{No-change baseline.} The no-change prediction is that the person remains unemployed one month later.\par

\textbf{Sample construction.} Starting from the monthly CPS, we keep people aged 16 or older who are unemployed and scheduled to be interviewed again in the following month. We match each person uniquely to the expected next interview. Next, we drop rows that are missing the next-month labor-force-status target; any current demographic, household, geographic, or macroeconomic predictor; unemployment duration; search or layoff status; or time since last full-time work. Worker class on the most recent job is allowed to remain missing.\par

\begin{tcolorbox}[
  breakable,
  lines before break=8,
  colback=white,
  colframe=black!55,
  boxrule=0.45pt,
  arc=0pt,
  left=7pt,
  right=7pt,
  top=7pt,
  bottom=7pt
]
\begingroup
\setlength{\parindent}{0pt}
\setlength{\parskip}{5pt}
\hyphenpenalty=1800
\exhyphenpenalty=1800
\emergencystretch=0.8em
{\centering\fontsize{7.7}{9.0}\selectfont\bfseries Prompt example\par
\texttt{labor\_cps\_jobfind}\par}
\medskip
{\color{black!25}\hrule height 0.35pt}
\vspace{5pt}
\fontsize{7.2}{8.55}\selectfont\rmfamily
\paragraph{\fontsize{7.2}{8.55}\selectfont System.} You are an American person answering questions about your work, job search, and labor-force situation.\par
\paragraph{\fontsize{7.2}{8.55}\selectfont User.}
Here is some background information about yourself and your household. You are a 31-year-old multiracial woman with some college or an associate degree. You have never been married and live in New Jersey, in a metropolitan area in the Northeast. There are 5 people in your household, and you do not live with any of your own children. You are not Hispanic. You were born outside the United States and are a naturalized U.S. citizen. You are not a veteran.\par

Here is your current and recent labor-market situation. Note that the current month is December. You are currently unemployed and are seeking full-time work. You have been unemployed for 29 weeks. The last time you worked full-time for at least 2 consecutive weeks was more than 12 months ago. In your most recent job, you worked in the private sector.\par

You are also aware of recent national economic conditions. Over the four most recently completed calendar quarters, listed from oldest to most recent, U.S. real GDP changed by 0.46\%, -0.16\%, 0.95\%, and 1.08\% quarter over quarter; the unemployment rate was 4.13\%, 4.13\%, 4.20\%, and 4.33\%; overall consumer prices changed by 0.79\%, 0.91\%, 0.41\%, and 0.76\% quarter over quarter; and the effective federal funds rate was 4.65\%, 4.33\%, 4.33\%, and 4.29\%. Over the five years ending in the most recently completed quarter, compound average quarterly real GDP growth was 0.78\%, the average unemployment rate was 4.30\%, compound average quarterly CPI inflation was 1.11\%, and the average effective federal funds rate was 3.04\%.\par

Next month, what will be your work status? For this question, "employed" means working for pay or profit or being temporarily absent from a job, "unemployed" means not employed but available for work and actively searching or on temporary layoff, and "not\_\allowbreak{}in\_\allowbreak{}labor\_\allowbreak{}force" means neither employed nor unemployed.\par

Return exactly one valid JSON array containing one string, using one of: "employed", "unemployed", "not\_\allowbreak{}in\_\allowbreak{}labor\_\allowbreak{}force". Your output must follow this exact array structure: ["s\_\allowbreak{}1"]. Replace every s placeholder with one of the permitted strings. Do not include explanatory text.\par
\paragraph{\fontsize{7.2}{8.55}\selectfont Assistant.} ["not\_\allowbreak{}in\_\allowbreak{}labor\_\allowbreak{}force"]\par
\endgroup
\end{tcolorbox}

\hypertarget{task:labor_cps_retire}{}
\subsection{labor\_\allowbreak{}cps\_\allowbreak{}retire}

\textbf{Target.} Whether a person who is employed in the current CPS interview is retired 12 months later. We match the person to the CPS interview one year later and classify the person as retired when they are out of the labor force and report retirement as the reason they are not working.\par

\textbf{Predictors.} Current work situation (actual and usual hours, worker class, multiple-job status, union coverage, weekly earnings, occupation, industry), household circumstances (family-income range, spouse or partner status and, when present, that person's age and labor-force status), health (functional difficulty), respondent demographics and geography (age, sex, race and ethnicity, education, marital status, state, region, metropolitan status, nativity), and core macro context (national economic conditions over the most recently completed quarters).\par

\textbf{No-change baseline.} The no-change prediction is that the person is not retired 12 months later.\par

\textbf{Sample construction.} Starting from CPS interviews from 1995 onward, we keep employed people between 50 and 75 years old who are scheduled to be interviewed again 12 months later. We match each person uniquely to the expected later interview. Next, we drop rows that are missing the later labor-force-status target, any required current demographic, household, geographic, or macroeconomic predictor, usual weekly hours, or hours worked in the previous week. Spouse or partner information, worker class, multiple-job and union status, weekly earnings, work-limiting difficulty, family income, occupation, and industry are allowed to remain missing. We also require sex and race to agree across the two interviews, age to advance consistently, and the person identifier to agree when it is available in both records. Finally, within each survey month, we trim the top and bottom percentiles of current weekly earnings when earnings are observed.\par

\begin{tcolorbox}[
  breakable,
  lines before break=8,
  colback=white,
  colframe=black!55,
  boxrule=0.45pt,
  arc=0pt,
  left=7pt,
  right=7pt,
  top=7pt,
  bottom=7pt
]
\begingroup
\setlength{\parindent}{0pt}
\setlength{\parskip}{5pt}
\hyphenpenalty=1800
\exhyphenpenalty=1800
\emergencystretch=0.8em
{\centering\fontsize{7.7}{9.0}\selectfont\bfseries Prompt example\par
\texttt{labor\_cps\_retire}\par}
\medskip
{\color{black!25}\hrule height 0.35pt}
\vspace{5pt}
\fontsize{7.2}{8.55}\selectfont\rmfamily
\paragraph{\fontsize{7.2}{8.55}\selectfont System.} You are an American person answering questions about your work, job search, and labor-force situation.\par
\paragraph{\fontsize{7.2}{8.55}\selectfont User.}
Here is some background information about yourself and your household. You are a 62-year-old Black woman with a high school diploma or GED. You are widowed and live in Alabama, in a metropolitan area in the South. There are 2 people in your household, including 1 of your own children. You are not Hispanic, were born in the United States, are a U.S. citizen, and are not a veteran. You do not live with a spouse or partner.\par

Here are your current work, resources, and health. Note that the current month is September, and that \$ denotes amounts in U.S. dollars. You are currently employed by a private for-profit employer. You worked 40 hours last week, usually work 40 hours per week, and do not hold multiple jobs. Your union coverage and weekly earnings are not observed. Whether you have a functional difficulty is not observed. Your family-income category is not observed. Your occupation is Building and Grounds Cleaning and Maintenance, and your industry is Private households.\par

You are also aware of recent national economic conditions. Over the four most recently completed calendar quarters, listed from oldest to most recent, U.S. real GDP changed by 0.58\%, 1.15\%, 0.35\%, and 0.30\% quarter over quarter; the unemployment rate was 6.00\%, 5.63\%, 5.47\%, and 5.67\%; overall consumer prices changed by 0.93\%, 0.58\%, 0.73\%, and 0.82\% quarter over quarter; and the effective federal funds rate was 4.49\%, 5.17\%, 5.81\%, and 6.02\%. Over the five years ending in the most recently completed quarter, compound average quarterly real GDP growth was 0.59\%, the average unemployment rate was 6.62\%, compound average quarterly CPI inflation was 0.82\%, and the average effective federal funds rate was 4.67\%.\par

Will you be retired one year from now? Use "retired" if you will be retired and "not\_\allowbreak{}retired" if you will not be retired.\par

Return exactly one valid JSON array containing one string: "retired" or "not\_\allowbreak{}retired". Your output must follow this exact array structure: ["s\_\allowbreak{}1"]. Replace every s placeholder with one of the permitted strings. Do not include explanatory text.\par
\paragraph{\fontsize{7.2}{8.55}\selectfont Assistant.} ["not\_\allowbreak{}retired"]\par
\endgroup
\end{tcolorbox}

\hypertarget{task:labor_cps_separation}{}
\subsection{labor\_\allowbreak{}cps\_\allowbreak{}separation}

\textbf{Target.} Whether a person who is employed in the current CPS interview is employed, unemployed, or out of the labor force one month later. We obtain the outcome by matching the person to the next monthly CPS interview and use the survey's standard classification based on reported work, job search, availability, and reasons for not working.\par

\textbf{Predictors.} Current work situation (employment status, actual and usual hours, worker class, multiple-job status), respondent demographics (age, sex, race, education, marital status, Hispanic origin, nativity, citizenship, veteran status), geography (state, region, metropolitan status), household composition (household role, household size, own children), and core macro context (national economic conditions over the most recently completed quarters).\par

\textbf{No-change baseline.} The no-change prediction is that the person remains employed one month later.\par

\textbf{Sample construction.} Starting from the monthly CPS, we keep people aged 16 or older who are employed and scheduled to be interviewed again in the following month. We match each person uniquely to the expected next interview and require a valid identifier that appears only once in the baseline month. Next, we drop rows that are missing the next-month labor-force-status target; any current demographic, household, geographic, or macroeconomic predictor; actual weekly hours; or usual weekly hours. Worker class and multiple-job status are allowed to remain missing.\par

\begin{tcolorbox}[
  breakable,
  lines before break=8,
  colback=white,
  colframe=black!55,
  boxrule=0.45pt,
  arc=0pt,
  left=7pt,
  right=7pt,
  top=7pt,
  bottom=7pt
]
\begingroup
\setlength{\parindent}{0pt}
\setlength{\parskip}{5pt}
\hyphenpenalty=1800
\exhyphenpenalty=1800
\emergencystretch=0.8em
{\centering\fontsize{7.7}{9.0}\selectfont\bfseries Prompt example\par
\texttt{labor\_cps\_separation}\par}
\medskip
{\color{black!25}\hrule height 0.35pt}
\vspace{5pt}
\fontsize{7.2}{8.55}\selectfont\rmfamily
\paragraph{\fontsize{7.2}{8.55}\selectfont System.} You are an American person answering questions about your work, job search, and labor-force situation.\par
\paragraph{\fontsize{7.2}{8.55}\selectfont User.}
Here is some background information about yourself and your household. You are a 61-year-old White man with some college or an associate degree. You are divorced and live in Texas, in a metropolitan area in the South. There is 1 person in your household, and you do not live with any of your own children. You are not Hispanic, were born in the United States, are a U.S. citizen, and are not a veteran.\par

Here is your current job. Note that the current month is January. You are currently employed as self-employed in an incorporated business. You worked 30 hours last week, usually work 30 hours per week, and do not hold multiple jobs.\par

You are also aware of recent national economic conditions. Over the four most recently completed calendar quarters, listed from oldest to most recent, U.S. real GDP changed by 0.17\%, 0.58\%, 0.48\%, and 1.36\% quarter over quarter; the unemployment rate was 7.13\%, 7.07\%, 6.80\%, and 6.63\%; overall consumer prices changed by 0.73\%, 0.72\%, 0.46\%, and 0.83\% quarter over quarter; and the effective federal funds rate was 3.04\%, 3.00\%, 3.06\%, and 2.99\%. Over the five years ending in the most recently completed quarter, compound average quarterly real GDP growth was 0.57\%, the average unemployment rate was 6.42\%, compound average quarterly CPI inflation was 0.97\%, and the average effective federal funds rate was 5.91\%.\par

One month from now, what will be your work status? For this question, "employed" means working for pay or profit or being temporarily absent from a job, "unemployed" means not employed but available for work and actively searching or on temporary layoff, and "not\_\allowbreak{}in\_\allowbreak{}labor\_\allowbreak{}force" means neither employed nor unemployed.\par

Return exactly one valid JSON array containing one string, using one of: "employed", "unemployed", "not\_\allowbreak{}in\_\allowbreak{}labor\_\allowbreak{}force". Your output must follow this exact array structure: ["s\_\allowbreak{}1"]. Replace every s placeholder with one of the permitted strings. Do not include explanatory text.\par
\paragraph{\fontsize{7.2}{8.55}\selectfont Assistant.} ["employed"]\par
\endgroup
\end{tcolorbox}

\hypertarget{task:labor_sce_offer}{}
\subsection{labor\_\allowbreak{}sce\_\allowbreak{}offer}

\textbf{Target.} Whether the respondent accepted or rejected a job offer. Respondents list as many as three of their best offers from the previous four months; we take the first listed, which is not necessarily the first one received. We treat the offer as accepted whether or not the respondent still works in that job, and omit respondents who were still deciding.\par

\textbf{Predictors.} Job-offer context (number of recent offers, annual salary of the first listed offer, whether that offer is full time or part time), reservation wage (the annual full-time wage the respondent would require), household finances and expectations (current and expected finances, one-year and longer-run inflation expectations), work situation (current work-status indicators), respondent demographics, resources, and geography (age, education, household income, region, numeracy), and core macro context (national economic conditions over the most recently completed quarters).\par

\textbf{No-change baseline.} The no-change prediction is that the respondent rejected the first listed job offer.\par

\textbf{Sample construction.} We merge the SCE Labor Market supplement with respondent information from the SCE Core survey and national economic variables from FRED-QD. We keep respondents with at least one job offer. Next, we drop rows that are missing the first-offer accept-or-reject target or any required demographic, income, or macroeconomic predictor. The offered salary, full-time or part-time status, reservation wage, household-finance and inflation-expectation measures, and work-status indicators are allowed to remain missing. We then require respondents to be between 18 and 100 years old and require offered annual pay and reservation wage to be at least \$500 when those amounts are reported. Finally, within each survey month, we trim the top and bottom percentiles of offered pay and reservation wage when observed.\par

\begin{tcolorbox}[
  breakable,
  lines before break=8,
  colback=white,
  colframe=black!55,
  boxrule=0.45pt,
  arc=0pt,
  left=7pt,
  right=7pt,
  top=7pt,
  bottom=7pt
]
\begingroup
\setlength{\parindent}{0pt}
\setlength{\parskip}{5pt}
\hyphenpenalty=1800
\exhyphenpenalty=1800
\emergencystretch=0.8em
{\centering\fontsize{7.7}{9.0}\selectfont\bfseries Prompt example\par
\texttt{labor\_sce\_offer}\par}
\medskip
{\color{black!25}\hrule height 0.35pt}
\vspace{5pt}
\fontsize{7.2}{8.55}\selectfont\rmfamily
\paragraph{\fontsize{7.2}{8.55}\selectfont System.} You are an American person answering questions about your work, job search, and labor-force situation.\par
\paragraph{\fontsize{7.2}{8.55}\selectfont User.}
Here is some background information about yourself and your household. You are a 66-year-old, have a college education or more, live in the Midwest, and have a high numeracy score.\par

Here are your household finances, work situation, and expectations. Note that the current month is November, and that \$ denotes amounts in U.S. dollars. Your household's total pre-tax income over the past 12 months was between \$75,000 and \$99,999. Your household is somewhat better off than it was 12 months ago, and you expect it to be somewhat better off 12 months from now. You expect inflation to be 2.1\% over the next 12 months and 2.2\% over the next three years. You work part-time.\par

You are also aware of recent national economic conditions. Over the four most recently completed calendar quarters, listed from oldest to most recent, U.S. real GDP changed by 0.87\%, -0.35\%, 1.29\%, and 1.22\% quarter over quarter; the unemployment rate was 6.93\%, 6.67\%, 6.20\%, and 6.07\%; overall consumer prices changed by 0.37\%, 0.62\%, 0.53\%, and 0.26\% quarter over quarter; and the effective federal funds rate was 0.09\%, 0.07\%, 0.09\%, and 0.09\%. Over the five years ending in the most recently completed quarter, compound average quarterly real GDP growth was 0.60\%, the average unemployment rate was 8.24\%, compound average quarterly CPI inflation was 0.49\%, and the average effective federal funds rate was 0.12\%.\par

For this question, consider the following job offer. You received at least one job offer in the last four months. The first offer was part-time and had an annual salary of \$2,200. Your reservation wage for a full-time job was approximately \$45,760 per year.\par

Thinking about the first job offer you received in the last four months, did you accept or reject that offer? Use "accepted" if you accepted the offer, whether or not you still work at that job. Use "rejected" if you did not accept it.\par

Return exactly one valid JSON array containing one string, using one of: "accepted", "rejected". Your output must follow this exact array structure: ["s\_\allowbreak{}1"]. Replace every s placeholder with one of the permitted strings. Do not include explanatory text.\par
\paragraph{\fontsize{7.2}{8.55}\selectfont Assistant.} ["rejected"]\par
\endgroup
\end{tcolorbox}

\hypertarget{task:labor_sce_risk}{}
\subsection{labor\_\allowbreak{}sce\_\allowbreak{}risk}

\textbf{Target.} The respondent's stated chances of three labor-market events: losing their current job over the next 12 months, leaving it voluntarily over the next 12 months, and finding a new job within three months if they lost their current job. Each answer comes directly from a survey question asking for a probability from 0 to 100 percent. For evaluation, we express each answer as the predicted probabilities that the event does and does not occur.\par

\textbf{Predictors.} Previous expectations (the same three stated chances one month earlier), current work situation (work-status indicators, number of paid jobs, and type of employment arrangement when available), household finances and health (current finances, expected finances, health when available), respondent demographics, resources, and geography (age, education, household income, region, numeracy), and core macro context (national economic conditions over the most recently completed quarters).\par

\textbf{No-change baseline.} For each event, the no-change prediction is the respondent's stated probability one month earlier.\par

\textbf{Sample construction.} We link each respondent's SCE Core survey record to their interview in the immediately preceding month and merge national economic variables from FRED-QD. Next, we drop rows that are missing one or more of the three probability targets, any of the corresponding first-history probabilities, or any required demographic, income, or macroeconomic predictor. Current work, household-finance, and health predictors are allowed to remain missing. We then require every probability to lie between 0 and 100 percent and respondents to be between 18 and 100 years old.\par

\begin{tcolorbox}[
  breakable,
  lines before break=8,
  colback=white,
  colframe=black!55,
  boxrule=0.45pt,
  arc=0pt,
  left=7pt,
  right=7pt,
  top=7pt,
  bottom=7pt
]
\begingroup
\setlength{\parindent}{0pt}
\setlength{\parskip}{5pt}
\hyphenpenalty=1800
\exhyphenpenalty=1800
\emergencystretch=0.8em
{\centering\fontsize{7.7}{9.0}\selectfont\bfseries Prompt example\par
\texttt{labor\_sce\_risk}\par}
\medskip
{\color{black!25}\hrule height 0.35pt}
\vspace{5pt}
\fontsize{7.2}{8.55}\selectfont\rmfamily
\paragraph{\fontsize{7.2}{8.55}\selectfont System.} You are an American person answering questions about your work, job search, and labor-force situation.\par
\paragraph{\fontsize{7.2}{8.55}\selectfont User.}
Here is some background information about yourself and your household. You are a 40-year-old, have a high-school education, live in the West, and have a low numeracy score.\par

Here are your household finances, work situation, and expectations. Note that the current month is August, and that \$ denotes amounts in U.S. dollars. Your household's total pre-tax income over the past 12 months was between \$10,000 and \$19,999. Your household is somewhat worse off than it was 12 months ago, but you expect your household finances to be about the same 12 months from now. You work full-time and currently have 1 paid job, and you work for someone else. One month ago, you assigned probability 0.500 to losing your current or main job in the following 12 months, probability 0.490 to voluntarily leaving that job in the following 12 months, and probability 0.490 to finding an acceptable job within 3 months if you lost the job that month.\par

You are also aware of recent national economic conditions. Over the four most recently completed calendar quarters, listed from oldest to most recent, U.S. real GDP changed by 0.14\%, 0.12\%, 0.99\%, and 0.27\% quarter over quarter; the unemployment rate was 8.03\%, 7.80\%, 7.73\%, and 7.53\%; overall consumer prices changed by 0.45\%, 0.66\%, 0.40\%, and -0.11\% quarter over quarter; and the effective federal funds rate was 0.14\%, 0.16\%, 0.14\%, and 0.12\%. Over the five years ending in the most recently completed quarter, compound average quarterly real GDP growth was 0.22\%, the average unemployment rate was 8.59\%, compound average quarterly CPI inflation was 0.37\%, and the average effective federal funds rate was 0.25\%.\par

For your current or main job, what probability do you think there is that each of the following will happen: you lose the job in the next 12 months; you voluntarily leave the job in the next 12 months; and you find an acceptable job within 3 months if you lose the job this month?\par

Return only a valid JSON array containing exactly three two-element probability distributions, in this order: job loss in the next 12 months; voluntary job leaving in the next 12 months; acceptable job finding within 3 months conditional on job loss. Within each distribution, use this order: event does not occur, event occurs. Express each probability as a number from 0 to 1. Each distribution must sum to 1. Your output must follow this exact array structure: [[p\_\allowbreak{}1\_\allowbreak{}1, p\_\allowbreak{}1\_\allowbreak{}2], [p\_\allowbreak{}2\_\allowbreak{}1, p\_\allowbreak{}2\_\allowbreak{}2], [p\_\allowbreak{}3\_\allowbreak{}1, p\_\allowbreak{}3\_\allowbreak{}2]]. Replace every p placeholder with a number from 0 to 1. Do not include keys or explanatory text.\par
\paragraph{\fontsize{7.2}{8.55}\selectfont Assistant.} [[0.5, 0.5], [0.49, 0.51], [0.5, 0.5]]\par
\endgroup
\end{tcolorbox}

\hypertarget{task:labor_sce_search}{}
\subsection{labor\_\allowbreak{}sce\_\allowbreak{}search}

\textbf{Target.} The respondent's stated chance of finding an acceptable job within three months and within 12 months. Both answers come directly from survey questions asked of people who are looking for work and are reported as probabilities from 0 to 100 percent. For evaluation, we express each answer as the predicted probabilities that the person does and does not find an acceptable job within the stated period.\par

\textbf{Predictors.} Previous expectations (the same two stated chances one month earlier), work and job-search situation (current work-status indicators, active job-search status, unemployment duration), household finances and health (current finances, expected finances, health when available), respondent demographics, resources, and geography (age, education, household income, region, numeracy), and core macro context (national economic conditions over the most recently completed quarters).\par

\textbf{No-change baseline.} For each horizon, the no-change prediction is the respondent's stated probability one month earlier.\par

\textbf{Sample construction.} We link each respondent's SCE Core survey record to their interview in the immediately preceding month and merge national economic variables from FRED-QD. We keep respondents who are currently looking for work. Next, we drop rows that are missing either probability target, either corresponding first-history probability, or any required demographic, income, or macroeconomic predictor. Current work and job-search details, household-finance measures, and health are allowed to remain missing. We then require every probability to lie between 0 and 100 percent and respondents to be between 18 and 100 years old.\par

\begin{tcolorbox}[
  breakable,
  lines before break=8,
  colback=white,
  colframe=black!55,
  boxrule=0.45pt,
  arc=0pt,
  left=7pt,
  right=7pt,
  top=7pt,
  bottom=7pt
]
\begingroup
\setlength{\parindent}{0pt}
\setlength{\parskip}{5pt}
\hyphenpenalty=1800
\exhyphenpenalty=1800
\emergencystretch=0.8em
{\centering\fontsize{7.7}{9.0}\selectfont\bfseries Prompt example\par
\texttt{labor\_sce\_search}\par}
\medskip
{\color{black!25}\hrule height 0.35pt}
\vspace{5pt}
\fontsize{7.2}{8.55}\selectfont\rmfamily
\paragraph{\fontsize{7.2}{8.55}\selectfont System.} You are an American person answering questions about your work, job search, and labor-force situation.\par
\paragraph{\fontsize{7.2}{8.55}\selectfont User.}
Here is some background information about yourself and your household. You are an 83-year-old, have a college education or more, live in the West, and have a low numeracy score.\par

Here are your household finances, work situation, and expectations. Your household's total pre-tax income over the past 12 months was between \$30,000 and \$39,999. Your household is somewhat worse off than it was 12 months ago, and you expect it to remain somewhat worse off 12 months from now.\par

Here is your current work and job-search situation. Note that the current month is July, and that \$ denotes amounts in U.S. dollars. You are not working but would like to work. You are currently looking for a job and have been unemployed for 37 months. One month ago, you assigned probability 0.450 to finding an acceptable job within the following 12 months and probability 0.250 to finding one within the following 3 months.\par

You are also aware of recent national economic conditions. Over the four most recently completed calendar quarters, listed from oldest to most recent, U.S. real GDP changed by 0.14\%, 0.12\%, 0.99\%, and 0.27\% quarter over quarter; the unemployment rate was 8.03\%, 7.80\%, 7.73\%, and 7.53\%; overall consumer prices changed by 0.45\%, 0.66\%, 0.40\%, and -0.11\% quarter over quarter; and the effective federal funds rate was 0.14\%, 0.16\%, 0.14\%, and 0.12\%. Over the five years ending in the most recently completed quarter, compound average quarterly real GDP growth was 0.22\%, the average unemployment rate was 8.59\%, compound average quarterly CPI inflation was 0.37\%, and the average effective federal funds rate was 0.25\%.\par

What probability do you think there is that you will find an acceptable job within the coming 12 months? What probability do you think there is that you will find an acceptable job within the coming 3 months?\par

Return only a valid JSON array containing exactly two two-element probability distributions, in this order: acceptable job finding within the coming 12 months; acceptable job finding within the coming 3 months. Within each distribution, use this order: event does not occur, event occurs. Express each probability as a number from 0 to 1. Each distribution must sum to 1. Your output must follow this exact array structure: [[p\_\allowbreak{}1\_\allowbreak{}1, p\_\allowbreak{}1\_\allowbreak{}2], [p\_\allowbreak{}2\_\allowbreak{}1, p\_\allowbreak{}2\_\allowbreak{}2]]. Replace every p placeholder with a number from 0 to 1. Do not include keys or explanatory text.\par
\paragraph{\fontsize{7.2}{8.55}\selectfont Assistant.} [[0.65, 0.35], [0.8, 0.2]]\par
\endgroup
\end{tcolorbox}

\hypertarget{task:labor_cps_displace}{}
\subsection{labor\_\allowbreak{}cps\_\allowbreak{}displace}

\textbf{Target.} Whether a worker who reports a past job displacement is employed, unemployed, or out of the labor force at the time of the Displaced Worker Supplement. We take this outcome from the CPS's standard labor-force classification based on reported work, job search, availability, and reasons for not working.\par

\textbf{Predictors.} Displacement history (lookback period, reason and timing of job loss, advance notice, full-time status, union status when available, employer type, health-insurance provision, and other characteristics of the lost job), respondent demographics and background (age, sex, race and ethnicity, education, marital status, Hispanic origin, nativity, citizenship, veteran status), geography (state, region, metropolitan status), household composition (household role, household size, own children), and core macro context (national economic conditions over the most recently completed quarters).\par

\textbf{No-change baseline.} The no-change prediction is that the displaced worker is employed at the time of the supplement.\par

\textbf{Sample construction.} Starting from the Displaced Worker Supplement, we keep respondents who completed the displacement questions and report a qualifying job loss. Next, we drop rows that are missing the current labor-force-status target or any required current demographic, household, geographic, or macroeconomic predictor. The detailed reason and timing of displacement, advance notice, full-time status, union status, employer type, health-insurance provision, and other lost-job characteristics are allowed to remain missing.\par

\begin{tcolorbox}[
  breakable,
  lines before break=8,
  colback=white,
  colframe=black!55,
  boxrule=0.45pt,
  arc=0pt,
  left=7pt,
  right=7pt,
  top=7pt,
  bottom=7pt
]
\begingroup
\setlength{\parindent}{0pt}
\setlength{\parskip}{5pt}
\hyphenpenalty=1800
\exhyphenpenalty=1800
\emergencystretch=0.8em
{\centering\fontsize{7.7}{9.0}\selectfont\bfseries Prompt example\par
\texttt{labor\_cps\_displace}\par}
\medskip
{\color{black!25}\hrule height 0.35pt}
\vspace{5pt}
\fontsize{7.2}{8.55}\selectfont\rmfamily
\paragraph{\fontsize{7.2}{8.55}\selectfont System.} You are an American person answering questions about your work, job search, and labor-force situation.\par
\paragraph{\fontsize{7.2}{8.55}\selectfont User.}
Here is some background information about yourself and your household. You are a 40-year-old White woman with a bachelor's degree. You are divorced and live in Alabama, in a metropolitan area in the South. There is 1 person in your household, and you do not live with any of your own children. You are not Hispanic, were born in the United States, are a U.S. citizen, and are not a veteran.\par

You are also aware of recent national economic conditions. Over the four most recently completed calendar quarters, listed from oldest to most recent, U.S. real GDP changed by 1.40\%, 1.70\%, 0.83\%, and 1.72\% quarter over quarter; the unemployment rate was 6.23\%, 5.93\%, 5.07\%, and 4.17\%; overall consumer prices changed by 1.02\%, 1.82\%, 1.59\%, and 2.18\% quarter over quarter; and the effective federal funds rate was 0.08\%, 0.07\%, 0.09\%, and 0.08\%. Over the five years ending in the most recently completed quarter, compound average quarterly real GDP growth was 0.65\%, the average unemployment rate was 5.08\%, compound average quarterly CPI inflation was 0.71\%, and the average effective federal funds rate was 1.09\%.\par

Note that the current month is January. During the past three years, you lost or left a full-time job at a private for-profit employer because there was insufficient work. You last worked at that job two years ago and received less than 1 month of notice before the job loss. You were not a labor-union member at that job, and the job provided health insurance.\par

What is your current employment status?\par

Return exactly one valid JSON array containing one string, using one of: "employed", "unemployed", "not\_\allowbreak{}in\_\allowbreak{}labor\_\allowbreak{}force". Your output must follow this exact array structure: ["s\_\allowbreak{}1"]. Replace every s placeholder with one of the permitted strings. Do not include explanatory text.\par
\paragraph{\fontsize{7.2}{8.55}\selectfont Assistant.} ["employed"]\par
\endgroup
\end{tcolorbox}

\hypertarget{task:labor_cps_ui}{}
\subsection{labor\_\allowbreak{}cps\_\allowbreak{}ui}

This task is inspired by \citet{farber2015extended}.\par

\textbf{Target.} Whether an unemployed person in the current CPS interview is employed, unemployed, or out of the labor force one month later. We obtain the outcome by matching the person to the next monthly CPS interview and use the survey's standard classification based on reported work, job search, availability, and reasons for not working.\par

\textbf{Predictors.} Current unemployment (duration and whether the person is searching or on temporary layoff), recent work (worker class, occupation, and industry on the most recent job when available), respondent demographics and geography (age, sex, race and ethnicity, education, marital status, state, region, nativity, citizenship, veteran status), household composition (household role, household size, own children), calendar month, core macro context (national economic conditions over the most recently completed quarters), and unemployment-insurance policy (the number of weeks of regular state benefits, temporary federal extensions, and total available benefits in the person's state and month). Policy values are measured on the fifth day of each month, matching the timing of the CPS interview week.\par

\textbf{No-change baseline.} The no-change prediction is that the person remains unemployed one month later.\par

\textbf{Sample construction.} Starting from monthly CPS interviews from January 2008 through August 2014, we keep unemployed people aged 18--69 who report an eligible reason for unemployment and are scheduled to be interviewed again in the following month. We merge the unemployment-insurance durations in force in each state and month and national and state economic variables, then match each person uniquely to the next interview. Next, we drop rows that are missing the next-month labor-force-status target; any required current demographic, household, geographic, macroeconomic, or policy predictor; current unemployment duration and search status; or the occupation and industry of the most recent job. Time since last full-time work and worker class are allowed to remain missing. We also require consistent identifiers and demographic characteristics across interviews, a valid survey weight, and internally consistent nonnegative measures of regular, temporary, and total benefit duration.\par

\begin{tcolorbox}[
  breakable,
  lines before break=8,
  colback=white,
  colframe=black!55,
  boxrule=0.45pt,
  arc=0pt,
  left=7pt,
  right=7pt,
  top=7pt,
  bottom=7pt
]
\begingroup
\setlength{\parindent}{0pt}
\setlength{\parskip}{5pt}
\hyphenpenalty=1800
\exhyphenpenalty=1800
\emergencystretch=0.8em
{\centering\fontsize{7.7}{9.0}\selectfont\bfseries Prompt example\par
\texttt{labor\_cps\_ui}\par}
\medskip
{\color{black!25}\hrule height 0.35pt}
\vspace{5pt}
\fontsize{7.2}{8.55}\selectfont\rmfamily
\paragraph{\fontsize{7.2}{8.55}\selectfont System.} You are an American person answering questions about your work, job search, and labor-force situation.\par
\paragraph{\fontsize{7.2}{8.55}\selectfont User.}
Here is some background information about yourself and your household. You are a 41-year-old White man with a bachelor's degree. You have never been married and live in New York, in a metropolitan area in the Northeast. There is 1 person in your household, and you do not live with any of your own children. You are not Hispanic, were born in the United States, are a U.S. citizen, and are not a veteran.\par

Here is your current and recent labor-market situation. Note that the current month is January. You are currently unemployed and are seeking full-time work. You have been unemployed for 2 weeks. The last time you worked full-time for at least 2 consecutive weeks was within the last 12 months. In your most recent job, you were self-employed in an unincorporated business.\par

Here is some additional information about your unemployment spell and most recent job. Your unemployment began because a temporary job ended. You are currently actively looking for work. In your most recent job, your industry was construction, and your occupation was management occupations.\par

You are also aware of recent national economic conditions. Over the four most recently completed calendar quarters, listed from oldest to most recent, U.S. real GDP changed by 0.30\%, 0.61\%, 0.58\%, and 0.63\% quarter over quarter; the unemployment rate was 4.50\%, 4.50\%, 4.67\%, and 4.80\%; overall consumer prices changed by 0.98\%, 1.13\%, 0.63\%, and 1.23\% quarter over quarter; and the effective federal funds rate was 5.26\%, 5.25\%, 5.07\%, and 4.50\%. Over the five years ending in the most recently completed quarter, compound average quarterly real GDP growth was 0.76\%, the average unemployment rate was 5.17\%, compound average quarterly CPI inflation was 0.74\%, and the average effective federal funds rate was 3.13\%.\par

Under the rules in your state in the current month, a job loser who met the program requirements could potentially receive unemployment benefits for up to 26 weeks. This total consists of 26 weeks of regular state benefits and 0 weeks of temporary additional benefits. This maximum does not establish that you personally qualified, received benefits continuously, or had that many weeks remaining.\par

Next month, what will be your work status? For this question, "employed" means working for pay or profit or being temporarily absent from a job, "unemployed" means not employed but available for work and actively searching or on temporary layoff, and "not\_\allowbreak{}in\_\allowbreak{}labor\_\allowbreak{}force" means neither employed nor unemployed.\par

Return exactly one valid JSON array containing one string, using one of: "employed", "unemployed", "not\_\allowbreak{}in\_\allowbreak{}labor\_\allowbreak{}force". Your output must follow this exact array structure: ["s\_\allowbreak{}1"]. Replace every s placeholder with one of the permitted strings. Do not include explanatory text.\par
\paragraph{\fontsize{7.2}{8.55}\selectfont Assistant.} ["employed"]\par
\endgroup
\end{tcolorbox}

\hypertarget{task:labor_psid_addedworker}{}
\subsection{labor\_\allowbreak{}psid\_\allowbreak{}addedworker}

This task is inspired by \citet{stephens2002addedworker}.\par

\textbf{Target.} Spouse or partner's annual work hours and labor earnings in the calendar year of the other adult's job loss and in the following calendar year. We take both measures from the PSID's annual questions about hours worked and labor earnings during the preceding year.\par

\textbf{Predictors.} Work and earnings history for both adults (work status, annual hours, and annual labor earnings in up to three earlier records), household resources (family income in those earlier records), household composition and demographics (each adult's age, sex, education, housing tenure, region, family size, number of children), job-loss circumstances (which adult lost the job, whether the loss followed a layoff, firing, employer closure, or employer move, and earlier losses when observed), and core macro context (national economic conditions before and during the job-loss period). \par

\textbf{No-change baseline.} For both target years, the no-change prediction is the nondisplaced adult's annual work hours and labor earnings before the job loss.\par

\textbf{Sample construction.} Starting from PSID couples observed from 1968--1992, we link the household's interviews before the job loss, during the job-loss period, and in the following period. We keep households in which exactly one adult reports a layoff, firing, plant closure, or employer move after having worked, both adults are observed throughout the relevant interviews, and neither adult reports another job loss in the adjacent periods. Next, we drop rows that are missing one or more of the four labor-supply targets; either adult's current age, sex, or family-composition predictors; either adult's first-history work status, hours, or earnings; first-history family income; or the macroeconomic variables. Current education, housing tenure, and region, prior job-loss details, and the second and third histories are allowed to remain missing. We also require both adults to be between 25 and 65 years old in all three periods, the couple to remain together, and both adults to have answered the job-loss questions. Finally, within each survey year, we trim the top and bottom percentiles of every family-income value shown in the household's history.\par

\begin{tcolorbox}[
  breakable,
  lines before break=8,
  colback=white,
  colframe=black!55,
  boxrule=0.45pt,
  arc=0pt,
  left=7pt,
  right=7pt,
  top=7pt,
  bottom=7pt
]
\begingroup
\setlength{\parindent}{0pt}
\setlength{\parskip}{5pt}
\hyphenpenalty=1800
\exhyphenpenalty=1800
\emergencystretch=0.8em
{\centering\fontsize{7.7}{9.0}\selectfont\bfseries Prompt example\par
\texttt{labor\_psid\_addedworker}\par}
\medskip
{\color{black!25}\hrule height 0.35pt}
\vspace{5pt}
\fontsize{7.2}{8.55}\selectfont\rmfamily
\paragraph{\fontsize{7.2}{8.55}\selectfont System.} You are an American adult making decisions about work, income, and household finances.\par
\paragraph{\fontsize{7.2}{8.55}\selectfont User.}
Here is some background information about yourself and your household. You are a 31-year-old woman. Your spouse or partner is a 30-year-old man. Your household has 5 people, including 3 children. You completed 12 years of schooling. Your spouse or partner completed 14 years of schooling. You currently rent your home. You live in the South.\par

Here is your household's work and income history. Note that \$ denotes amounts in U.S. dollars. Three years ago, you worked 2,080 hours in total and had \$6,000 in annual labor earnings. Your spouse or partner worked 2,000 hours in total and had \$7,800 in annual labor earnings. Your total family income was \$13,800. Two years ago, you worked 2,080 hours in total and had \$6,400 in annual labor earnings. Your spouse or partner worked 2,000 hours in total and had \$7,000 in annual labor earnings. Your total family income was \$13,400. One year ago, you worked 2,000 hours in total and had \$6,300 in annual labor earnings. Your spouse or partner worked 2,600 hours in total and had \$9,000 in annual labor earnings. Your total family income was \$15,300.\par

You are also aware of recent national economic conditions. Over the four most recently completed calendar quarters, listed from oldest to most recent, U.S. real GDP changed by -1.22\%, 0.71\%, 1.71\%, and 1.35\% quarter over quarter; the unemployment rate was 8.27\%, 8.87\%, 8.47\%, and 8.30\%; overall consumer prices changed by 2.14\%, 1.20\%, 2.01\%, and 1.84\% quarter over quarter; and the effective federal funds rate was 6.30\%, 5.42\%, 6.16\%, and 5.41\%. Over the five years ending in the most recently completed quarter, compound average quarterly real GDP growth was 0.78\%, the average unemployment rate was 6.11\%, compound average quarterly CPI inflation was 1.68\%, and the average effective federal funds rate was 6.83\%.\par

During the current calendar year, your spouse or partner lost their job because of a layoff or firing.\par

The first two outcomes below cover the entire current calendar year. They can therefore include work before as well as after your spouse or partner lost their job. The last two outcomes cover one year from now.\par

Given this information, what are your annual work hours and labor earnings in the current calendar year, and what will they be in one year?\par

Return only a valid JSON array with exactly four numbers, in this order: your annual work hours in the current calendar year; your annual labor earnings in the current calendar year; your annual work hours in one year; your annual labor earnings in one year. Earnings must be in current U.S. dollars. Your output must follow this exact array structure: [v\_\allowbreak{}1, v\_\allowbreak{}2, v\_\allowbreak{}3, v\_\allowbreak{}4]. Replace every v placeholder with a single number. Do not include dollar signs, commas, keys, or explanatory text.\par
\paragraph{\fontsize{7.2}{8.55}\selectfont Assistant.} [1760.0,8300.0,1984.0,8200.0]\par
\endgroup
\end{tcolorbox}

\hypertarget{task:labor_sce_reswage}{}
\subsection{labor\_\allowbreak{}sce\_\allowbreak{}reswage}

\textbf{Target.} The lowest hourly wage the respondent would accept for a new job and the number of hours per week they would prefer to work at that wage. Both targets come directly from the SCE Job Search supplement. When the respondent reports an acceptable wage by week or year instead of by hour, we convert it to an hourly amount using the reported preferred hours.\par

\textbf{Predictors.} Work and job-search situation (current work-status indicators, number of current jobs when available, duration of job search, months since last paid work, whether the respondent has ever had paid work), pay benchmark (hourly-equivalent pay and usual weekly hours in the current job or, when unavailable, the most recent job), household finances and expectations (current and expected finances, one-year and longer-run inflation expectations), respondent demographics, resources, and household context (age, education, household income, region, numeracy, household size, children under 18), and core macro context (national economic conditions over the most recently completed quarters).\par

\textbf{No-change baseline.} The no-change prediction uses hourly pay and weekly hours in the respondent's current job, or in the last job if the respondent is not currently working.\par

\textbf{Sample construction.} We merge the SCE Job Search supplement with respondent information from the SCE Core survey and national economic variables from FRED-QD, and convert all reported pay amounts to hourly values. Next, we drop rows that are missing either target, any required demographic, income, or macroeconomic predictor, or a complete pay-and-hours benchmark from either the current job or the most recent job. Household size, number of children, household-finance and inflation-expectation measures, and other work and job-search details are allowed to remain missing. We then require the acceptable wage to be positive, preferred weekly hours to be between 1 and 168, and respondents to be between 18 and 100 years old. Finally, within each survey month, we trim the top and bottom percentiles of the acceptable wage and current-job or last-job hourly pay.\par

\begin{tcolorbox}[
  breakable,
  lines before break=8,
  colback=white,
  colframe=black!55,
  boxrule=0.45pt,
  arc=0pt,
  left=7pt,
  right=7pt,
  top=7pt,
  bottom=7pt
]
\begingroup
\setlength{\parindent}{0pt}
\setlength{\parskip}{5pt}
\hyphenpenalty=1800
\exhyphenpenalty=1800
\emergencystretch=0.8em
{\centering\fontsize{7.7}{9.0}\selectfont\bfseries Prompt example\par
\texttt{labor\_sce\_reswage}\par}
\medskip
{\color{black!25}\hrule height 0.35pt}
\vspace{5pt}
\fontsize{7.2}{8.55}\selectfont\rmfamily
\paragraph{\fontsize{7.2}{8.55}\selectfont System.} You are an American person answering questions about your work, job search, and labor-force situation.\par
\paragraph{\fontsize{7.2}{8.55}\selectfont User.}
Here is some background information about yourself and your household. You are a 40-year-old, have a high-school education, live in the West, and have a low numeracy score. Your household has 4 people, including 2 children under 18.\par

Here are your household finances, work situation, and expectations. Note that the current month is October, and that \$ denotes amounts in U.S. dollars. Your household's total pre-tax income over the past 12 months was between \$10,000 and \$19,999. Your household is somewhat worse off than it was 12 months ago, and you expect it to be about the same 12 months from now. You expect inflation to be 2.2\% over the next 12 months and 6.5\% over the next three years. You work full-time. You recently spent 14 days searching for work. Your current or main job pays approximately \$9.00 per hour, and you usually work 40 hours per week.\par

You are also aware of recent national economic conditions. Over the four most recently completed calendar quarters, listed from oldest to most recent, U.S. real GDP changed by 0.12\%, 0.99\%, 0.27\%, and 0.85\% quarter over quarter; the unemployment rate was 7.80\%, 7.73\%, 7.53\%, and 7.23\%; overall consumer prices changed by 0.66\%, 0.40\%, -0.11\%, and 0.54\% quarter over quarter; and the effective federal funds rate was 0.16\%, 0.14\%, 0.12\%, and 0.08\%. Over the five years ending in the most recently completed quarter, compound average quarterly real GDP growth was 0.29\%, the average unemployment rate was 8.65\%, compound average quarterly CPI inflation was 0.32\%, and the average effective federal funds rate was 0.16\%.\par

For this question, consider the following job preferences.\par

What is the lowest hourly wage, before taxes and other deductions, that you would accept for a job in a line of work you would consider? Assuming you could find suitable work, how many hours per week would you prefer to work on a new job?\par

Return only a valid JSON array with exactly two numeric elements, in this order: hourly reservation wage; preferred weekly work hours. Your output must follow this exact array structure: [v\_\allowbreak{}1, v\_\allowbreak{}2]. Replace every v placeholder with a single number. Do not include keys or explanatory text.\par
\paragraph{\fontsize{7.2}{8.55}\selectfont Assistant.} [9.81, 40.0]\par
\endgroup
\end{tcolorbox}

\hypertarget{task:macro_mich_outlook}{}
\subsection{macro\_\allowbreak{}mich\_\allowbreak{}outlook}

\textbf{Target.} The respondent's expected inflation rate over the next 12 months and their expected average annual inflation rate over the next five to ten years. For each horizon, the Michigan Survey first asks whether prices are expected to rise, fall, or remain unchanged and then asks by what percentage. We combine the direction and magnitude into a signed inflation expectation.\par

\textbf{Predictors.} Previous expectations (the same two inflation expectations from the respondent's earlier interview six to eight months before), household and national assessments (personal finances and national business conditions compared with one year earlier, assessment of government economic policy), buying conditions (major household items, homes, and vehicles when available), household resources (annual income), household composition and respondent demographics (age, sex, education, marital or partner status, number of adults, number of children), geography (region), and core macro context (national economic conditions over the most recently completed quarters).\par

\textbf{No-change baseline.} For each horizon, the no-change prediction is the respondent's inflation expectation from the earlier interview.\par

\textbf{Sample construction.} We link each respondent to a unique earlier Michigan Survey interview conducted six, seven, or eight months before and merge national economic variables from FRED-QD. We keep pairs in the correct chronological order whose earlier interview matches the survey's recorded link and was available by the date of the current response. Next, we drop rows that are missing either current inflation-expectation target, either corresponding first-lag expectation, or any required current predictor. Assessments of whether it is a good time to buy a home, sell a house, or buy a vehicle are allowed to remain missing. We also require the categorical responses shown to the model to have valid meanings. Finally, within each survey month, we trim the top and bottom percentiles of both current inflation expectations and annual household income, and apply the same trimming rule to each earlier expectation within its earlier survey month.\par

\begin{tcolorbox}[
  breakable,
  lines before break=8,
  colback=white,
  colframe=black!55,
  boxrule=0.45pt,
  arc=0pt,
  left=7pt,
  right=7pt,
  top=7pt,
  bottom=7pt
]
\begingroup
\setlength{\parindent}{0pt}
\setlength{\parskip}{5pt}
\hyphenpenalty=1800
\exhyphenpenalty=1800
\emergencystretch=0.8em
{\centering\fontsize{7.7}{9.0}\selectfont\bfseries Prompt example\par
\texttt{macro\_mich\_outlook}\par}
\medskip
{\color{black!25}\hrule height 0.35pt}
\vspace{5pt}
\fontsize{7.2}{8.55}\selectfont\rmfamily
\paragraph{\fontsize{7.2}{8.55}\selectfont System.} You are the head of an American household forming expectations about economic conditions, both for the country in general and for your household in particular.\par
\paragraph{\fontsize{7.2}{8.55}\selectfont User.}
Here is some background information about yourself and your household. You are a 62-year-old man living in the West. You are married or partnered, and your highest completed education is some college without a college degree. Your household has 2 adults and no children.\par

Here are your current views on your finances and the economy. Note that the current month is November, and that \$ denotes amounts in U.S. dollars. Your current household income is \$29,000 per year. You report that your personal finances are better than they were a year ago, while national business conditions are worse than they were a year ago. You rate the government's handling of economic policy as only fair. You think now is a good time to buy major household items, buy a home, sell a house, and buy a vehicle.\par

6 months ago, your expected price change over the next 12 months was 2.0\%. Your expected average annual price change over the next 5 to 10 years was 2.0\%.\par

You are also aware of recent national economic conditions. Over the four most recently completed calendar quarters, listed from oldest to most recent, U.S. real GDP changed by 0.35\%, 1.20\%, 1.08\%, and 0.99\% quarter over quarter; the unemployment rate was 7.10\%, 7.37\%, 7.60\%, and 7.63\%; overall consumer prices changed by 0.83\%, 0.68\%, 0.77\%, and 0.76\% quarter over quarter; and the effective federal funds rate was 4.82\%, 4.02\%, 3.77\%, and 3.26\%. Over the five years ending in the most recently completed quarter, compound average quarterly real GDP growth was 0.66\%, the average unemployment rate was 6.06\%, compound average quarterly CPI inflation was 1.05\%, and the average effective federal funds rate was 7.01\%.\par

Over the next 12 months, by what percent do you expect prices in general to change? Over the next 5 to 10 years, by what percent per year on average do you expect prices in general to change? Use a negative number if you expect prices to fall.\par

Return only a valid JSON array with exactly 2 elements, in this order: expected\_\allowbreak{}inflation\_\allowbreak{}1y, expected\_\allowbreak{}inflation\_\allowbreak{}5\_\allowbreak{}to\_\allowbreak{}10y\_\allowbreak{}annual. Both elements must be numbers in percent. Your output must follow this exact array structure: [v\_\allowbreak{}1, v\_\allowbreak{}2]. Replace every v placeholder with a single number. Do not include keys or explanatory text.\par
\paragraph{\fontsize{7.2}{8.55}\selectfont Assistant.} [3.0, 2.0]\par
\endgroup
\end{tcolorbox}

\hypertarget{task:macro_sce_uncertainty}{}
\subsection{macro\_\allowbreak{}sce\_\allowbreak{}uncertainty}

\textbf{Target.} Five probability vectors reported by the respondent: inflation over the next year, inflation over the 12-month period two to three years ahead, and whether U.S. unemployment, the average savings-account interest rate, and U.S. stock prices will be higher one year ahead. For each inflation horizon, the survey asks the respondent to divide 100 percentage points across ten possible inflation ranges. For each of the other three outcomes, it asks directly for the chance that the outcome will be higher.\par

\textbf{Predictors.} Previous expectations (all five probability vectors from one month earlier), household finances and work situation (current household finances, current work-status indicators), respondent demographics, resources, and geography (age, education, household before-tax income bracket, region, numeracy), and core macro context (real GDP growth, unemployment rate, headline CPI inflation, and the effective federal funds rate over the four most recently completed quarters, plus five-year quarterly references).\par

\textbf{No-change baseline.} The no-change prediction carries forward all five probability vectors from one month earlier.\par

\textbf{Sample construction.} We link each respondent to their SCE interview in the immediately preceding month and merge national economic variables from FRED-QD. Next, we drop rows that are missing one or more target answers, any of the five first-history probability vectors, or any required demographic, income, or macroeconomic predictor. Current household-finance and work-status predictors are allowed to remain missing. We require every probability to lie between 0 and 100 percent, the probabilities over the inflation ranges to sum to 100 percent at each horizon apart from rounding, and respondents to be between 18 and 100 years old. Finally, within each survey month, we trim the top and bottom percentiles of the interquartile range implied by each respondent's inflation probabilities.\par

\begin{tcolorbox}[
  breakable,
  lines before break=8,
  colback=white,
  colframe=black!55,
  boxrule=0.45pt,
  arc=0pt,
  left=7pt,
  right=7pt,
  top=7pt,
  bottom=7pt
]
\begingroup
\setlength{\parindent}{0pt}
\setlength{\parskip}{5pt}
\hyphenpenalty=1800
\exhyphenpenalty=1800
\emergencystretch=0.8em
{\centering\fontsize{7.7}{9.0}\selectfont\bfseries Prompt example\par
\texttt{macro\_sce\_uncertainty}\par}
\medskip
{\color{black!25}\hrule height 0.35pt}
\vspace{5pt}
\fontsize{7.2}{8.55}\selectfont\rmfamily
\paragraph{\fontsize{7.2}{8.55}\selectfont System.} You are an American adult making household decisions and forming expectations about your finances, work, housing, and the economy.\par
\paragraph{\fontsize{7.2}{8.55}\selectfont User.}
Here is some background information about yourself and your household. You are a 66-year-old, have a college education or more, live in the Northeast, and have a high numeracy score.\par

Here are your household finances, work situation, and expectations. Note that the current month is July, and that \$ denotes amounts in U.S. dollars. Your household's total pre-tax income over the past 12 months was between \$100,000 and \$149,999. Your household is somewhat better off than it was 12 months ago, but you expect your household finances to be somewhat better off 12 months from now. One month ago, your probabilities for inflation over the following 12 months, in the same ten-bin order used below, were [0.1, 0.15, 0.25, 0.5, 0.0, 0.0, 0.0, 0.0, 0.0, 0.0]. Your probabilities for the 12-month period two to three years ahead were [0.15, 0.2, 0.5, 0.15, 0.0, 0.0, 0.0, 0.0, 0.0, 0.0]. You assigned probability 0.400 to the U.S. unemployment rate being higher 12 months later, probability 0.700 to the average interest rate on savings accounts being higher, and probability 0.500 to U.S. stock prices being higher.\par

You are also aware of recent national economic conditions. Over the four most recently completed calendar quarters, listed from oldest to most recent, U.S. real GDP changed by 0.14\%, 0.12\%, 0.99\%, and 0.27\% quarter over quarter; the unemployment rate was 8.03\%, 7.80\%, 7.73\%, and 7.53\%; overall consumer prices changed by 0.45\%, 0.66\%, 0.40\%, and -0.11\% quarter over quarter; and the effective federal funds rate was 0.14\%, 0.16\%, 0.14\%, and 0.12\%. Over the five years ending in the most recently completed quarter, compound average quarterly real GDP growth was 0.22\%, the average unemployment rate was 8.59\%, compound average quarterly CPI inflation was 0.37\%, and the average effective federal funds rate was 0.25\%.\par

Now think about the different things that may happen to inflation. For each of the following ten ranges, what probability do you think there is that inflation over the next 12 months will fall in that range? Use this order: inflation at least 12 percent, inflation 8 to 12 percent, inflation 4 to 8 percent, inflation 2 to 4 percent, inflation 0 to 2 percent, deflation 0 to 2 percent, deflation 2 to 4 percent, deflation 4 to 8 percent, deflation 8 to 12 percent, deflation at least 12 percent. Then, for the 12-month period two to three years from now, what probability do you think there is that inflation will fall in each of the same ranges? Finally, what probability do you think there is that 12 months from now the U.S. unemployment rate will be higher than it is now, that the average interest rate on savings accounts will be higher than it is now, and that U.S. stock prices will be higher than they are now?\par

Return only a valid JSON array containing exactly five probability distributions, in this order: the ten-element one-year inflation distribution; the ten-element two-to-three-year inflation distribution; the two-element higher-unemployment distribution; the two-element higher-savings-rate distribution; the two-element higher-stock-price distribution. Preserve the inflation-bin order stated above. Within each two-element distribution, use this order: event does not occur, event occurs. Express each probability as a number from 0 to 1. Each distribution must sum to 1. Your output must follow this exact array structure: [[p\_\allowbreak{}1\_\allowbreak{}1, ..., p\_\allowbreak{}1\_\allowbreak{}10], [p\_\allowbreak{}2\_\allowbreak{}1, ..., p\_\allowbreak{}2\_\allowbreak{}10], [p\_\allowbreak{}3\_\allowbreak{}1, p\_\allowbreak{}3\_\allowbreak{}2], [p\_\allowbreak{}4\_\allowbreak{}1, p\_\allowbreak{}4\_\allowbreak{}2], [p\_\allowbreak{}5\_\allowbreak{}1, p\_\allowbreak{}5\_\allowbreak{}2]]. Replace every p placeholder with a number from 0 to 1. Do not include keys or explanatory text.\par
\paragraph{\fontsize{7.2}{8.55}\selectfont Assistant.} [[0.0, 0.0, 0.75, 0.2, 0.05, 0.0, 0.0, 0.0, 0.0, 0.0], [0.0, 0.0, 0.05, 0.65, 0.3, 0.0, 0.0, 0.0, 0.0, 0.0], [0.84, 0.16], [0.25, 0.75], [0.44, 0.56]]\par
\endgroup
\end{tcolorbox}

\hypertarget{task:macro_sce_revision}{}
\subsection{macro\_\allowbreak{}sce\_\allowbreak{}revision}

This task is inspired by \citet{armantier2022expectations}.\par

\textbf{Target.} The respondent's inflation expectations in their twelfth monthly SCE interview: expected inflation over the next 12 months and expected average annual inflation over the 12-month period two to three years ahead. For both horizons, we use the mean implied by the probabilities the respondent assigns to different possible inflation ranges.\par

\textbf{Predictors.} Earlier expectations (both inflation expectations from the respondent's first monthly interview), respondent demographics and resources at the final interview (age, education, household income, region, numeracy, household finances, work situation), core macro context (national economic conditions over the most recently completed quarters), and inflation since the first interview (the observed change in the seasonally adjusted U.S. consumer price index over the intervening 11 months, the earlier one-year expectation converted to the same 11-month period, and the difference between the two). The task therefore asks how the respondent revises expectations after observing inflation during the panel.\par

\textbf{No-change baseline.} For each horizon, the no-change prediction is the respondent's inflation expectation in their first monthly interview.\par

\textbf{Sample construction.} We link each respondent's first and twelfth SCE interviews, require the two interviews to be exactly 11 months apart, and merge national economic variables from FRED-QD and the seasonally adjusted U.S. consumer price index. We calculate inflation over the 11 completed months between the interviews, convert the earlier 12-month expectation to an 11-month rate using constant monthly growth, and subtract the converted expectation from observed inflation. Next, we drop pairs that are missing either final-expectation target, either first-history expectation, any required final-interview demographic, income, or macroeconomic predictor, either price-index value, or any of the three inflation-comparison measures. Final-interview household-finance and work-status predictors are allowed to remain missing. We also require the earlier one-year expectation to exceed -100 percent and respondents to be between 18 and 100 years old. Finally, within each survey month, we trim the top and bottom percentiles of both final inflation expectations.\par

\begin{tcolorbox}[
  breakable,
  lines before break=8,
  colback=white,
  colframe=black!55,
  boxrule=0.45pt,
  arc=0pt,
  left=7pt,
  right=7pt,
  top=7pt,
  bottom=7pt
]
\begingroup
\setlength{\parindent}{0pt}
\setlength{\parskip}{5pt}
\hyphenpenalty=1800
\exhyphenpenalty=1800
\emergencystretch=0.8em
{\centering\fontsize{7.7}{9.0}\selectfont\bfseries Prompt example\par
\texttt{macro\_sce\_revision}\par}
\medskip
{\color{black!25}\hrule height 0.35pt}
\vspace{5pt}
\fontsize{7.2}{8.55}\selectfont\rmfamily
\paragraph{\fontsize{7.2}{8.55}\selectfont System.} You are an American adult making household decisions and forming expectations about your finances, work, housing, and the economy.\par
\paragraph{\fontsize{7.2}{8.55}\selectfont User.}
Here is some background information about yourself and your household. You are a 28-year-old, have a college education or more, live in the Midwest, and have a low numeracy score.\par

Here are your household finances, work situation, and expectations. Your household's total pre-tax income over the past 12 months was between \$10,000 and \$19,999. Your household is about as well off as it was 12 months ago, and you expect your household finances to remain about the same 12 months from now.\par

You are also aware of recent national economic conditions. Over the four most recently completed calendar quarters, listed from oldest to most recent, U.S. real GDP changed by 0.27\%, 0.85\%, 0.87\%, and -0.35\% quarter over quarter; the unemployment rate was 7.53\%, 7.23\%, 6.93\%, and 6.67\%; overall consumer prices changed by -0.11\%, 0.54\%, 0.37\%, and 0.62\% quarter over quarter; and the effective federal funds rate was 0.12\%, 0.08\%, 0.09\%, and 0.07\%. Over the five years ending in the most recently completed quarter, compound average quarterly real GDP growth was 0.48\%, the average unemployment rate was 8.57\%, compound average quarterly CPI inflation was 0.52\%, and the average effective federal funds rate was 0.13\%.\par

Now consider how actual consumer prices have evolved compared with your expectations over recent months. Note that the current month is May, and that \$ denotes amounts in U.S. dollars. Eleven months ago, you expected consumer prices to increase by 12.4\% over the following 12 months, and to increase by 10.3\% over the 12-month period two to three years ahead. Assuming a constant monthly expected inflation rate over the following 12 months, your first expectation corresponds to an expected increase of 11.3\% over the first 11 months of that period. Actual consumer prices increased by 2.0\% over those 11 months. This was 9.3 percentage points below the implied 11-month expectation.\par

Given this information, by what percentage do you now expect consumer prices to change over the next 12 months? By what percentage do you now expect them to change over the 12-month period starting two years from now?\par

Return only a valid JSON array with exactly two numbers, in this order: expected consumer-price change over the next 12 months; expected consumer-price change over the 12-month period starting two years from now. Use a negative number if you expect consumer prices to fall. Your output must follow this exact array structure: [v\_\allowbreak{}1, v\_\allowbreak{}2]. Replace every v placeholder with a single number. Do not include keys or explanatory text.\par
\paragraph{\fontsize{7.2}{8.55}\selectfont Assistant.} [5.071, 3.569]\par
\endgroup
\end{tcolorbox}

\hypertarget{task:house_census_move}{}
\subsection{house\_\allowbreak{}census\_\allowbreak{}move}

\textbf{Target.} Whether an adult lives in the same home as five years earlier, a different home in the same state, or a different state. We construct the three categories from the Census questions on residence five years ago and current and previous state of residence.\par

\textbf{Predictors.} Respondent characteristics five years earlier (age, sex, race and ethnicity, education), location five years earlier (state of residence), economic conditions in that state (annual unemployment rate, per-capita personal income growth, house-price growth), and core macro context before the five-year period (real GDP growth, unemployment rate, headline CPI inflation, and the effective federal funds rate over the preceding four quarters, plus five-year quarterly references).\par

\textbf{No-change baseline.} The no-change prediction is that the respondent remains in the same home.\par

\textbf{Sample construction.} Starting from the 1990 and 2000 Census samples, we use each person's reported residence five years earlier to identify the origin state, then merge state unemployment, income, and house-price data and national economic variables. We keep people who were at least 30 years old at the start of the five-year period. Next, we drop rows that are missing the migration target or any current predictor. No target or predictor shown to the model is allowed to remain missing.\par

\begin{tcolorbox}[
  breakable,
  lines before break=8,
  colback=white,
  colframe=black!55,
  boxrule=0.45pt,
  arc=0pt,
  left=7pt,
  right=7pt,
  top=7pt,
  bottom=7pt
]
\begingroup
\setlength{\parindent}{0pt}
\setlength{\parskip}{5pt}
\hyphenpenalty=1800
\exhyphenpenalty=1800
\emergencystretch=0.8em
{\centering\fontsize{7.7}{9.0}\selectfont\bfseries Prompt example\par
\texttt{house\_census\_move}\par}
\medskip
{\color{black!25}\hrule height 0.35pt}
\vspace{5pt}
\fontsize{7.2}{8.55}\selectfont\rmfamily
\paragraph{\fontsize{7.2}{8.55}\selectfont System.} You are an American adult making decisions about housing and residential location.\par
\paragraph{\fontsize{7.2}{8.55}\selectfont User.}
Here is some background information about yourself and where you lived five years ago. Note that the current month is April. Five years ago, you were a 64-year-old mexican, Hispanic woman with grade 5, 6, 7, or 8. Your state of residence five years ago was Texas.\par

You are also aware of national economic conditions before that five-year period began. Over the four calendar quarters immediately before that period, listed from oldest to most recent, U.S. real GDP changed by 1.95\%, 1.73\%, 0.96\%, and 0.82\% quarter over quarter; the unemployment rate was 7.87\%, 7.43\%, 7.43\%, and 7.30\%; overall consumer prices changed by 1.42\%, 0.94\%, 0.87\%, and 0.86\% quarter over quarter; and the effective federal funds rate was 9.69\%, 10.56\%, 11.39\%, and 9.27\%. Over the five years ending in the last of those four quarters, compound average quarterly real GDP growth was 0.64\%, the average unemployment rate was 8.32\%, compound average quarterly CPI inflation was 1.64\%, and the average effective federal funds rate was 12.26\%.\par

You are also aware of annual state and national conditions in the completed calendar year immediately before that five-year period began. Unemployment in your state was 6.10\%, compared with 7.51\% nationally; per-capita personal income was growing at 7.60\% in your state and 9.35\% nationally; and house prices were growing at 0.98\% in your state and 4.47\% nationally.\par

Compared with where you lived five years ago, do you now live at the same residence, at a different residence in the same state, or in a different state?\par

Return only a valid JSON array containing exactly one string, using one of: "stay", "move\_\allowbreak{}within\_\allowbreak{}state", "move\_\allowbreak{}to\_\allowbreak{}different\_\allowbreak{}state". Your output must follow this exact array structure: ["s\_\allowbreak{}1"]. Replace every s placeholder with one of the permitted strings. Do not include explanatory text.\par
\paragraph{\fontsize{7.2}{8.55}\selectfont Assistant.} ["stay"]\par
\endgroup
\end{tcolorbox}

\hypertarget{task:house_psid_owner}{}
\subsection{house\_\allowbreak{}psid\_\allowbreak{}owner}

\textbf{Target.} Whether a household that currently rents owns a home in its next PSID interview, usually two years later and sometimes one year later in the early part of the sample. We use the housing-tenure answer in the matched later interview and classify the household as an owner or nonowner.\par

\textbf{Predictors.} Household resources (income and available wealth components), household history (up to three prior samples of income, wealth, and housing tenure), household composition (presence of a spouse or partner, marital status, household size, number of children), respondent demographics and geography (age, sex, race and ethnicity, region, metropolitan status when available), core macro context (real GDP growth, unemployment rate, headline CPI inflation, and the effective federal funds rate over the most recently completed quarters, plus five-year quarterly references), and financial and housing-market context (house-price growth, S\&P 500 growth, and the 30-year mortgage rate over the same periods).\par

\textbf{No-change baseline.} The no-change prediction is that the household remains a renter in the next PSID interview.\par

\textbf{Sample construction.} Starting from our clean PSID household panel, we keep current renters between 25 and 75 years old and link each household to its next consecutive interview. We merge national economic, financial-market, and housing-market variables. Next, we drop rows that are missing the next-interview homeownership target or any required current predictor: age, sex, race or ethnicity, region, housing tenure, spouse or partner status, marital status, family size, number of children, family income, or the core macroeconomic variables. Metropolitan status, current wealth fields, changes since the previous interview, all prior household histories, and the financial- and housing-market variables are allowed to remain missing. Finally, within each survey year, we trim the top and bottom percentiles of current and available lagged family income and of current net worth when observed.\par

\begin{tcolorbox}[
  breakable,
  lines before break=8,
  colback=white,
  colframe=black!55,
  boxrule=0.45pt,
  arc=0pt,
  left=7pt,
  right=7pt,
  top=7pt,
  bottom=7pt
]
\begingroup
\setlength{\parindent}{0pt}
\setlength{\parskip}{5pt}
\hyphenpenalty=1800
\exhyphenpenalty=1800
\emergencystretch=0.8em
{\centering\fontsize{7.7}{9.0}\selectfont\bfseries Prompt example\par
\texttt{house\_psid\_owner}\par}
\medskip
{\color{black!25}\hrule height 0.35pt}
\vspace{5pt}
\fontsize{7.2}{8.55}\selectfont\rmfamily
\paragraph{\fontsize{7.2}{8.55}\selectfont System.} You are an American adult making decisions about housing and residential location.\par
\paragraph{\fontsize{7.2}{8.55}\selectfont User.}
Here is some background information about yourself and your household. You are a 57-year-old White woman living in the South, in a metropolitan area. You do not have a spouse or partner in your household and are not legally married. Your household has 1 person and no children. Compared with two years ago, household size has fallen by 1, the number of children is unchanged, partner status is unchanged, and marital status is unchanged.\par

Here are your household's housing, finances, and past household records. Note that \$ denotes amounts in U.S. dollars. You currently rent your home. Your current household income is \$77,187, and your total net worth is \$1,269. You have \$0 in savings, \$0 in financial investments, and \$0 in home equity. Six years ago, your household rented its home, household income was \$72,924, total net worth was \$23,610, savings were \$5,555, and financial investments were \$0. Four years ago, your household rented its home, household income was \$25,142, total net worth was \$13,307, savings were \$0, and financial investments were \$0. Two years ago, your household rented its home, household income was \$71,953, total net worth was \$5,157, savings were \$5,157, and financial investments were \$0.\par

You are also aware of recent national economic conditions. Over the four most recently completed calendar quarters, listed from oldest to most recent, U.S. real GDP changed by -0.35\%, 1.29\%, 1.22\%, and 0.51\% quarter over quarter; the unemployment rate was 6.67\%, 6.20\%, 6.07\%, and 5.70\%; overall consumer prices changed by 0.62\%, 0.53\%, 0.26\%, and -0.25\% quarter over quarter; and the effective federal funds rate was 0.07\%, 0.09\%, 0.09\%, and 0.10\%. Over the five years ending in the most recently completed quarter, compound average quarterly real GDP growth was 0.57\%, the average unemployment rate was 8.03\%, compound average quarterly CPI inflation was 0.44\%, and the average effective federal funds rate was 0.12\%. You are also aware of recent financial and housing-market conditions. Over the same four completed quarters, listed from oldest to most recent, U.S. house prices changed by 0.49\%, 1.50\%, 1.06\%, and 0.54\% quarter over quarter; the S\&P 500 changed by 3.61\%, 3.60\%, 3.98\%, and 1.83\% quarter over quarter; and the average 30-year mortgage rate was 4.36\%, 4.23\%, 4.14\%, and 3.97\%. Over the same five-year period, compound average quarterly house-price growth was -0.15\%, compound average quarterly S\&P 500 growth was 3.12\%, and the average 30-year mortgage rate was 4.19\%.\par

In two years, will your household own its home?\par

Return only a valid JSON array containing exactly one string, using one of: "yes", "no". Your output must follow this exact array structure: ["s\_\allowbreak{}1"]. Replace every s placeholder with one of the permitted strings. Do not include explanatory text.\par
\paragraph{\fontsize{7.2}{8.55}\selectfont Assistant.} ["no"]\par
\endgroup
\end{tcolorbox}

\hypertarget{task:house_sce_move}{}
\subsection{house\_\allowbreak{}sce\_\allowbreak{}move}

\textbf{Target.} The respondent's stated chance of moving to a different primary residence over the next 12 months. The answer comes directly from an SCE question asking for a probability from 0 to 100 percent. For evaluation, we express it as the predicted probabilities that the respondent moves and does not move.\par

\textbf{Predictors.} Previous expectation (the same stated moving probability one month earlier), residence and moving circumstances (current residence and moving-related answers when available), household finances and expectations (current and expected finances), work and job-search situation (current work-status indicators and search information when available), respondent demographics, resources, and geography (age, education, household before-tax income bracket, region, numeracy), and core macro context (real GDP growth, unemployment rate, headline CPI inflation, and the effective federal funds rate over the four most recently completed quarters, plus five-year quarterly references).\par

\textbf{No-change baseline.} The no-change prediction is the respondent's stated moving probability one month earlier.\par

\textbf{Sample construction.} We link each SCE Core survey record to the respondent's interview in the immediately preceding month and merge national economic variables from FRED-QD. Next, we drop rows that are missing the current moving-probability target, the corresponding first-history probability, or any required demographic, income, or macroeconomic predictor. Current residence and moving circumstances, household-finance measures, work-status indicators, and job-search information are allowed to remain missing. We then require both probabilities to lie between 0 and 100 percent and respondents to be between 18 and 100 years old.\par

\begin{tcolorbox}[
  breakable,
  lines before break=8,
  colback=white,
  colframe=black!55,
  boxrule=0.45pt,
  arc=0pt,
  left=7pt,
  right=7pt,
  top=7pt,
  bottom=7pt
]
\begingroup
\setlength{\parindent}{0pt}
\setlength{\parskip}{5pt}
\hyphenpenalty=1800
\exhyphenpenalty=1800
\emergencystretch=0.8em
{\centering\fontsize{7.7}{9.0}\selectfont\bfseries Prompt example\par
\texttt{house\_sce\_move}\par}
\medskip
{\color{black!25}\hrule height 0.35pt}
\vspace{5pt}
\fontsize{7.2}{8.55}\selectfont\rmfamily
\paragraph{\fontsize{7.2}{8.55}\selectfont System.} You are an American adult making decisions about housing and residential location.\par
\paragraph{\fontsize{7.2}{8.55}\selectfont User.}
Here is some background information about yourself and your household. You are a 58-year-old, have some college education, live in the West, and have a low numeracy score.\par

Here are your household finances, work situation, expectations, and housing. Note that the current month is July, and that \$ denotes amounts in U.S. dollars. Your household's total pre-tax income over the past 12 months was between \$30,000 and \$39,999. Your household is about the same than it was 12 months ago, and you expect your household finances to be somewhat better off 12 months from now. You have not moved recently. One month ago, you thought the probability that you would move to a different primary residence over the following 12 months was 0.300.\par

You are also aware of recent national economic conditions. Over the four most recently completed calendar quarters, listed from oldest to most recent, U.S. real GDP changed by 0.14\%, 0.12\%, 0.99\%, and 0.27\% quarter over quarter; the unemployment rate was 8.03\%, 7.80\%, 7.73\%, and 7.53\%; overall consumer prices changed by 0.45\%, 0.66\%, 0.40\%, and -0.11\% quarter over quarter; and the effective federal funds rate was 0.14\%, 0.16\%, 0.14\%, and 0.12\%. Over the five years ending in the most recently completed quarter, compound average quarterly real GDP growth was 0.22\%, the average unemployment rate was 8.59\%, compound average quarterly CPI inflation was 0.37\%, and the average effective federal funds rate was 0.25\%.\par

What probability do you think there is that over the next 12 months you will move to a different primary residence?\par

Return only a valid JSON array containing exactly one two-element probability distribution. Use this order: probability of not moving, probability of moving. Express each probability as a number from 0 to 1. The distribution must sum to 1. Your output must follow this exact array structure: [[p\_\allowbreak{}1\_\allowbreak{}1, p\_\allowbreak{}1\_\allowbreak{}2]]. Replace every p placeholder with a number from 0 to 1. Do not include keys or explanatory text.\par
\paragraph{\fontsize{7.2}{8.55}\selectfont Assistant.} [[1.0, 0.0]]\par
\endgroup
\end{tcolorbox}

\hypertarget{task:house_sce_financing}{}
\subsection{house\_\allowbreak{}sce\_\allowbreak{}financing}

This task is inspired by \citet{fuster2021sensitivity}.\par

\textbf{Target.} The respondent's maximum home purchase price and down payment under three financing scenarios: freedom to choose any down payment at the original mortgage rate, a mortgage rate two percentage points above or below the original rate, and receipt of a cash inheritance. The six targets are the direct dollar answers to the price and down-payment questions under these three scenarios.\par

\textbf{Predictors.} Initial housing choice (maximum purchase price, down payment, mortgage rate, monthly payment, comparable-home value), household housing and financial circumstances (tenure, savings, debt, credit-score range, moving expectations, views about property investment), respondent preferences and characteristics (risk tolerance, numeracy, demographics, income, location), and core macro context (national economic conditions over the most recently completed quarters). The experiment randomly assigns an initial mortgage rate of 4.5 or 6.5 percent, then allows a flexible down payment, changes the rate by two percentage points, and adds an inheritance. The model sees the initial choice and predicts all six later answers together.\par

\textbf{No-change baseline.} The no-change prediction repeats the respondent's initial maximum price and down payment under each of the three later scenarios.\par

\textbf{Sample construction.} Starting from the February 2014 home-financing experiment data from \citet{fuster2021sensitivity}'s replication package, we add the survey release date and national economic variables available before the survey. Next, we drop rows that are missing one or more of the six targets or any current predictor. Current home value and housing debt are required for homeowners but are allowed to remain missing for renters; no other predictor shown to the model is allowed to remain missing. We require every stated purchase price to be positive, each down payment to be at least the experiment's rounded 5 percent minimum and no greater than the stated price, and all answers to lie within the ranges allowed by the questionnaire. Finally, we trim the top and bottom percentiles of all the monetary variables.\par

\begin{tcolorbox}[
  breakable,
  lines before break=8,
  colback=white,
  colframe=black!55,
  boxrule=0.45pt,
  arc=0pt,
  left=7pt,
  right=7pt,
  top=7pt,
  bottom=7pt
]
\begingroup
\setlength{\parindent}{0pt}
\setlength{\parskip}{5pt}
\hyphenpenalty=1800
\exhyphenpenalty=1800
\emergencystretch=0.8em
{\centering\fontsize{7.7}{9.0}\selectfont\bfseries Prompt example\par
\texttt{house\_sce\_financing}\par}
\medskip
{\color{black!25}\hrule height 0.35pt}
\vspace{5pt}
\fontsize{7.2}{8.55}\selectfont\rmfamily
\paragraph{\fontsize{7.2}{8.55}\selectfont System.} You are an American adult making decisions about housing, borrowing, and household finances.\par
\paragraph{\fontsize{7.2}{8.55}\selectfont User.}
Here is some background information about yourself and your household. You are a 40-year-old woman, are married, and your highest completed education is less than a college degree. You live in the West and have 2 children under 18 and 1 adult child living with you. You answered 2 of 5 numeracy questions correctly.\par

Here are your household finances, housing situation, preferences, and expectations. Note that \$ denotes amounts in U.S. dollars. Your household income is between \$10,000 and \$20,000. You rent your current home. Your liquid savings are less than \$5,000, your non-housing debt is between \$5,000 and \$30,000, and your credit-score band is below 620. A typical comparable home costs about \$230,000. On a scale from 1 to 10, your willingness to take risks is 2. You give a 50.0\% chance of moving within the next three years. You do not regard property in your ZIP code as a good investment.\par

You are also aware of recent national economic conditions. Over the four most recently completed calendar quarters, listed from oldest to most recent, U.S. real GDP changed by 0.99\%, 0.27\%, 0.85\%, and 0.87\% quarter over quarter; the unemployment rate was 7.73\%, 7.53\%, 7.23\%, and 6.93\%; overall consumer prices changed by 0.40\%, -0.11\%, 0.54\%, and 0.37\% quarter over quarter; and the effective federal funds rate was 0.14\%, 0.12\%, 0.08\%, and 0.09\%. Over the five years ending in the most recently completed quarter, compound average quarterly real GDP growth was 0.44\%, the average unemployment rate was 8.65\%, compound average quarterly CPI inflation was 0.45\%, and the average effective federal funds rate was 0.14\%.\par

Imagine that you move to a comparable city and are considering a comparable home where you plan to remain for a long time. You do not need to sell a current home. With a 20\% down payment and a 30-year mortgage rate of 4.5\%, the most you would pay for the home is \$60,000, implying a down payment of \$12,000. This is equivalent to an effective total monthly payment of about \$827 for a 30-year fixed-rate loan, including principal and interest, estimated mortgage-interest tax savings assuming that 95\% of interest is deductible at a 34\% income tax rate, and maintenance, property taxes, and homeowners insurance equal in total to 3.35\% of the comparable-home value per year. For reference, the principal-and-interest payment alone on each \$100,000 borrowed is about \$507 per month at 4.5\% and \$632 per month at 6.5\%.\par

Now consider three further situations, in this order.\par

Situation 1: The mortgage rate remains 4.5\%, but you may choose any down payment of at least 5\% of the home price. You receive no extra cash before purchase.\par

Situation 2: The mortgage rate is 6.5\%, and you may again choose any down payment of at least 5\% of the home price. You receive no extra cash before purchase.\par

Situation 3: The mortgage rate remains 6.5\%, you may choose any down payment of at least 5\%, and you unexpectedly receive \$100,000 in cash before buying the home. You may use all, some, or none of this cash for the down payment.\par

For each situation, state the maximum home price you would be willing to pay and the dollar down payment you would choose.\par

Return only a valid JSON array with exactly 6 numbers in this order: situation\_\allowbreak{}1\_\allowbreak{}maximum\_\allowbreak{}price, situation\_\allowbreak{}1\_\allowbreak{}down\_\allowbreak{}payment, situation\_\allowbreak{}2\_\allowbreak{}maximum\_\allowbreak{}price, situation\_\allowbreak{}2\_\allowbreak{}down\_\allowbreak{}payment, situation\_\allowbreak{}3\_\allowbreak{}maximum\_\allowbreak{}price, situation\_\allowbreak{}3\_\allowbreak{}down\_\allowbreak{}payment. All six numbers must be nominal U.S. dollars. Your output must follow this exact array structure: [v\_\allowbreak{}1, v\_\allowbreak{}2, v\_\allowbreak{}3, v\_\allowbreak{}4, v\_\allowbreak{}5, v\_\allowbreak{}6]. Replace every v placeholder with a single number. Do not include keys or explanatory text.\par
\paragraph{\fontsize{7.2}{8.55}\selectfont Assistant.} [65000, 5000, 65000, 7000, 85000, 35000]\par
\endgroup
\end{tcolorbox}

\hypertarget{task:house_sce_lockin}{}
\subsection{house\_\allowbreak{}sce\_\allowbreak{}lockin}

This task is inspired by \citet{aidala2024mortgage}.\par

\textbf{Target.} The respondent's stated chance of moving to a different primary residence over the next three years if they could keep their current mortgage interest rate after moving. The answer comes directly from an SCE question asking for a probability from 0 to 100 percent. For evaluation, we express it as the predicted probabilities that the respondent moves and does not move.\par

\textbf{Predictors.} Ordinary moving expectation (the respondent's stated three-year moving probability without mortgage portability), mortgage and housing circumstances (current mortgage rate, home value, mortgage balance, homeownership, the latest national 30-year mortgage rate, and the difference between the current and national rates), household finances and expectations (current and expected finances), work situation (current work-status indicators), household composition and respondent demographics, resources, and geography (age, education, household before-tax income bracket, region, numeracy, marital or partner status), and core macro context (national economic conditions over the most recently completed quarters). The hypothetical scenario allows the household to carry its current mortgage rate to a new home.\par

\textbf{No-change baseline.} The no-change prediction is the respondent's ordinary three-year moving probability reported in the same survey.\par

\textbf{Sample construction.} We merge the SCE Housing module with respondent information from the SCE Core survey, national economic variables from FRED-QD, and the national 30-year mortgage rate. We keep respondents with an active mortgage. Next, we drop rows that are missing the moving-probability target under mortgage portability; the ordinary moving probability; current mortgage status or rate; or any required demographic, income, macroeconomic, or national-mortgage-rate predictor. Current home values, mortgage balance, spouse or partner status, household-finance measures, and work-status indicators are allowed to remain missing. We then require both the ordinary and mortgage-portability moving probabilities to lie between 0 and 100 percent, the current mortgage rate to be positive, and respondents to be between 18 and 100 years old. Finally, within each survey month, we trim the top and bottom percentiles of the current mortgage rate and observed housing values.\par

\begin{tcolorbox}[
  breakable,
  lines before break=8,
  colback=white,
  colframe=black!55,
  boxrule=0.45pt,
  arc=0pt,
  left=7pt,
  right=7pt,
  top=7pt,
  bottom=7pt
]
\begingroup
\setlength{\parindent}{0pt}
\setlength{\parskip}{5pt}
\hyphenpenalty=1800
\exhyphenpenalty=1800
\emergencystretch=0.8em
{\centering\fontsize{7.7}{9.0}\selectfont\bfseries Prompt example\par
\texttt{house\_sce\_lockin}\par}
\medskip
{\color{black!25}\hrule height 0.35pt}
\vspace{5pt}
\fontsize{7.2}{8.55}\selectfont\rmfamily
\paragraph{\fontsize{7.2}{8.55}\selectfont System.} You are an American adult making decisions about housing and residential location.\par
\paragraph{\fontsize{7.2}{8.55}\selectfont User.}
Here is some background information about yourself and your household. You are a 38-year-old, have a high-school education, live in the Midwest, have a high numeracy score, and are married or living with a partner.\par

Here are your household finances, work situation, expectations, housing, and mortgage situation. Note that the current month is February, and that \$ denotes amounts in U.S. dollars. Your household's total pre-tax income over the past 12 months was between \$150,000 and \$199,999. Your household is about as well off as it was 12 months ago, and you expect your household finances to remain about the same 12 months from now. You expect inflation to be -0.5\% over the next 12 months and -2.6\% over the next three years. You work full-time. A typical local home is worth about \$200,000, while your primary residence was purchased for about \$520,000 and is now worth about \$600,000. Your home loan consists of mortgage debt only, with a balance of about \$400,000 and an interest rate of 3.0\%. The national average 30-year mortgage rate is 6.7\%, which is 3.7 percentage points above your current mortgage rate. Given your current situation, you think the probability that you would move to a different primary residence over the next three years is 0.010.\par

You are also aware of recent national economic conditions. Over the four most recently completed calendar quarters, listed from oldest to most recent, U.S. real GDP changed by -0.25\%, 0.16\%, 0.72\%, and 0.69\% quarter over quarter; the unemployment rate was 3.87\%, 3.63\%, 3.53\%, and 3.57\%; overall consumer prices changed by 2.20\%, 2.35\%, 1.32\%, and 1.05\% quarter over quarter; and the effective federal funds rate was 0.12\%, 0.77\%, 2.19\%, and 3.65\%. Over the five years ending in the most recently completed quarter, compound average quarterly real GDP growth was 0.57\%, the average unemployment rate was 4.93\%, compound average quarterly CPI inflation was 0.95\%, and the average effective federal funds rate was 1.23\%.\par

For this question, consider the following mortgage-rate scenario. Suppose that, if you moved and bought a different home, you could keep the same interest rate as your current mortgage.\par

Under that scenario, what probability do you think there is that you would move to a different primary residence over the next three years?\par

Return only a valid JSON array containing exactly one two-element probability distribution. Use this order: probability of not moving under the scenario, probability of moving under the scenario. Express each probability as a number from 0 to 1. The distribution must sum to 1. Your output must follow this exact array structure: [[p\_\allowbreak{}1\_\allowbreak{}1, p\_\allowbreak{}1\_\allowbreak{}2]]. Replace every p placeholder with a number from 0 to 1. Do not include keys or explanatory text.\par
\paragraph{\fontsize{7.2}{8.55}\selectfont Assistant.} [[0.99, 0.01]]\par
\endgroup
\end{tcolorbox}

\end{document}